\RequirePackage{fix-cm}
\documentclass[smallextended]{svjour3}       
\smartqed  
\usepackage{graphicx}
\usepackage{color}
\usepackage{cite}
\usepackage{subfigure}
\usepackage{booktabs}
\usepackage{multirow}
\usepackage{tabulary}
\usepackage{url}
\usepackage[figuresright]{rotating}
\usepackage{longtable}
\usepackage{rotating}
\usepackage{adjustbox}

\begin{document}
\bibliographystyle{unsrt}

\title{What Can Machine Vision Do for Lymphatic Histopathology Image Analysis: A Comprehensive Review}


\author{Xiaoqi Li \and Haoyuan Chen \and Chen Li \and Md Mamunur Rahaman \and 
Xintong Li \and Jian Wu \and Xiaoyan Li \and Hongzan Sun \and Marcin Grzegorzek}


\institute{Xiaoqi Li, Chen Li, Xintong Li, Jian Wu \at
              Microscopic Image and Medical Image Analysis Group, College of Medicine and Biological Information Engineering, Northeastern University, Shenyang, 110169, China \\
              Corresponding author: Chen Li, E-mail: lichen201096@hotmail.com
           \and 
           Md Mamunur Rahaman
           School of Computer Science and Engineering, University of New South Wales, Sydney, NSW 2052, Australia
           \and
           Xiaoyan Li, Hongzan Sun \at
               China Medical University, Shenyang, 110001, China\\
           \and 
           Marcin Grzegorzek \at
               Institute of Medical Informatics, University of Luebeck, Luebeck, Germany 
}

\date{Received: date / Accepted: date}

\maketitle

\begin{abstract}
Over the past ten years, machine vision (MV) algorithms for image analysis have been developing rapidly with computing power. At the same time, histopathological slices can be stored as digital images. Therefore, MV algorithms can provide diagnostic references to doctors. In particular, the continuous improvement of deep learning algorithms has further improved the accuracy of MV in disease detection and diagnosis. This paper reviews the application of image processing techniques based on MV in lymphoma histopathological images in recent years, including segmentation, classification and detection. Finally, the current methods are analyzed, some potential methods are proposed, and further prospects are made. 
\end{abstract}

\section{Introduction}
\label{Sec:1}
\subsection{Basic Clinical Information about Lymph}
\label{Sec:1.1}
Lymph is also known as lymphatic fluid, a colorless and transparent liquid in humans and animals. It contains lymphocytes and is partly formed by tissue fluid infiltrating into the lymphatic vessels. Lymph exists in all parts of the human body and plays an important role in the immune system. The lymph nodes are organized as lymphoid organs, containing lymphocytes within a fine reticular stroma~\cite{Elmore-2006-Histopathology}. The lymphatic system consists of a network of vessels, termed lymphatics, lymph nodes and lymphoid organs. It is an important part of the circulatory and immune systems and plays an important role in homeostasis by controlling extracellular fluid volume and combating infection~\cite{Margaris-2012-Modelling}. For example, the lymphatic system can drain the accumulated fluid to restore normal fluid circulation if the body tissue swells in injury.

Malignant lymphoma, also known as lymph cancer, is a malignant tumor that originates in the lymph nodes and tissues of organs, which is caused by the malignant proliferation of lymphocytes and tissue cells. According to the different pathological characteristics, malignant lymphoma includes two categories: Hodgkin’s lymphoma (HL) and Non-Hodgkin’s lymphoma (NHL). HL accounts for about 10$\%$ of all lymphomas, and NHL accounts for 90$\%$ of the remaining lymphomas~\cite{Shankland-2012-Non}. According to World Health Organization’s (WHO), HL includes two categories: classic Hodgkin's lymphoma (CHL) and nodular lymphocyte-predominant Hodgkin's lymphoma (NLPHL). HL is one of the most common lymphomas in the Western world, with approximately 3 cases per 100,000 persons every year~\cite{Kuppers-2012-Hodgkin}. In 2015, there were approximately 9,050 new HL cases in the United States~\cite{Siegel-2015-Cancer}. The exact cause of HL is unclear, but the factors that increase the risk of HL include exposure to viral infection, family factors, and immunosuppression~\cite{Ansell-2015-Hodgkin}. Nevertheless, HL is a curable malignant tumor~\cite{Gobbi-2013-Hodgkin}, today's medical technology has made great progress, soaring with the possibility of a cure for patients with HL. Currently, over 80$\%$ of patients newly diagnosed with HL can be cured~\cite{Ansell-2015-Hodgkin}. NHL is a group of heterogeneous malignant tumors of the lymphatic system. NHL results in many deaths worldwide, and its incidence has been rising. According to the WHO classification of blood and lymphoid tumors, NHL includes B-cell and T-cell tumors. B-cell lymphoma accounts for about 90$\%$ of all lymphomas~\cite{Ansell-2005-Non}. In the United States, among new cancers, NHL accounts for 5$\%$ of men, and 4$\%$ of women each year, and causes 5$\%$ of all cancer deaths~\cite{Greenlee-2000-Cancer}. Although some cases are related to immunodeficiency, autoimmunity, or viral infection, the causes of NHL are unclear in most cases~\cite{Hennessy-2004-Non}. 

Although some of the causes of lymphoma as mentioned above are not clear, pathologists and doctors have not given up exploring the causes of lymphoma. Traditionally, lymphoma is diagnosed by surgical excision biopsy~\cite{Vandervelde-2008-Study}. Biopsy refers to the use of local excision~\cite{Gannon-2006-Accuracy}, forceps~\cite{Herth-2012-Endobronchial}, fine-needle aspiration~\cite{Amedee-2001-Fine}, scraping~\cite{Teal-2007-Evaluation} to obtain diseased tissue from the living body for pathological diagnosis. Besides, due to the symptoms of some lymphatic tissue diseases are not typical, and imageology lacks specificity, a biopsy is frequently required for clinical diagnosis. The diagnosis of NHL is most easily established by examining the tissue obtained by open biopsy of the involved site~\cite{Sandlund-2020-Childhood}, and the initial diagnosis of HL can only be made by biopsy~\cite{Ansell-2018-Hodgkin}. For example, lymphatic mapping with sentinel lymph node biopsy is widely used to reduce complications associated with axillary lymph node dissection in low-risk breast cancer patients~\cite{Kim-2006-Lymphatic}. Diffuse large B-cell lymphoma (DLBCL) is diagnosed ideally by biopsy of abnormally enlarged and suspected lymph nodes during clinical examination and radiographic imaging~\cite{Liu-2019-Diffuse}. 

In addition to the biopsy mentioned above, the commonly used methods in the diagnosis of lymphoma include histopathological examination and immunohistochemistry (IHC). In histopathological observation, the removed diseased tissues are fixed with formalin solution and embedded in paraffin to make tissue sections~\cite{Wong-2001-Use}. And then after different staining, the tissues sections are observed with an optical microscope. By analyzing and synthesizing the characteristics of the diseased tissue, the pathological diagnosis of the disease can be made. The most commonly used staining method for tissue sections is hematoxylin-eosin (H$\&$E) staining, which is usually the only staining method used in lymph node diagnosis~\cite{Wright-2011-Diagnostic}. Moreover, most H$\&$E stained tissue sections can be used for initial analysis~\cite{Perkins-2007-Biology}, the pathological evaluation of sentinel nodes also includes H$\&$E staining~\cite{Terwisscha-2006-Sentinel}. H$\&$E staining can help detect cells that are not typical, and determine their histological affiliation~\cite{Lepathology}. Currently, the intraoperative histopathological examination of sentinel lymph nodes is mainly based on frozen sections stained with H$\&$E~\cite{Nahrig-2003-Intraoperative}. If the histopathological observation failed to make a diagnosis or requires further research, observation techniques such as IHC can be assisted. IHC is very helpful to confirm the diagnosis with a high sensitivity~\cite{Lepathology}~\cite{Kansal-2002-Follicular}. The presence of lymph node metastasis is one of the most important prognostic factors and treatment options for many kinds of cancer~\cite{Mesker-2003-Automated}. However, sometimes micrometastasis in lymph nodes cannot be detected only based on H$\&$E staining, and can only be detected after IHC investigation on paraffin-embedded tissues~\cite{Lepathology}~\cite{Toda-2011-Novel}. For example, in~\cite{Cote-1999-Role}, 20$\%$ of breast cancer patients have occult lymph node metastasis detected by IHC, and only 7$\%$ have been detected by H$\&$E staining. Therefore, the laboratory should control the quality of reagents and equipment to ensure that histopathological observations such as H$\&$E staining and IHC can make a correct diagnosis of lymphoma.

\begin{figure}[htbp!]
\centering
\includegraphics[width=0.98\linewidth]{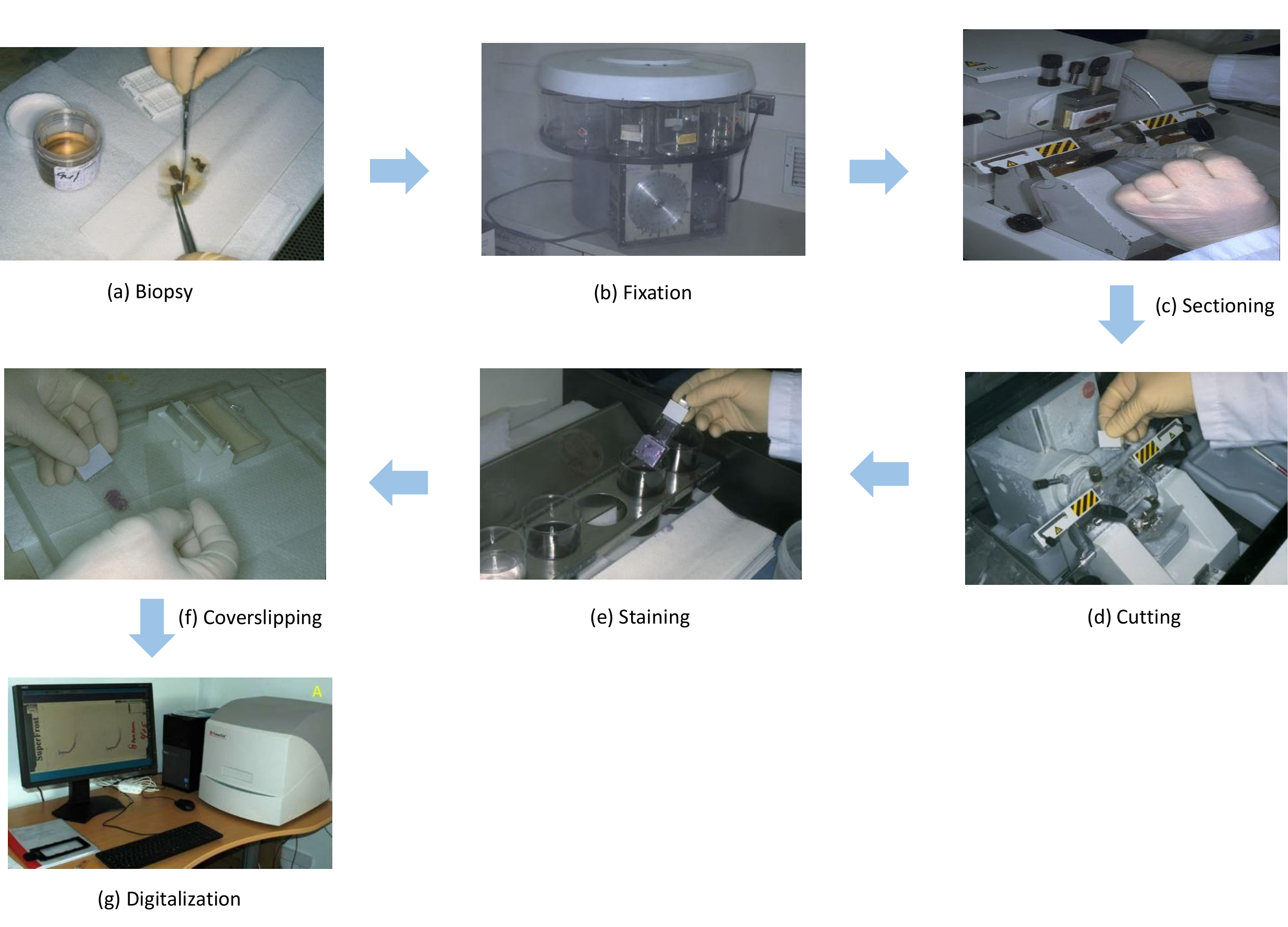}
\caption{The process of histopathology image acquisition. ((a)-(f) are taken from http://library.med.utah.edu/WebPath/HISTHTML/HISTOTCH/HISTOTCH.html. (g) corresponds to Fig.1 (a) in~\cite{Phdthesis})}
\label{fig:preparation}
\end{figure}

To make the above content more intuitive, we can observe the acquisition process of the histopathological image in Fig.~\ref{fig:preparation}. First, a biological sample is taken from a living body. Then, the biopsy is fixed to avoid chemical changes in the tissue~\cite{Hewitson-2010-Histology}. After that, the tissue is cut into sections to be placed on a glass slide for staining. Then the stained tissue is covered on the slide with a thin piece of plastic or glass, which can protect the tissue and facilitate observation under a microscope. Finally, the slide is observed and digitized with a microscope.

\begin{figure}[htbp!]
\centering
\includegraphics[width=1\linewidth]{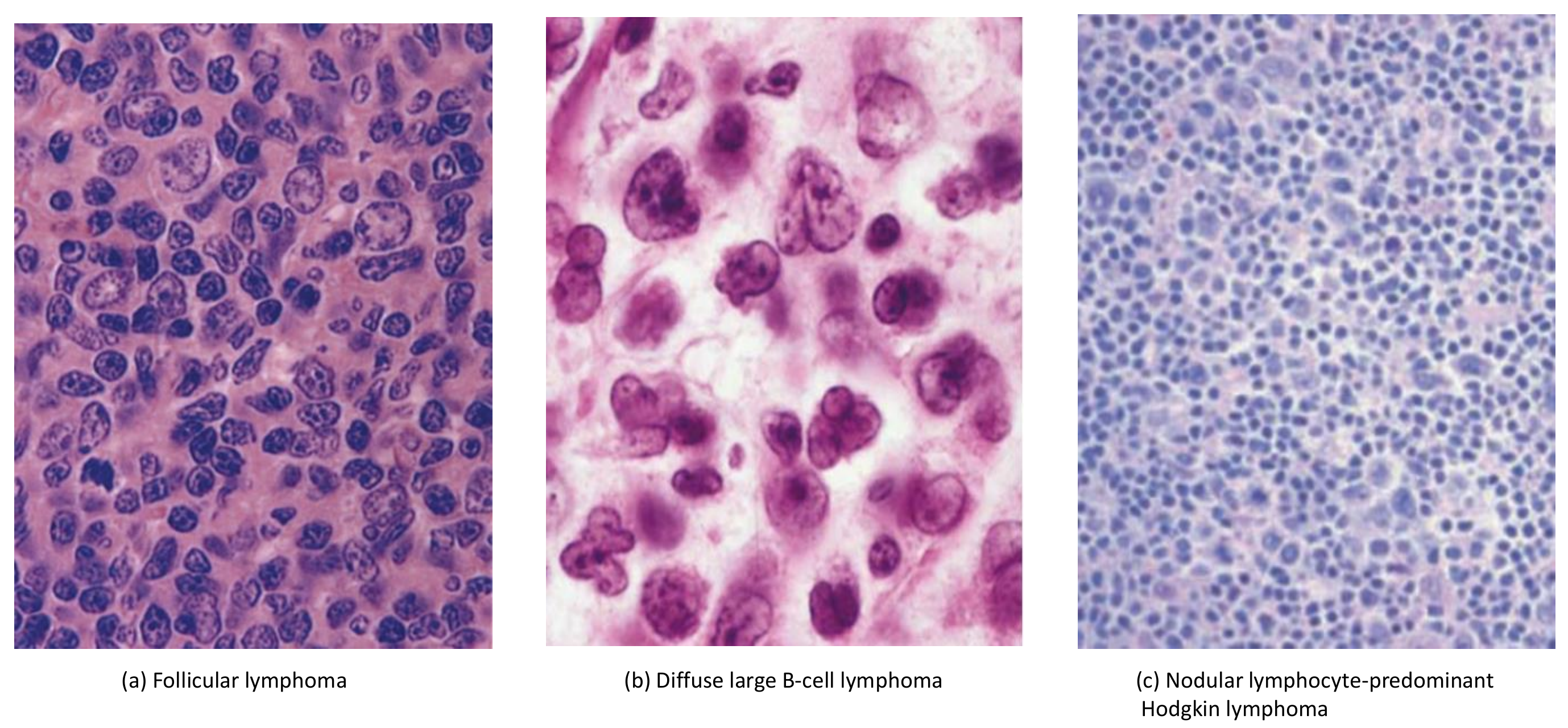}
\caption{Pathological pictures of three types of lymphomas (H$\&$E stain). (a) corresponds to Fig. 6.67 (b) in ~\cite{Feller-2003-Histopathology}. (b) corresponds to Fig. 6.61 in ~\cite{Weinstein-2007-Pediatric}. (c) corresponds to Fig. 3.1 in ~\cite{Engert-2016-Hodgkin}.}
\label{fig:three}
\end{figure}

In this paragraph, we introduce the morphological features of some typical lymphomas. Because every type of lymphoma has a characteristic morphology, so many lymphomas can be accurately subcategorized based on morphological characteristics~\cite{Xu-2002-Assessment}. Pathologist's morphological analysis (i.e., tissue structure, cell shape, and staining characteristics) is still the gold standard for the current routine lymph node histopathological staging and grading~\cite{Isabelle-2010-Correlation}~\cite{Pannu-2000-Mr}. Moreover, because the prognosis of patients with lymphoma varies greatly, the morphology of lymphoma is the main determinant of treatment outcome and prognosis~\cite{Devita-2001-Cancer}. In NHL, follicular lymphoma (FL) is the common type of B-cell lymphoma (As shown in Fig.~\ref{fig:three} (a)). FL cells usually show a distinct follicular-like growth pattern, with follicles similar in size and shape, with unclear boundaries. Tumorous follicles are mainly composed of centrocyte (CC) and centroblast (CB) in different proportions. The CC of FL is small in size and medium in large, with irregular nuclei and inconspicuous nucleoli. The CB has a large volume, 2 to 3 times larger than normal lymphocytes. DLBCL is a diffusely hyperplastic large B cell malignant tumor. The cell morphology of DLBCL is diverse, the nucleus is round or oval, with single or multiple nucleoli (As shown in Fig.~\ref{fig:three} (b)). In HL, NLPHL presents a vaguely large nodular shape, and the background structure is a large spherical network composed of follicular dendritic cells (As shown in Fig.~\ref{fig:three} (c)).

Early detection of lymphoma can greatly increase the possibility of cure, so a fast and reliable way is needed for early detection and screening. Cytological and histological techniques are generally used to examine biopsy tissue~\cite{Sattlecker-2011-0ptimisation}. However, these techniques have several limitations. For instance, they have a high subjectivity, because the result of diagnosis depends on the pathologists' opinion. Therefore, different pathologists may have different diagnostic results when examining the same tissue sample~\cite{Robbins-1995-Histological}. In addition, the examination process for biopsy tissue is very time-consuming~\cite{Crow-2003-0ptical}. At the same time, problems with the preparation of slides and differences in staining can also cause pathologists to make mistakes when analyzing diseases. For example, lymph node metastasis is one of the most important factors for the prognosis of breast cancer. The status of regional lymph nodes is generally assessed through sentinel lymph node (SLN) surgery. The SLN is removed, histopathologically processed and examined by a pathologist. This cumbersome examination process is likely to cause small metastases to be missed~\cite{Litjens-2018-1399}~\cite{Van-2010-Pathology}. Due to the increase in the incidence of lymphoma and the treatment of specific patients, the diagnosis and grading of lymphoma have become more and more complicated~\cite{Litjens-2016-Deep}. Nowadays, pathologists must examine a large number of slides, and their histopathological cancer diagnosis workload and complexity have increased dramatically. Finally, the more serious problem is that the number of pathologists has decreased a lot in the past ten years~\cite{Kandel-2020-Novel}. The author of ~\cite{Robboy-2013-Pathologist} estimates that the number of pathologists in the United States will decrease from 5.7 to 3.7 per 100,000 in the next 20 years. Therefore, histopathological examination technology and biopsy technology for lymphoma urgently need to be improved. This is not only the liberation of the pathologist's hands and eyes but also the responsibility for the patient.

\subsection{The Development of Machine Vision in the Diagnosis and Treatment of Lymphoma}
\label{Sec:1.2}
Many kinds of diseases can be evaluated and diagnosed by analyzing cells and tissues. As mentioned above, most of the analysis is performed manually by pathologists and the accuracy of the results largely depends on the proficiency of pathologists. Besides, with the increasing workload and limited number of pathologists, \emph{Computer-aided Diagnosis} (CAD) systems become a major research subject~\cite{Bergeron-2017-Investigation}. A CAD system generally consists of three parts~\cite{Samsi-2012-Computer}. First, CAD systems segment the object from the background tissue. Second, CAD systems extract features related to the classification. Finally, a diagnosis result based on the extracted features is obtained from a classifier. In recent years, CAD has become a major research direction of medical image analysis. Usually, the medical images that require computer assistance are histopathological slides. Compared with pathologists, the analysis of diseases by CAD systems is performed quickly, and usually, only one sample image is needed to get accurate results~\cite{Di-2015-Different}. With CAD systems, pathological images can be automatically classified. This can enable doctors to be more efficient in diagnosing diseases, and the diagnosis results can be more accurate and objective~\cite{Zhang-2016-Large}~\cite{Zhang-2015-Fine}. Today, CAD is used in many fields of medicine. For example, CAD systems play a very important role in the early detection of breast cancer~\cite{Rangayyan-2007-Review}, lung cancer diagnosis~\cite{Reeves-2000-Computer}, and arrhythmia detection~\cite{Oweis-2006-Computer}. In~\cite{Kourou-2015-Machine}, a CAD system is mainly used in cancer diagnosis and prognosis. ~\cite{Kong-2008-New} describes a CAD system that can classify stromal development and neuroblastoma grades.

\begin{figure}[htbp!]
\centering
\centerline{\includegraphics[width=0.9\textwidth]{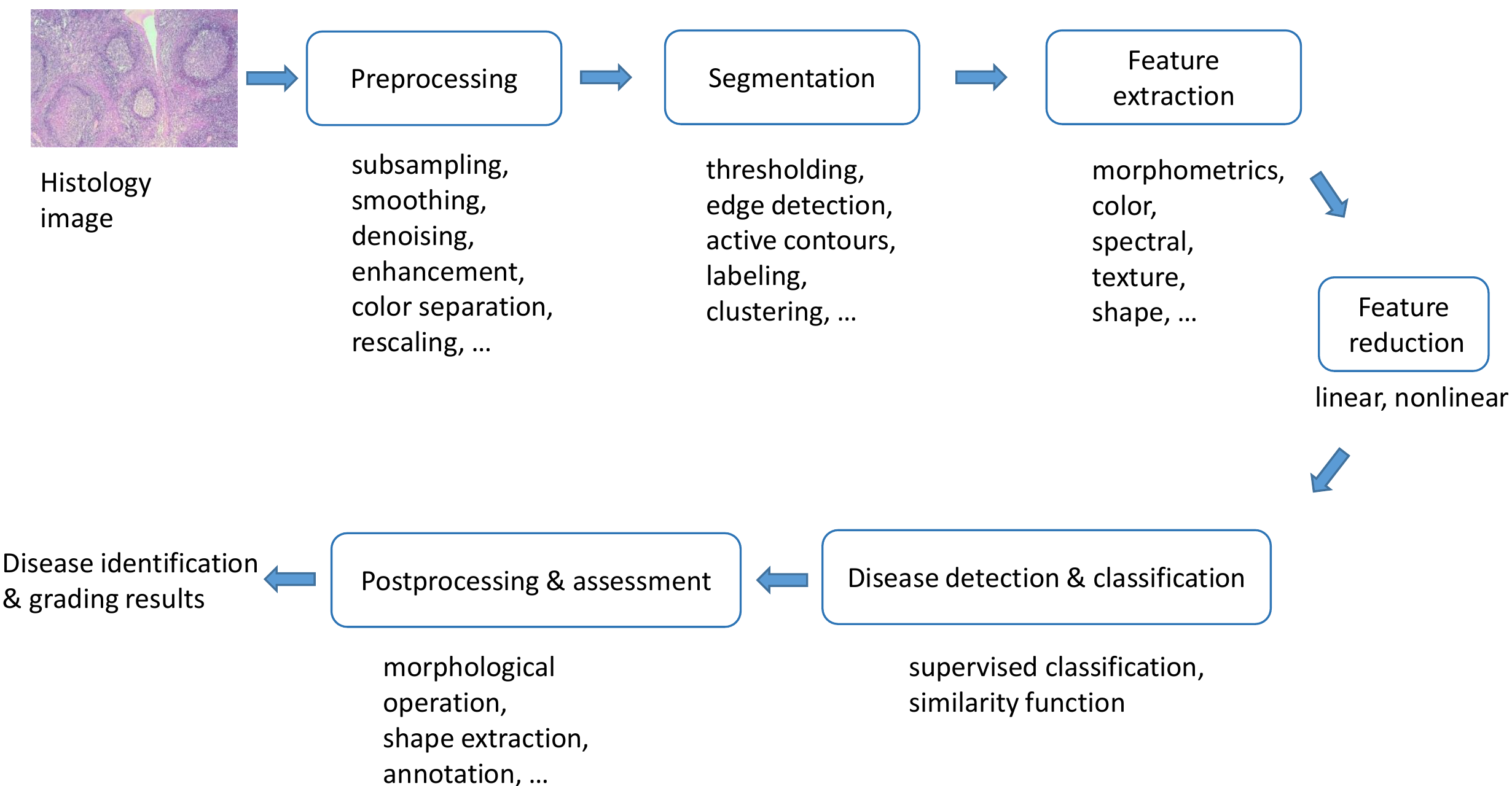}}
\caption{Computer-assisted diagnostic-system flowchart.}
\label{fig:CAD}
\end{figure}

A CAD system is composed of image preprocessing, segmentation, feature extraction, feature reduction, detection and classification, and post-processing modules (as shown in Fig.~\ref{fig:CAD}).

\begin{figure}[htbp!]
\centering
\centerline{\includegraphics[width=1\textwidth]{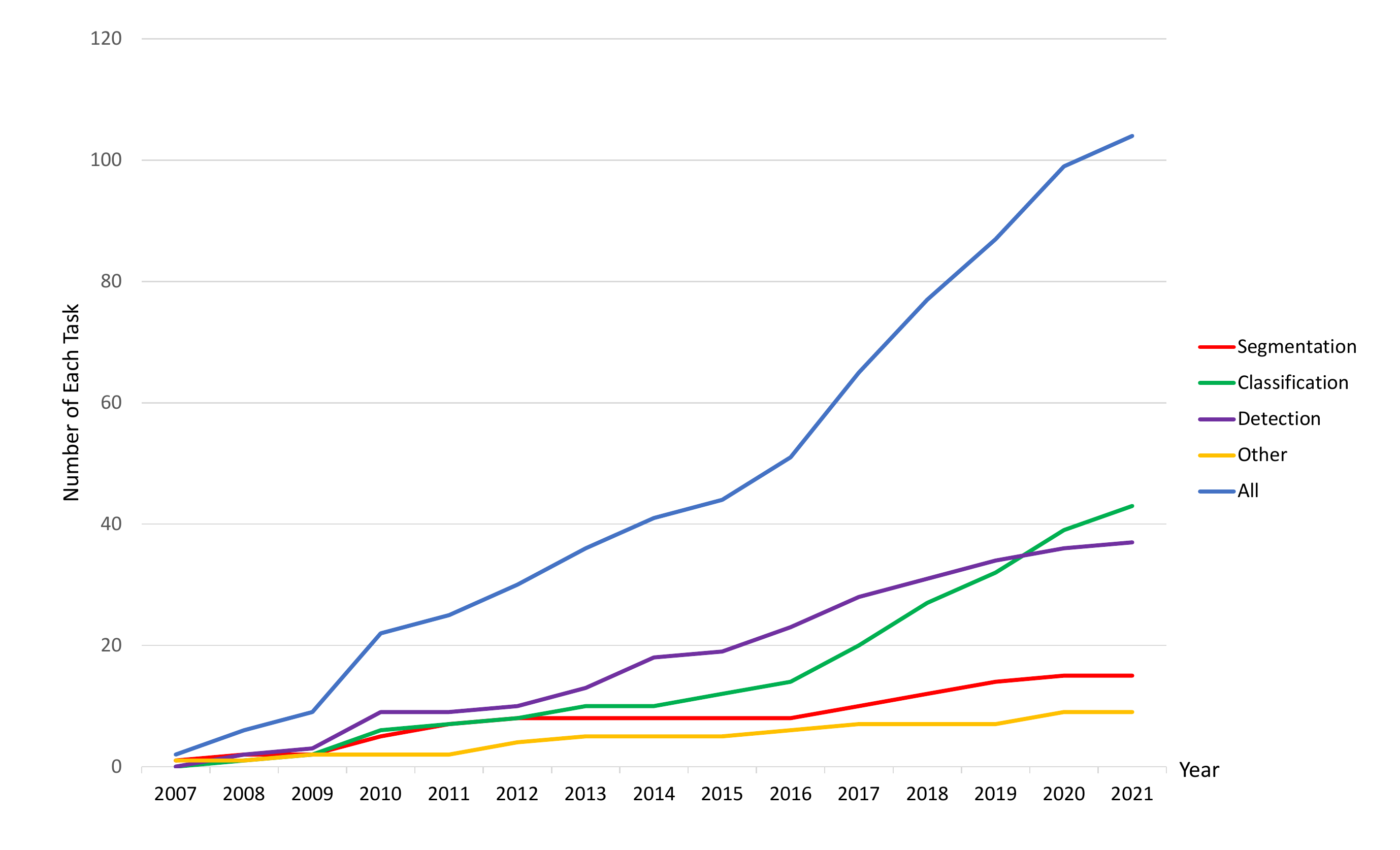}}
\caption{The development trend of the applications of MV in the diagnosis and treatment of lymphoma. The horizontal axis represents the year, and the vertical axis represents the number of studies.}
\label{fig:trend of MV}
\end{figure}

In CAD systems, MV is an important research direction. For example, in~\cite{Lee-2018-Deep}, a CAD system based on deep learning is developed to locate and differentiate metastatic lymph nodes of thyroid cancer using ultrasound images, a thriving example of MV as a high-sensitivity screening tool. In recent years, MV has developed rapidly and has received much attention, and it also represents the most cutting-edge research direction of deep learning. For example, from the earliest ImageNet large-scale visual recognition competition (ILSVRC)~\cite{Krizhevsky-2012-Imagenet} to the AlphaGo man-machine battles~\cite{Granter-2017-Alphago}, we have seen the powerful vitality and potential of computers in the field of vision processing. These superhuman performances and results can be achieved because of the development and improvement of deep learning and its deep neural network technology~\cite{Zhang-2018-Deep}. It can be seen that deep learning is driving the development of MV and artificial intelligence (AI) at high speed, and it is also changing our lives. At present, MV has been used in agriculture~\cite{Patricio-2018-Computer}, medicine~\cite{Obermeyer-2016-Predicting}, military fields~\cite{Budiharto-2020-Design} and various industrial applications. The current trends in the application of MV in the diagnosis and treatment of lymphoma are roughly shown in Fig.~\ref{fig:trend of MV}.

As shown in Fig.~\ref{fig:trend of MV}, the applications of MV in the diagnosis and treatment of lymphoma mainly include segmentation, classification, and detection. Besides, the number of applications as well as tasks has increased, such as quantitative analysis~\cite{Shi-2016-Automated}, analysis of characteristics of lymphoma~\cite{Martins-2019-Colour} and so on.

\subsection{Motivation of this Review}
\label{Sec:1.3}
This paper comprehensively reviews the MV methods for lymphatic histopathology image analysis. The motivation of this study is to study the popular technologies and trends of MV and to explore the future potential of MV in the diagnosis and treatment of lymphoma. In the course of our work, we found that some survey papers are related to the applications of MV in this area.

The survey of\cite{Loukas-2004-Survey} in 2004 reviews various image analysis methods which involve pathological image segmentation, cell detection and classification tasks. However, only one reference in this paper mentions the methods of extracting features for lymphoma classification.

The survey of\cite{Cruz-2006-Applications} in 2006 reviews the machine learning (ML) methods used in cancer prediction and prognosis. The author summarizes the percentage of papers using different ML methods in predicting lymphoma in a histogram. These ML methods include artificial neural networks (ANN), support vector machines (SVM), clustering, and so on. The paper does not review the applications of MV in lymphatic histopathology image analysis (LHIA).

The survey of\cite{Cai-2008-Content} in 2008 summarizes the current techniques for content-based medical image retrieval. The author mainly reviews retrieval techniques based on features such as color, texture, shape, and spatial relationship. However, only one reference in this paper mentions that the shape of various nucleus segmented from original leukocyte images can be characterized by Fourier descriptors for efficient retrieval of chronic lymphocytic leukemia (CLL), follicular center cell lymphoma and mantle cell lymphoma (MCL).

The survey of\cite{Gurcan-2009-Histopathological} in 2009 reviews the latest developments in CAD technology for digital histopathology including histopathological image preprocessing, feature extraction and selection, image detection and segmentation. This paper reviews about 120 references, but only one summarizes in detail that combining statistical texture features with low-level color texture features can improve the classification performance of FL.

The survey of\cite{Belsare-2012-Histopathological} in 2012 reviews the segmentation and classification methods in histological images. More than 34 papers have been summarized, but only 4 papers are related to lymphoma.

The survey of\cite{He-2012-Histology} in 2012 summarizes the methods of histopathological images including pre-processing, segmentation, feature extraction and dimensionality reduction, and post-processing. However, the paper does not summarize the processing methods in LHIA, only one example in the paper is related to FL grading.

The survey of\cite{Kumar-2013-Content} in 2013 reviews the applications of content-based medical image retrieval technology in 2D images, multi-dimensional images and medical multi-modal images. But the images discussed in this paper are not the histopathological images that we are interested in, and only 1 reference is related to lymph nodes.

The survey of\cite{Irshad-2013-Methods} in 2013 reviews the development trends of various cell nucleus detection, segmentation, feature calculation and classification techniques for histopathological images in cancer detection and grading. However, the paper only shows lymphocyte nuclei in one picture. More than 100 papers are summarized in this paper, but only 2 papers related to lymphoma are summarized.

The survey of\cite{Arevalo-2014-Histopathology} in 2014 reviews the current techniques and applications for describing the visual content of histopathological images. An overview of the process of obtaining histopathological images is described, including biopsy, fixation, sectioning, cutting, staining, coverslipping and digitalization. This article also summarizes the steps of automatic histopathology image analysis including image preprocessing, feature extraction, and pattern recognition using machine learning methods. At the same time, this paper also reviews in detail four different feature extraction schemes and the applications of automatic histopathology image analysis in four different medical fields. But among the more than 70 references in the paper, only 5 are related to lymphoma.

The survey of\cite{Greenspan-2016-Guest} in 2016 summarizes the applications of deep learning technology in medical images. An overview of the applications of convolutional neural networks (CNNs) and other deep learning methods in medical tasks is presented. These tasks include lesion detection, image segmentation, shape modeling, and image registration. However, only 3 of the more than 30 references in the paper are related to lymph node detection and the images used are CT scan images.

The survey of\cite{Madabhushi-2016-image} in 2016 reviews the developments of computer image analysis tools used in predictive modeling of digital pathology images through detection, feature extraction, segmentation and tissue classification. This paper summarizes more than 20 references, only 2 are related to lymph.

The survey of\cite{Jothi-2017-Survey} in 2017 reviews various image preprocessing, segmentation, feature extraction, feature reduction, and classification techniques used in histopathological image analysis. More than 150 references are summarized, but only 11 are related to lymphoma.

The survey of\cite{Litjens-2017-Survey} in 2017 reviews the tasks of deep learning technology in medical image analysis in different fields. These tasks mainly include detection, segmentation, classification, retrieval, registration, image generation and enhancement. However, only 11 are related to lymphoma among more than 350 references in the paper.

The survey of\cite{Ker-2017-Deep} in 2017 reviews the ML algorithms used in medical image analysis, especially CNNs. The important research fields and applications of medical image segmentation, detection, localization, classification and registration are covered in the paper. However, only 1 is related to the detection of lymph node enlargement among more than 120 references in the paper, and the images used are CT slices that we are not interested in.

The survey of\cite{Robertson-2018-Digital} in 2018 reviews the applications of AI and deep learning technology in the diagnosis of breast pathology, as well as other recent developments in digital image analysis. However, only 8 are related to lymph among more than 180 references in the paper.

The survey of\cite{Bera-2019-Artificial} in 2019 reviews the applications of AI in digital pathology, including tumor diagnosis and prognosis, and hand-crafted feature-based ML methods. At the same time, this paper also summarizes the applications of several deep neural networks to the segmentation, detection, and classification of pathological images. These Deep neural network approaches include CNNs, fully convolutional networks (FCN), recurrent neural networks (RNN), generative adversarial networks (GAN). However, only 14 are related to lymph among more than 180 references in the paper. The survey of\cite{Levine-2019-Rise} in 2019 reviews the development and current status of deep learning in cancer diagnosis involving radiology and pathology. However, only 5 are related to lymph among about 90 references in the paper.

The survey of\cite{Zhou-2020-Comprehensive} in 2020 reviews the application of ANNs in breast histopathology images, including image classification, segmentation, and detection. However, only 3 are related to lymph among more than 150 references in the paper.

\begin{figure}[htbp!]
\centering
\centerline{\includegraphics[width=0.5\textwidth]{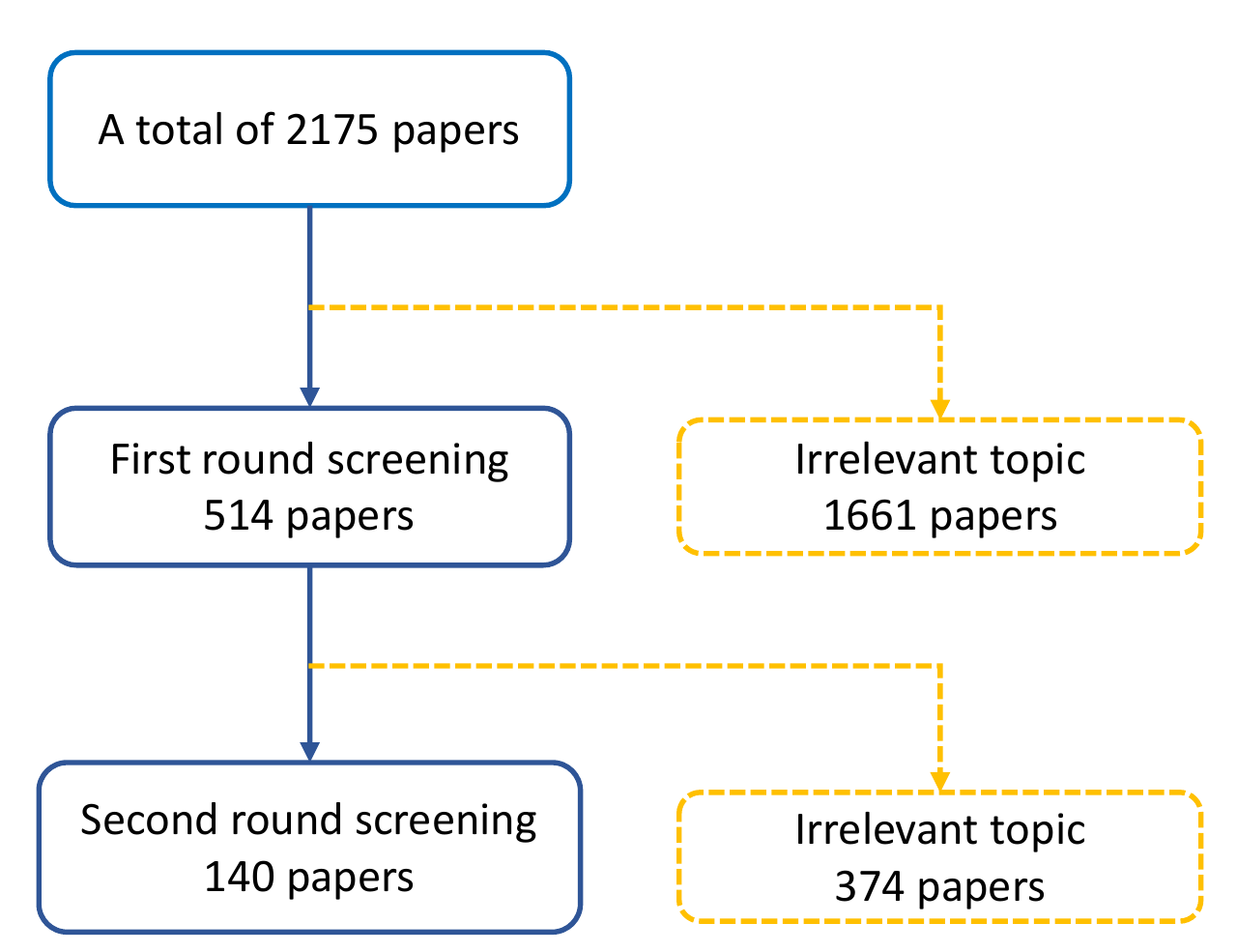}}
\caption{The flowchart of the paper searching and screening process.}
\label{fig:screening flowchart}
\end{figure}

According to the existing survey papers described above, we can find that none of the survey papers contributed to a comprehensive summary of the applications of MV in LHIA. Therefore, this study summarizes more than 170 works from 1999 to 2020 to fill the gap of MV in this field. The works summarized in this paper are scoured from arXiv, IEEE, Springer, Elsevier, and so on. The screening process of the whole work is shown in Fig.~\ref{fig:screening flowchart}. Twenty-six keywords are used to search for papers we are interested in, and a total of 2,175 papers are collected. After two rounds of screening, we retain 171 papers, including 135 papers on lymphoma histopathology image analysis and 36 survey papers.

\subsection{Structure of this Review}
\label{Sec:1.4}

The content introduced in each section of this paper is as follows: Sec.~\ref{Sec:2} introduces commonly used datasets and evaluation methods. Sec.\ref{Sec:3} introduces commonly used preprocessing methods, and ~Sec.\ref{Sec:4} introduces commonly used image segmentation methods. ~Sec.\ref{Sec:5}, ~Sec.\ref{Sec:6}, ~Sec.\ref{Sec:7} introduce the use of CAD technology for feature extraction, classification, and detection methods. ~Sec.\ref{Sec:8} analyses commonly used methods and potential methods, and finally ~Sec.\ref{Sec:9} summarizes this article and made further prospects for future work. 

\section{Datasets and Evaluation Methods}
\label{Sec:2}
In the applications of MV in analysing histopathological images of lymphoma, we have discussed some publicly available datasets that are frequently used. Besides, we listed evaluation metrics used in segmentation, classification, and detection tasks.

\subsection{Publicly Available Datasets}
\label{Sec:2.1}
Four publicly available datasets are discussed in this section; among them, three of them are Camelyon16~\cite{Camelyon16-2016}, Camelyon17~\cite{Camelyon16-2016}, IICUB-2008~\cite{Shamir-2008-Iicbu}. and the fourth dataset is collected from the National Cancer Institute (NCI) [86] and the National Institute on Aging (NIA)~\cite{NCI} and the National Institute on Aging (NIA)~\cite{NIA}. Table.~\ref{datasets} shows the basic information of the datasets.

\newcommand{\tabincell}[2]{\begin{tabular}{@{}#1@{}}#2\end{tabular}}
\begin{table}[!htbp]
\caption{available lymph histopathology image datasets.}
\centering
\begin{tabular}{ccccc}
\toprule
Datasets & Year & Staining & Task & Number of images\\
\midrule
Camelyon16 & 2016 & H$\&$E	& Detection & \tabincell{c}{270 images for training,\\ 130 images for testing}\\ \\

Camelyon17 & 2017 & H$\&$E & \tabincell{c}{Detection and\\ classification} & \tabincell{c}{500 images for training,\\500 images for testing}\\ \\

IICBU-2008 & 2008 & H$\&$E & Classification & 113 CLL, 139 FL, 122 MCL\\ \\

the NCI and NIA & $\backslash$ & H$\&$E & Classification & 12 CLL, 62 FL, 99 MCL\\
\bottomrule
\end{tabular}
\label{datasets}
\end{table}

\begin{figure}[htbp!]
\centering
\centerline{\includegraphics[width=0.6\textwidth]{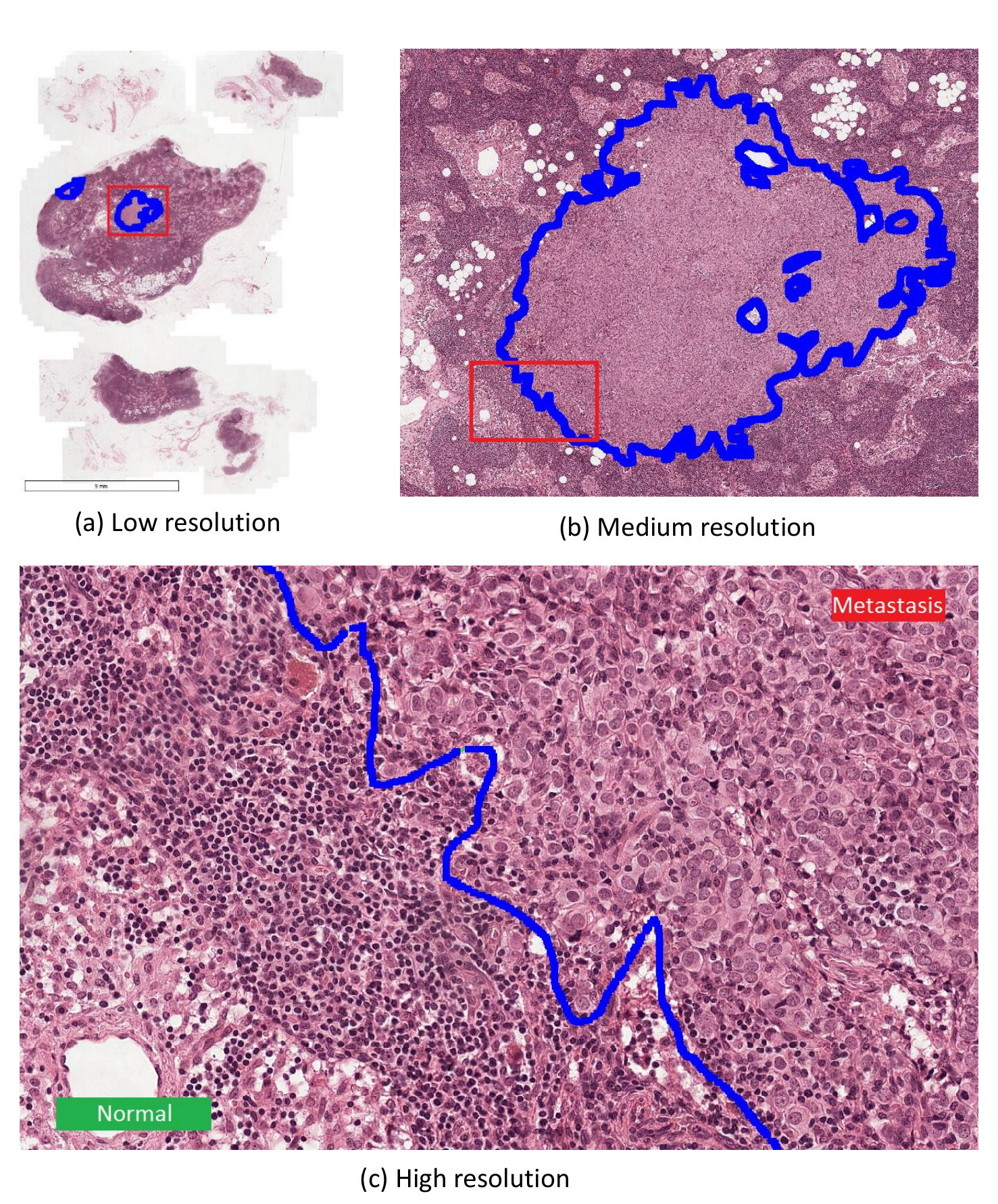}}
\caption{The WSIs in Camelyon16 dataset.}
\label{fig:Camelyon16}
\end{figure}

The Camelyon16 challenge is organized by International Symposium on Biomedical Imaging (ISBI) in 2016. The purpose of Camelyon16 is to advance the algorithms for automated detection of cancer metastases in H$\&$E stained lymph node images. This competition has a high clinical significance in reducing the workload of pathologists and the subjectivity in the diagnosis. The data used in Camelyon16 consists of 400 Whole-slide images (WSIs) of SLN, which are collected in Radboud University Medical Center (Radboudumc) (Nijmegen, the Netherlands), and the University Medical Center Utrecht (Utrecht, the Netherlands). The training dataset includes 270 WSIs, and the test dataset includes 130 WSIs. In Fig.~\ref{fig:Camelyon16}, (a), (b) and (c) are WSIs at low, medium and high resolution respectively~\cite{Camelyon16-2016}.

The Camelyon17 is the second challenge launched by the Diagnostic Image Analysis Group (DIAG) and Department of Pathology in Radboudumc. The purpose of Camelyon17 is to advance the algorithms for automated detection and classification of cancer metastases in H$\&$E stained lymph node images. Like Camelyon 16, Camalyon 17 also has a high clinical significance. The training dataset includes 500 WSIs, and the test dataset includes 500 WSIs (as shown in Fig.~\ref{fig:Camelyon16}).

\begin{figure}[htbp!]
\centering
\centerline{\includegraphics[width=1\textwidth]{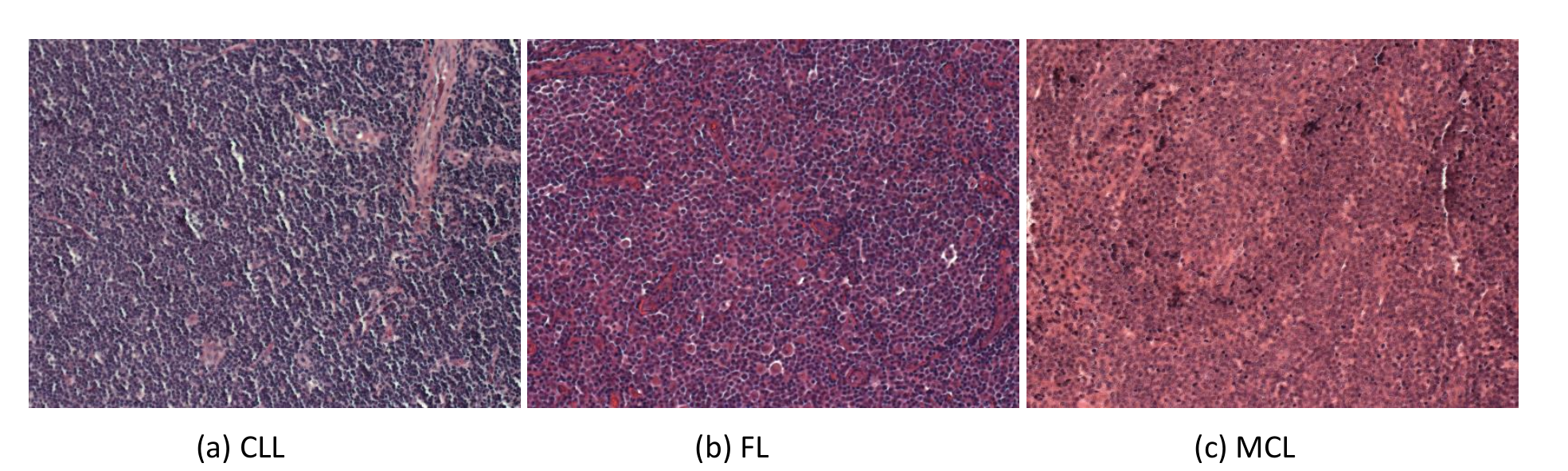}}
\caption{The histopathological images in IICBU-2008 dataset.}
\label{fig:IICBU-2008}
\end{figure}

The IICBU-2008 dataset is proposed by~\cite{Shamir-2008-Iicbu} in 2008 to provide free access of biological image datasets for computer experts, it includes nine small datasets containing different organs or cells. Three types of malignant lymphoma are proposed in the dataset containing lymphoma. There are 113 CLL images, 139 FL images and 122 MCL images (as shown in Fig.~\ref{fig:IICBU-2008}). The images are stained with H$\&$E and usually used for classification such as~\cite{Meng-2010-Histology}~\cite{Bai-2019-Nhl}.

\begin{figure}[htbp!]
\centering
\centerline{\includegraphics[width=1\textwidth]{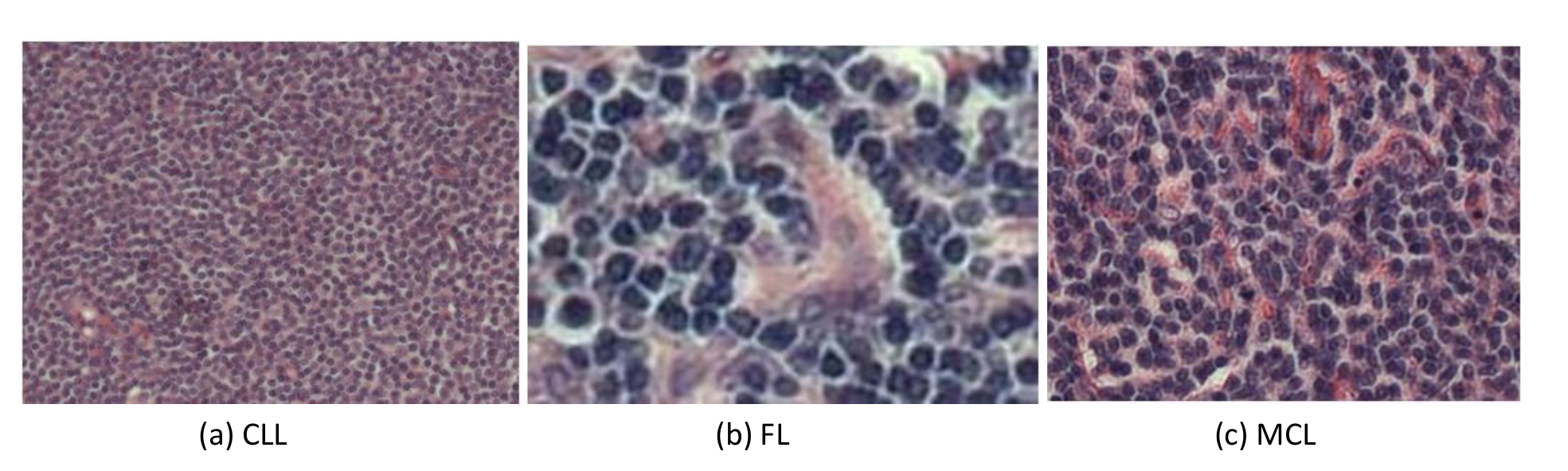}}
\caption{The histopathological images in the NCI and NIA dataset.}
\label{fig:NCI and NIA}
\end{figure}

The dataset used in~\cite{Roberto-2017-Features} and~\cite{Ribeiro-2018-Analysis} is from the NCI and NIA, contains 173 histological NHL images which are comprised of 12 CLL images, 62 FL images and 99 MCL images (as shown in Fig.\ref{fig:NCI and NIA}).

\subsection{Evaluation Method}
\label{Sec:2.2}
This section introduces some commonly used evaluation metrics in the classification, segmentation, and detection tasks of lymphoma histopathology images.

\subsubsection{Some Basic Metrics}
\label{Sec:2.2.1}

Here, we presented several commonly used evaluation metrics. Suppose we want to classify two types of samples, recorded as positive and negative, respectively: True positive (TP), which is the number of actually positive samples classified as positive by the classifier. False-positive (FP) is the number of negative samples classified as positive by the classifier. False-negative (FN) is the number of samples that are actually positive but classified as negative by the classifier. True negative (TN) is the number of negative samples classified as negative by the classifier. Table.~\ref{Confusion Matrix} shows the confusion matrix of the above four indexes. The value in the diagonal of the confusion matrix is the number of correct classifications for each category.

\begin{table}[!htbp]
\renewcommand\arraystretch{2}
\caption{Confusion Matrix. P represents positive, N represents negative.}
\label{Confusion Matrix}
\centering
\begin{tabular}{|c|c|c|c|}
\hline
 & \multicolumn{3}{c|}{Classification Result}\\
\hline
\multirow{3}*{\tabincell{c}{Actual \\Result}} &  & P & N\\
\cline{2-4}
& P & TP & FN\\
\cline{2-4}
& N & FP & TN\\
\hline
\end{tabular}
\end{table}

\subsubsection{Evaluation Criteria for Classification models}
\label{Sec:2.2.2}

In classification tasks, when a classifier is trained with enough training samples, the next step is to provide test samples to check whether the test samples are correctly classified. Then, a method to evaluate and quantify the results of this classification is necessary. Table.~\ref{classification metrics} shows the commonly used evaluation metrics in classification tasks.

Accuracy (ACC) is the most commonly used evaluation metric, which is the proportion of the number of correctly classified samples among all samples. A well-performing classifier generally has a high ACC. Precision indicates the proportion of samples classified as positive that are actually positive. Sensitivity (SEN) represents the proportion of all positive examples that are correctly classified and measures the classifier’s ability to recognize positive examples. Specificity (SPE) represents the proportion of all negative examples that are correctly classified and measures the classifier’s ability to recognize negative examples. SEN and SPE are also called true positive rates (TPR) and true negative rates (TNR). Recall measures how many positive samples are classified as positive. It can be seen that a Recall is equal to SEN in Table~\ref{classification metrics}. F1-score is often used as the final evaluation method in some multi-classification tasks. It is the harmonic average of precision and recall. In addition, Receiver Operating Characteristic Curve (ROC curve) can also be used to evaluate the quality of a classifier. TPR is used as the ordinate in the ROC curve, and one minus TNR is also called false positive rate (FPR) as the abscissa. The closer the ROC curve is to the upper left corner, the better the classifier performance. The classifier can also be evaluated by the area under the ROC curve (Area Under Curve, AUC). The larger the AUC, the more reliable the classifier model. In addition, some metrics such as positive predictive value (PPV) and negative predictive value (NPV) are also used in~\cite{Bollschweiler-2004-Artificial}. PPV represents the proportion of true positive samples among all samples classified as positive and is given as TP/(TP+FP). NPV represents the proportion of true negative samples among all samples classified as negative and is given as TN/(TN+FN).

\begin{table}[!htbp]
\renewcommand\arraystretch{2}
\caption{Evaluation metrics for classification tasks.}
\label{classification metrics}
\centering
\begin{tabular}{c|c}
\hline
Assessments & Formula\\
\hline
ACC & $\frac{TP+TN}{TP+TN+FP+FN}$\\

Precision & $\frac{TP}{TP+FP}$\\

SEN & $\frac{TP}{TP+FN}$\\

SPE & $\frac{TN}{TN+FP}$\\

Recall & $\frac{TP}{TP+FN}$\\

F1-score & $\frac{2TP}{2TP+FN+FP}$\\
\hline
\end{tabular}
\end{table}

\subsubsection{Evaluation Criteria for Segmentation Methods}
\label{Sec:2.2.3}

Segmentation can be used to detect the region of interest (ROI) in histopathological images. In tasks of histopathology image analysis, ROIs include tissue components such as lymphocytes and cell nuclei. Segmentation is the process to divide ROIs from the tissue. Since segmentation is generally used to prepare for subsequent detection and classification, so it is necessary to evaluate the segmentation result by appropriate metrics.

A segmentation algorithm is often measured by detection accuracy and segmentation accuracy. Detection accuracy is used to measure the ability of the segmentation algorithm in identifying ROIs, segmentation accuracy is used to measure the similarity between the ROIs obtained in the segmentation algorithm and the regions in the ground truth images~\cite{Jothi-2017-Survey}. The detection accuracy metrics commonly used in segmentation are discussed in Sec.~\ref{Sec:2.2.2} such as Precision, Recall, F-measure (namely F1-score). The segmentation accuracy metrics commonly used in segmentation include Dice coefficient (DICE), Hausdorff distance (HD), mean absolute distance (MAD)~\cite{Cheng-2010-Identifying}, Jaccard index (JAC)~\cite{Fatakdawala-2010-Expectation} and so on. The following are the formulas for the above metrics. $S_g$ is a set of points ($g1, g2, ..., gn$) constituting the automatically segmented contour, $S_t$ is a set of points ($t1, t2, ..., tn$) constituting the ground truth contour.

1) DICE represents the ratio of the area where the two regions intersect to the total area and is given by

\begin{equation}
{\rm DICE}=\frac{2 \vert S_g \cap S_t \vert}{\vert S_g \vert + \vert S_t \vert}
\label{eq:Dice}
\end{equation}

2) HD between $S_g$ and $S_t$ is the maximum among the minimum distance computed from each point of $S_g$ to each point of $S_t$. HD is defined by

\begin{equation}
{\rm HD}={\rm max}(h(S_g, S_t), h(S_t, S_g))
\label{eq:HD}
\end{equation}

\begin{equation}
h(S_g, S_t)=\mathop{{\rm max}}_{g \in S_g}\mathop{{\rm max}}_{t \in S_t}\parallel g-t \parallel
\end{equation}

3) MAD between $S_g$ and $S_t$ is the mean of the minimum distance computed from each point of $S_g$ to each point of $S_t$. MAD is defined by

\begin{equation}
{\rm MAD}=\frac{1}{n}{\rm \Sigma} _{g=1}^{n}\lbrace {\rm min}_{t \in S_t} \lbrace h(g,t)\rbrace\rbrace
\label{eq：MAD }
\end{equation}

4) JAC represents the intersection of two regions divided by their union and is defined by

\begin{equation}
{\rm JAC}=\frac{\vert S_g \cap S_t \vert}{\vert S_g \cup S_t \vert}
\label{eq:JAC}
\end{equation}

\subsubsection{Evaluation Criteria for Detection Methods}
\label{Sec:2.2.4}

Because most of the detection tasks are completed by classification, many metrics in the classification tasks can also be used to evaluate the results of the detection tasks. Such as ACC, Confusion matrix, Precision, Recall, ROC curve, AUC, SEN, SPE in Sec.~\ref{Sec:2.2.2}. In addition, free response operating characteristic (FROC) is used to evaluate the result in~\cite{Lin-2018-Scannet}, dice similarity coefficient (DSC) (as shown in Eq.~\ref{Eq:DSC}) is used to evaluate the result in~\cite{Senaras-2019-Segmentation} and AI (Artificial Intelligence) score is used to evaluate the probability of lymph node metastasis in~\cite{Harmon-2020-Multiresolution}.

\begin{equation}
{\rm DSC}=\frac{2{\rm TP}}{2{\rm TP}+{\rm FN}+{\rm FP}}
\label{Eq:DSC}
\end{equation}

\subsection{Summary}
\label{Sec:2.3}

In summary, we introduce the datasets commonly used in tasks of LHIA, among which the commonly used data sets are Camelyon16, Camelyon17 and IICBU-2008. Next, we summarize the evaluation metrics commonly used in classification, segmentation and detection tasks. We find that the commonly used metrics in classification and detection tasks are ACC, SEN, SPE, AUC. DICE and HD are the commonly used metrics in segmentation tasks.

\section{Image Preprocessing}
\label{Sec:3}

In order to obtain the expected results in tasks of LHIA, the images must be in good quality. Due to the difference in staining, the histopathological images obtained from the biopsy sample may show uneven staining and blurring cases~\cite{Jothi-2017-Survey}. In addition, the images can be simplified by removing irrelevant information and highlighting useful information for tasks such as feature extraction, segmentation and detection can be carried out smoothly. Therefore, the images need to be preprocessed by resizing, denoising, and normalization methods. In this section, we introduce the preprocessing methods commonly used in tasks of LHIA, which are based on color, filter, threshold, morphology, histogram and other methods. 

\subsection{Color-based Preprocessing Techniques}
\label{Sec:3.1}

At present, some CAD methods in tasks of LHIA convert images from RGB color space to other color spaces. Since the histopathological images show multiple colors, the RGB color space is not enough to describe these images. Therefore, the RGB images need to be converted to other color spaces to get better quantification and description of colors. In addition to the conversion of various color spaces, color-based preprocessing methods also include image grayscale, color deconvolution, color normalization, color equalization and enhancement, and extraction of mean value or maximum and minimum of R, G and B channel. The commonly used color-based preprocessing methods on LHIA as shown in Table.~\ref{Tab:color-based preprocessing}.

\begin{figure}[htbp!]
\centering
\centerline{\includegraphics[width=0.96\textwidth]{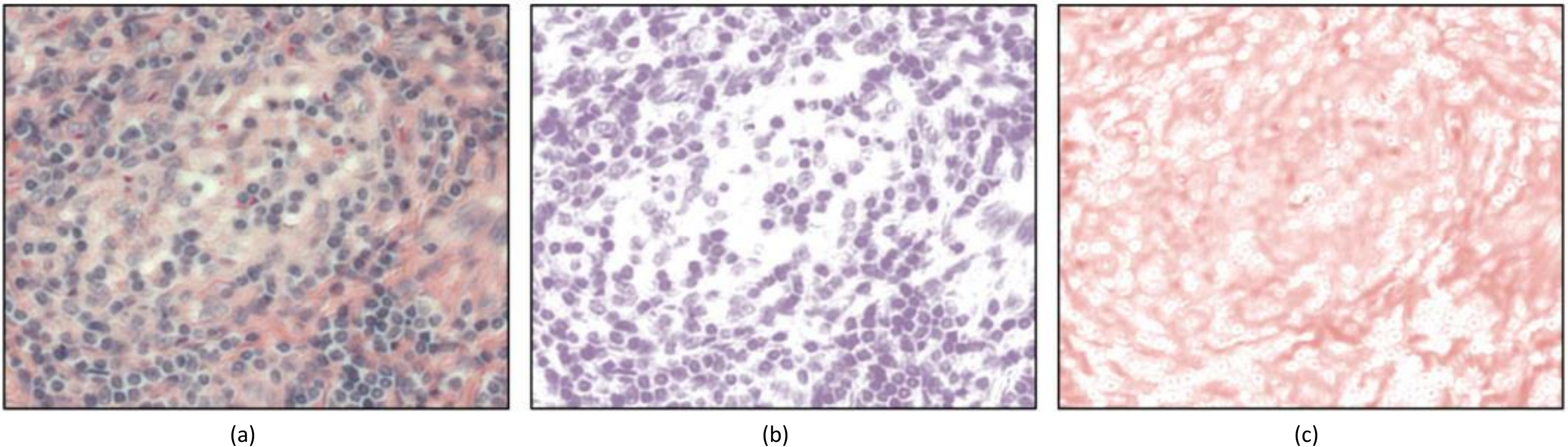}}
\caption{(a) MCL lymphoma sample stained with H$\&$E. (b) H channel. (c) E channel. This figure corresponds to Fig. 4 in~\cite{Orlov-2010-Automatic}.}
\label{fig:color decon}
\end{figure}

In~\cite{Angulo-2006-Ontology, Sertel-2008-Texture, Sertel-2009-Histopathological, Basavanhally-2009-Computerized, Belkacem-2009-Extraction, Belkacem-2010-Computer, Orlov-2010-Automatic, Belkacem-2010-Effect, Belkacem-2010-Segmentation, Akakin-2012-Content, Oztan-2012-Follicular, Saxena1-2013-Texture, Di-2015-Different, Tosta-2017-Computational, Zhu-2019-Novel, Bai-2019-Nhl, Martins-2019-Colour, Martins-2021-Hermite}, $L^*a^*b^*$ color space is used on LHIA for $L^*a^*b^*$ allowing color changes to be compatible with differences in visual perception. In~\cite{Chen-2005-Image, Sertel-2010-Image, Di-2015-Different, Tosta-2017-Computational}, $L^*u^*v^*$ color space is used on LHIA for $L^*u^*v^*$ color space are uniform in perception. In~\cite{Basavanhally-2008-Manifold, Sertel-2008-Computerized, Samsi-2010-Detection, Belkacem-2010-Effect, Han-2010-Multi, Akakin-2012-Content, Samsi-2012-Efficient, Ishikawa-2014-Gastric, Kuo-2014-Lymphatic, Zarella-2015-Lymph, Fauzi-2015-Classification, Di-2015-Different, Wang-2016-Deep, Chen-2016-Identifying, Tosta-2017-Computational, Zhu-2019-Novel}, HSV color space is used on LHIA for HSV separating the intensity from the color information. In~\cite{Belkacem-2010-Computer}, HSI color space is used on LHIA for HSI is hue based color space similar to HSV. In~\cite{Di-2015-Different, Tosta-2017-Computational}, Ycbcr color space is used on LHIA. In~\cite{Tosta-2017-Computational}, YIQ color space is used on LHIA. In~\cite{Zhu-2019-Novel}, YUV color space is used on LHIA. Ycbcr, YIQ and YUV are luminance based color spaces. In~\cite{Kong-2011-Partitioning}, a new color space is defined which is called MDC (the most discriminant color space). MDC is a linear combination of RGB color space, the texture features extracted from MDC can separate classes optimally. In~\cite{Orlov-2010-Automatic, Cheikh-2017-Spatial, Bianconi-2020-Experimental}, color deconvolution is used to separate H and E overlapped channels into individual channels. Fig.~\ref{fig:color decon} shows an example of application in color deconvolution. In~\cite{Samsi-2012-Efficient, Acar-2013-Tensor, Michail-2014-Detection, Dimitropoulos-2014-Using, Jiang-2018-Effective, Zhu-2019-Novel, Bandi-2019-Resolution}, RGB images are converted into grayscale images because the variations of color and intensity may hamper the classification performance, grayscale features need to be calculated, noise and small details in RGB images need to be removed and suppressed and the data in the individual channel is easier to process after image grayscale. In~\cite{Codella-2016-Lymphoma, Hashimoto-2020-Multi}, the color saturation of the original image has increased after color equalization and enhancement, color augmentation is used to reduce the effect of outlying colors. In~\cite{Tosta-2017-Computational, Ribeiro-2018-Analysis, Hashimoto-2020-Multi, Bianconi-2020-Experimental}, color normalization is used to improve the quality of images. In~\cite{Tosta-2017-Application, Tosta-2018-Fitness, Azevedo-2021-Evaluation}, the R channel is extracted from RGB color space because the R channel has the greatest contrast compared to the image background.

\begin{table}[!htbp]
\renewcommand\arraystretch{2}
\centering
\caption{Color-based Preprocessing Techniques.}
\label{Tab:color-based preprocessing}
\tiny
\begin{tabular}{cccc}
\hline
Reference & Year & Team & Details\\
\hline

\cite{Chen-2005-Image} & 2005 & W Chen & RGB to $L^*u^*v^*$\\

\cite{Angulo-2006-Ontology} & 2006 & J Angulo & RGB to $L^*a^*b^*$\\

\cite{Sertel-2008-Texture} & 2008 & O Sertel & RGB to $L^*a^*b^*$\\

\cite{Basavanhally-2008-Manifold} & 2008 & A Basavanhally & RGB to HSV\\

\cite{Sertel-2008-Computerized} & 2008 & O Sertel & RGB to HSV\\

\cite{Sertel-2009-Histopathological} & 2009 & O Sertel & RGB to $L^*a^*b^*$\\

\cite{Basavanhally-2009-Computerized} & 2009 & AN Basavanhally & RGB to $L^*a^*b^*$\\

\cite{Belkacem-2009-Extraction} & 2009 & K Belkacem-Boussaid & RGB to $L^*a^*b^*$\\

\cite{Belkacem-2010-Computer} & 2010 & K Belkacem-Boussaid & RGB to $L^*a^*b^*$, HSI\\

\cite{Sertel-2010-Image} & 2010 & O Sertel & RGB to $L^*u^*v^*$\\

\cite{Orlov-2010-Automatic} & 2010 & NV Orlov & \tabincell{c}{RGB to $L^*a^*b^*$, \\color deconvolution}\\

\cite{Sertel-2010-Computer} & 2010 & O Sertel & \tabincell{c}{RGB to a 1-D unitone image and further \\normalize the unitone image to the [0,1] range}\\

\cite{Samsi-2010-Detection} & 2010 & S Samsi & RGB to HSV\\

\cite{Belkacem-2010-Effect} & 2010 & K Belkacem-Boussaid & RGB to HSV, $L^*a^*b^*$\\

\cite{Han-2010-Multi} & 2010 & J Han & RGB to HSV\\

\cite{Belkacem-2010-Segmentation} & 2010 & K Belkacem-Boussaid & RGB to $L^*a^*b^*$\\

\cite{Kong-2011-Partitioning} & 2011 & H Kong & RGB to MDC\\

\cite{Akakin-2012-Content} & 2012 & HC Akakin & RGB to HSV, $L^*a^*b^*$\\

\cite{Samsi-2012-Efficient} & 2012 & S Samsi & RGB to grayscale, HSV\\

\cite{Oztan-2012-Follicular} & 2012 & B Oztan & RGB to $L^*a^*b^*$\\



\cite{Acar-2013-Tensor} & 2013 & E Acar & RGB to grayscale\\

\cite{Ishikawa-2014-Gastric} & 2014 & T Ishikawa & RGB to HSV\\

\cite{Michail-2014-Detection} & 2014 & E Michail & RGB to grayscale\\

\cite{Kuo-2014-Lymphatic} & 2014 & YL Kuo & RGB to HSV\\
 
\cite{Dimitropoulos-2014-Using} & 2014 & K Dimitropoulos & RGB to grayscale\\

\cite{Zarella-2015-Lymph} & 2015 & MD Zarella & RGB to HSV\\

\cite{Fauzi-2015-Classification} & 2015 & MFA Fauzi & RGB to HSV\\

\cite{Di-2015-Different} & 2015 & C Di Ruberto & RGB to HSV, $L^*a^*b^*$, $L^*u^*v^*$, Ycbcr\\

\cite{Chen-2016-Automatic} & 2016 & J Chen & Stain normalization\\

\cite{Wang-2016-Deep} & 2016 & D Wang & RGB to HSV\\

\cite{Chen-2016-Identifying} & 2016 & R Chen & RGB to HSV\\

\cite{Codella-2016-Lymphoma} & 2016 & N Codella & \tabincell{c}{Color equalization and enhancement, \\relative color enhancement}\\

\cite{Tosta-2017-Computational} & 2017 & TAA Tosta & \tabincell{c}{RGB to HSV, $L^*a^*b^*$, $L^*u^*v^*$, Ycbcr, YIQ.\\color normalization}\\

\cite{Tosta-2017-Application} & 2017 & TAA Tosta & Extract the R channel from RGB model\\

\cite{Cheikh-2017-Spatial} &2017 & BB Cheikh & Color deconvolution\\

\cite{Jiang-2018-Effective} & 2018 & H Jiang & RGB to grayscale\\

\cite{Ribeiro-2018-Analysis} & 2018 & MG Ribeiro & Color normalization\\

\cite{Tosta-2018-Fitness} & 2018 & TAA Tosta & Extract the R channel from RGB model\\



\cite{Zhu-2019-Novel} & 2019 & H Zhu & RGB to HSV, $L^*a^*b^*$, YUV, grayscale\\

\cite{Bai-2019-Nhl} & 2019 & J Bai & RGB to the blue ratio gray space, $L^*a^*b^*$\\

\cite{Martins-2019-Colour} & 2019 & AS Martins & RGB to $L^*a^*b^*$\\

\cite{Bandi-2019-Resolution} & 2019 & P Bándi & RGB to grayscale\\

\cite{Hashimoto-2020-Multi} & 2020 & N Hashimoto & Color normalization, color augmentation \\

\cite{Bianconi-2020-Experimental} & 2020 & F Bianconi & Color deconvolution, color normalization\\

\cite{Martins-2021-Hermite} & 2021 & AS Martins & RGB to $L^*a^*b^*$\\

\cite{Azevedo-2021-Evaluation} & 2021 & TA Azevedo Tosta & Extract the R channel from RGB model\\

\hline
\end{tabular}
\end{table}

\subsection{Filter-based Preprocessing Techniques}
\label{Sec:3.2}

Filtering can smooth images and remove uninteresting parts of the images such as noise, artifacts, and so on. The commonly used filter-based preprocessing methods are shown in Table.~\ref{Tab：Filter-based techniques}.

\begin{figure}[htbp!]
\centering
\centerline{\includegraphics[width=0.7\textwidth]{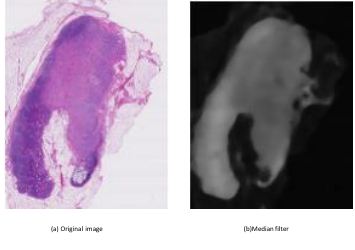}}
\caption{The median filter in preprocessing. (a) and (b) corresponds to Fig. 1 (a) and Fig. 1 (d) respectively in~\cite{Chen-2016-Identifying}.}
\label{fig:median filter}
\end{figure}

In~\cite{Samsi-2010-Detection, Oger-2012-General, Chen-2016-Identifying, Shi-2016-Automated, Wollmann-2017-Automatic, Wollmann-2018-Adversarial, Bandi-2019-Resolution}, a median filter is used to smooth texture variations, remove artifacts and noise, homogenize color intensity. An example using median filter is shown in Fig.~\ref{fig:median filter}. In~\cite{Han-2010-Multi, Schmitz-2012-Automated, Schafer-2013-Image, Michail-2014-Detection, Dimitropoulos-2014-Using, Shi-2016-Automated, Es-2017-Decision, Tosta-2017-Computational, Tosta-2017-Application, Tosta-2018-Fitness, Azevedo-2021-Evaluation}, Gaussian filter is used to remove noise and small details, convolve input images, smooth images, homogenize color intensity, adjust contrast. In~\cite{Belkacem-2010-Segmentation}, a matching filter is used for removing noise, flattening background and enhancing the contours of the follicle region to homogenize images. In~\cite{Codella-2016-Lymphoma}, region filter is used to remove areas with a surface area that is less than 10 pixels. In~\cite{Do-2018-Lymphoma}, the 2D order-statistics filter is used to remove noise and discrete points.

\begin{table}[!htbp]
\renewcommand\arraystretch{2}
\centering
\caption{Filter-based Preprocessing Techniques.}
\label{Tab：Filter-based techniques}
\tiny
\begin{tabular}{cccc}
\hline
Reference & Year & Team & Details\\
\hline

\cite{Samsi-2010-Detection} & 2010 & S Samsi & median filter\\

\cite{Han-2010-Multi} & 2010 & J Han & Gaussian filters\\

\cite{Belkacem-2010-Segmentation} & 2010 & K Belkacem-Boussaid & matching filter\\

\cite{Oger-2012-General} & 2012 & M Oger & median filter\\

\cite{Schmitz-2012-Automated} & 2012 & A Schmitz & Gaussian filtering\\

\cite{Schafer-2013-Image} & 2013 & T Schäfer & Gaussian filter\\

\cite{Michail-2014-Detection} & 2014 & E Michail & Gaussian filtering\\

\cite{Dimitropoulos-2014-Using} & 2014 & K Dimitropoulos & Gaussian filtering\\

\cite{Chen-2016-Identifying} & 2016 & R Chen & median filter\\

\cite{Shi-2016-Automated} & 2016 & P Shi & median filter, Gaussian filter\\

\cite{Codella-2016-Lymphoma} & 2016 & N Codella & region filter\\

\cite{Litjens-2016-Deep} & 2016 & G Litjens & for separating background from tissue\\

\cite{Wollmann-2017-Automatic} & 2017 & T. Wollmann & median filter\\

\cite{Es-2017-Decision} & 2017 & A ES Negm & Gaussian filter\\

\cite{Tosta-2017-Computational} & 2017 & TAA Tosta & Gaussian filter\\

\cite{Tosta-2017-Application} & 2017 & TAA Tosta & Gaussian filter\\

\cite{Do-2018-Lymphoma} & 2018 & MZ do Nascimento & 2D order-statistics filter\\

\cite{Wollmann-2018-Adversarial} & 2018 & T Wollmann & median filter\\

\cite{Tosta-2018-Fitness} & 2018 & TAA Tosta & Gaussian filter\\

\cite{Bandi-2019-Resolution} & 2019 & P Bándi & median filter\\

\cite{Azevedo-2021-Evaluation} & 2021 & TA Azevedo Tosta & Gaussian filter\\
\hline
\end{tabular}
\end{table}

\subsection{Threshold-based Preprocessing Techniques}
\label{Sec:3.3}

Thresholding can separate the area of interest in an image from the background. In RGB images, the brightnesses of R, G and B commonly are used to be a threshold. Table.~\ref{Tab:Threshold-based techniques} shows the references using threshold-based preprocessing techniques.

\begin{figure}[htbp!]
\centering
\centerline{\includegraphics[width=0.9\textwidth]{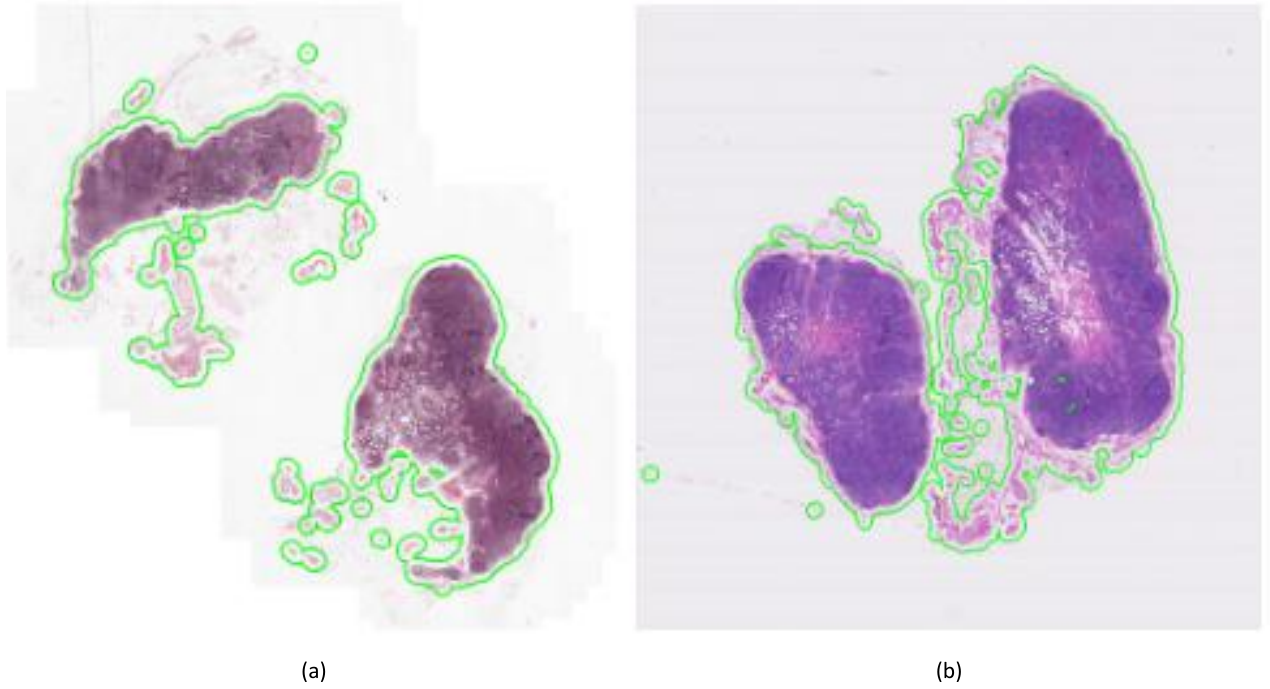}}
\caption{Visualization of tissue area detection during preprocessing. The figure corresponds to Figure 1 in~\cite{Wang-2016-Deep}.}
\label{fig:Otsu}
\end{figure}

In~\cite{Zorman-2007-Symbol, Zorman-2011-Classification}, the mean brightness of R component is selected to be a threshold of the binary image. The pixels whose values are less than the threshold are set to 1 and the rest is set to 0. In~\cite{Belkacem-2010-Computer, Wang-2016-Deep, Chen-2016-Identifying, Codella-2016-Lymphoma, Xiao-2017-Deep, Li-2018-Cancer, Lin-2018-Scannet, Bandi-2019-Resolution}, the Otsu threshold method is used to calculate the optimal threshold of each channel in color space, binarize images, segment color channels, remove the background regions of images. The detection result is shown in Fig.~\ref{fig:Otsu} after the preprocessing including conversion of color space and Otsu thresholding, the tissue regions are surrounded by a green curve. In~\cite{Schmitz-2012-Automated, Schafer-2013-Image, Ishikawa-2014-Gastric, Fauzi-2015-Classification, Es-2017-Decision, Bentaieb-2018-Predicting}, a threshold is selected to separate the background area and tissue area to highlight the tissue area. In~\cite{Michail-2014-Morphological}, the intensity threshold is used to remove non-candidate regions for nucleoli are dark and non-candidate regions are bright.

\begin{table}[!htbp]
\renewcommand\arraystretch{2}
\centering
\caption{Threshold-based Preprocessing Techniques.}
\label{Tab:Threshold-based techniques}
\tiny
\begin{tabular}{cccc}
\hline
Reference & Year & Team & Details\\
\hline
\cite{Zorman-2007-Symbol} & 2007 & M Zorman & \tabincell{c}{The mean brightness value of the R \\component of the RGB image is used as the threshold}\\

\cite{Belkacem-2010-Computer} & 2010 & K Belkacem-Boussaid & Otsu\\

\cite{Zorman-2011-Classification} & 2011 & M Zorman & \tabincell{c}{The mean brightness value of the R \\component of the RGB image is used as the threshold}\\

\cite{Schmitz-2012-Automated} & 2012 & A Schmitz & Background thresholding\\

\cite{Schafer-2013-Image} & 2013 & T Schäfer & a Threshold for background elimination\\

\cite{Ishikawa-2014-Gastric} & 2014 & T Ishikawa & \tabincell{c}{Extracting background area by \\binarization with threshold 30}\\

\cite{Michail-2014-Morphological} & 2014 & E Michail & Intensity thresholding\\

\cite{Kuo-2014-Lymphatic} & 2014 & YL Kuo & The threshold is used to create the binary image\\

\cite{Dimitropoulos-2014-Using} & 2014 & K Dimitropoulos & \tabincell{c}{A threshold (empirically set to 0.37) on \\the normalized R component of the RGB image}\\

\cite{Fauzi-2015-Classification} & 2015 & MFA Fauzi & A local thresholding operation\\

\cite{Wang-2016-Deep} & 2016 & D Wang & \tabincell{c}{Otsu is used to computed the optimal \\threshold values in H, S, V channels}\\

\cite{Chen-2016-Identifying} & 2016 & R Chen & \tabincell{c}{Otsu's binarization is \\performed on the S channel}\\

\cite{Codella-2016-Lymphoma} & 2016 & N Codella & Otsu thresholding technique\\

\cite{Xiao-2017-Deep} & 2017 & K Xiao & Otsu is used to binarize the image\\

\cite{Wollmann-2017-Automatic} & 2017 & T. Wollmann & Color thresholding\\

\cite{Es-2017-Decision} & 2017 & A ES Negm & A threshold for background elimination\\

\cite{Cheikh-2017-Spatial} &2017 & BB Cheikh & \tabincell{c}{Extracting nuclear objects \\by image thresholding}\\

\cite{Li-2018-Cancer} & 2018 & Y Li & Otsu is used to exclude the background\\

\cite{Do-2018-Lymphoma} & 2018 & MZ do Nascimento & \tabincell{c}{Contrast limited adaptive \\histogram equalisation (CLAHE)}\\

\cite{Wollmann-2018-Adversarial} & 2018 & T Wollmann & Color thresholding\\

\cite{Lin-2018-Scannet} & 2018 & H Lin & Otsu is used to determine the adaptive threshold\\

\cite{Bentaieb-2018-Predicting} & 2018 & A BenTaieb & \tabincell{c}{Removing background pixels using a \\threshold on the pixel intensity values}\\

\cite{Bandi-2019-Resolution} & 2019 & P Bándi & Otsu’s adaptive threshold\\
\hline
\end{tabular}
\end{table}

\subsection{Morphology-based Preprocessing Techniques}
\label{Sec:3.4}

Morphological operations can smooth the image contour, suppress the noise, fill in small holes and connect broken areas. There are four basic operations, namely dilation, erosion, opening operation and closing operation. Table.~\ref{Tab:Morphology-based techniques} shows the references using morphology-based preprocessing techniques.

\begin{figure}[htbp!]
\centering
\centerline{\includegraphics[width=0.7\textwidth]{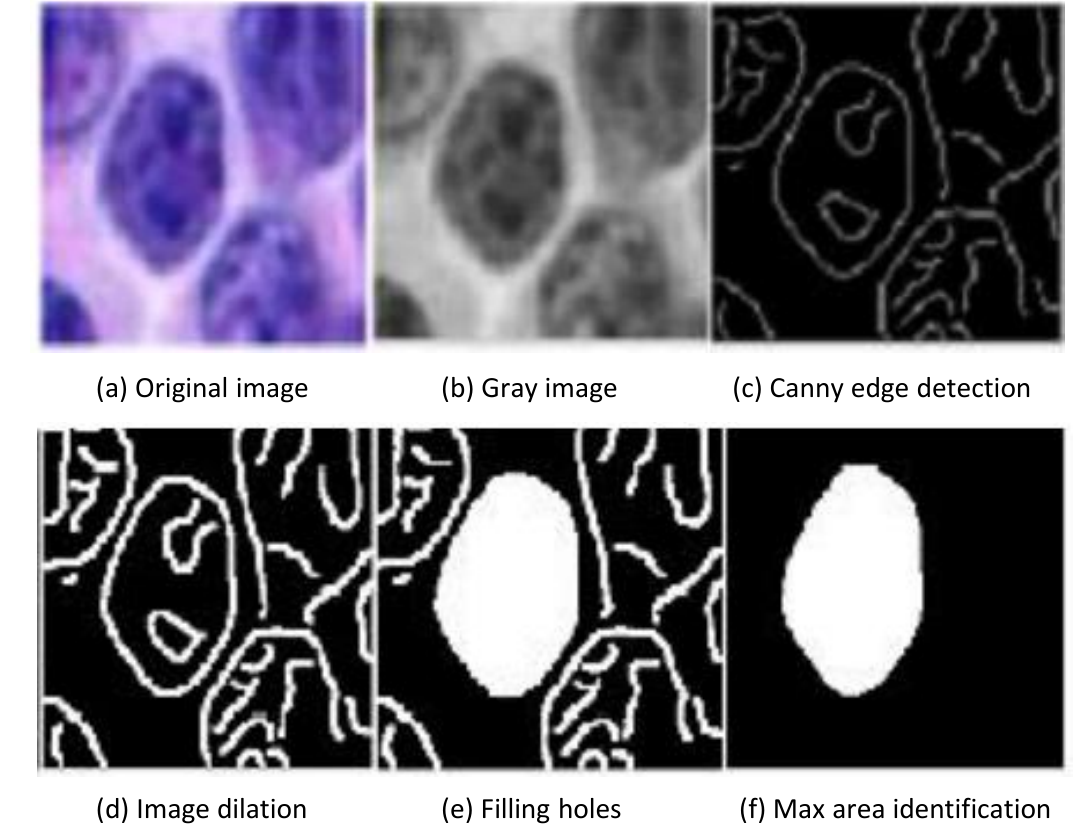}}
\caption{The results of preprocessing. The figure corresponds to Fig. 5 in~\cite{Sandhya-2013-Automated}.}
\label{fig:canny detector}
\end{figure}

In~\cite{Belkacem-2010-Computer, Cheng-2010-Identifying, Sandhya-2013-Automated, Bandi-2019-Resolution}, opening and closing operations are used to recover the shape of tissue, fill the holes in the segmented binary images. In~\cite{Sandhya-2013-Automated, Michail-2014-Morphological, Codella-2016-Lymphoma, Xiao-2017-Deep}, dilation and erosion operators are used to connect edges, remove noise and black holes. In~\cite{Sandhya-2013-Automated}, the preprocessing process includes removing noise, smoothing images by using disc smooth operator, making edges continuous by using canny edge detector, opening and closing operations, dilation operator, filling connected components using hole filling method, normalizing the maximum area of the connected area in images. Fig.~\ref{fig:canny detector} shows the results of preprocessing.

\begin{table}[!htbp]
\renewcommand\arraystretch{2}
\centering
\caption{Morphology-based Preprocessing Techniques.}
\label{Tab:Morphology-based techniques}
\tiny
\begin{tabular}{cccc}
\hline
Reference & Year & Team & Details\\
\hline

\cite{Belkacem-2010-Computer} & 2010 & K Belkacem-Boussaid & Opening and closing\\

\cite{Cheng-2010-Identifying} & 2010 & J Cheng & Hole-filling algorithm, opening\\

\cite{Neuman-2010-Segmentation} & 2010 & U Neuman & Removing small spots\\

\cite{Belkacem-2010-Segmentation} & 2010 & K Belkacem-Boussaid & \tabincell{c}{Eliminating undesired areas \\and filling holes}\\

\cite{Sandhya-2013-Automated} & 2013 & BS Sandhya & \tabincell{c}{Disc smooth operator, opening, \\closing, dilation,filling holes}\\

\cite{Michail-2014-Morphological} & 2014 & E Michail & Dilation\\

\cite{Kuo-2014-Lymphatic} & 2014 & YL Kuo & Hasan hole filling algorithm\\

\cite{Codella-2016-Lymphoma} & 2016 & N Codella & Disk dilation of 2 pixels\\

\cite{Xiao-2017-Deep} & 2017 & K Xiao & Erosion, dilation\\

\cite{Cheikh-2017-Spatial} &2017 & BB Cheikh. & Removing small object\\

\cite{Bandi-2019-Resolution} & 2019 & P Bándi & Closing, opening, hole filling\\

\hline
\end{tabular}
\end{table}

\subsection{Histogram-based Preprocessing Techniques}
\label{Sec:3.5}

The histogram shows the total number of pixels in each gray level in the images. Histogram equalization, histogram normalization and histogram stretching are commonly used to enhance image contrast. Table.~\ref{Tab:Histogram-based techniques} shows the references using histogram-based preprocessing techniques.

\begin{figure}[htbp!]
\centering
\centerline{\includegraphics[width=0.97\textwidth]{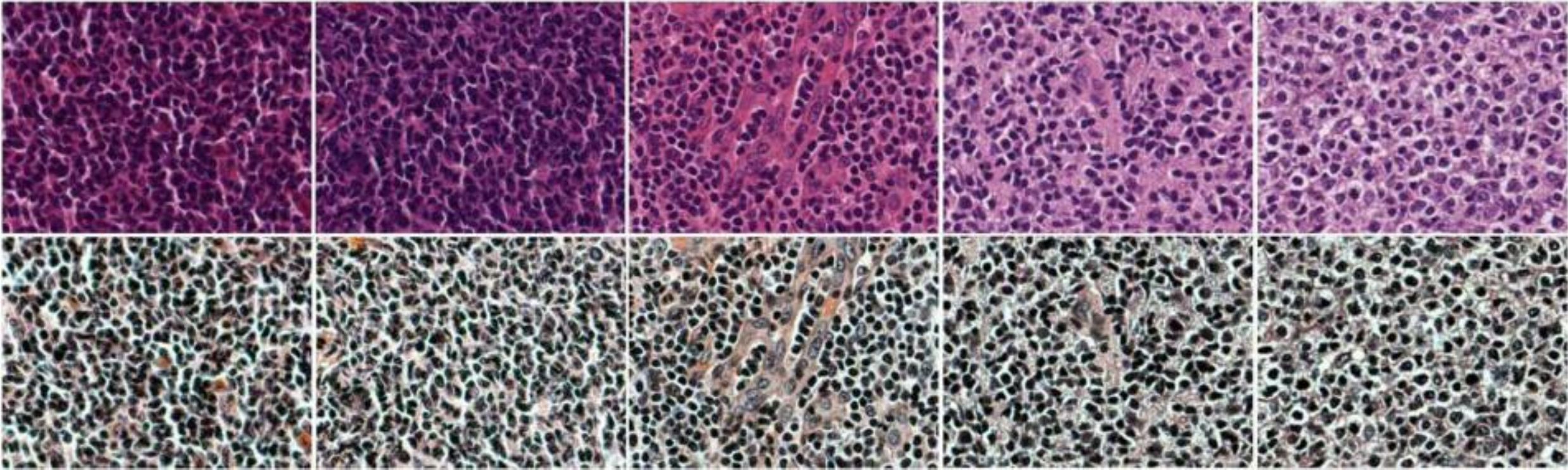}}
\caption{First row: the original images of FL. Second row: the images after histogram equalization. The figure corresponds to Fig. 1 in~\cite{Kong-2011-Partitioning}.}
\label{fig:histogram equ}
\end{figure}

In~\cite{Samsi-2010-Detection, Kong-2011-Partitioning, Oger-2012-General, Michail-2014-Detection, Dimitropoulos-2014-Using, Codella-2016-Lymphoma, Tosta-2017-Computational, Tosta-2017-Application, Tosta-2018-Fitness, Somaratne-2019-Improving, Bianconi-2020-Experimental, Azevedo-2021-Evaluation}, histogram equalization is used to increase the contrast of images, better identify the tissue regions, enhance the differences between background and tissue, normalize the color distribution of slides under different staining and lighting conditions to obtain a uniformly distributed histogram. As shown in the first row of Fig.~\ref{fig:histogram equ}, the histopathological images of FL show inconsistent colors caused by inconsistent lighting conditions, the second row shows more consistent images after histogram equalization. In~\cite{Akakin-2012-Content}, the co-occurrence histogram is normalized to extract texture features. In~\cite{Cheikh-2017-Spatial}, histogram stretching is used to solve the problem of changes in hematoxylin concentration after color deconvolution.

\begin{table}[!htbp]
\renewcommand\arraystretch{2}
\centering
\caption{Histogram-based Preprocessing Techniques.}
\label{Tab:Histogram-based techniques}
\tiny
\begin{tabular}{cccc}
\hline
Reference & Year & Team & Details\\
\hline

\cite{Samsi-2010-Detection} & 2010 & S Samsi & Histogram equalization\\

\cite{Kong-2011-Partitioning} & 2011 & H Kong & Histogram equalization\\

\cite{Oger-2012-General} & 2012 & M Oger & Adaptive equalization of histogram\\

\cite{Akakin-2012-Content} & 2012 & HC Akakin & Normalizing co-occurrence histograms\\

\cite{Michail-2014-Detection} & 2014 & E Michail & Histogram equalization\\


\cite{Dimitropoulos-2014-Using} & 2014 & K Dimitropoulos & Histogram equalization\\

\cite{Codella-2016-Lymphoma} & 2016 & N Codella & Histogram equalization\\

\cite{Tosta-2017-Computational} & 2017 & TAA Tosta & Histogram equalization\\

\cite{Tosta-2017-Application} & 2017 & TAA Tosta & Histogram equalization\\

\cite{Cheikh-2017-Spatial} &2017 & BB Cheikh & Histogram stretching\\

\cite{Tosta-2018-Fitness} & 2018 & TAA Tosta & Histogram equalization\\

\cite{Somaratne-2019-Improving} & 2019 & UV Somaratne & Histogram equalization\\

\cite{Bianconi-2020-Experimental} & 2020 & F Bianconi & Histogram equalization\\

\cite{Azevedo-2021-Evaluation} & 2021 & TA Azevedo Tosta & Histogram equalization\\
\hline
\end{tabular}
\end{table}

\subsection{Other Preprocessing Techniques}
\label{Sec:3.6}

In addition to the above-mentioned preprocessing methods, some papers design different image preprocessing methods in LHIA.

\begin{figure}[!htbp]
\centering
\centerline{\includegraphics[width=0.98\textwidth]{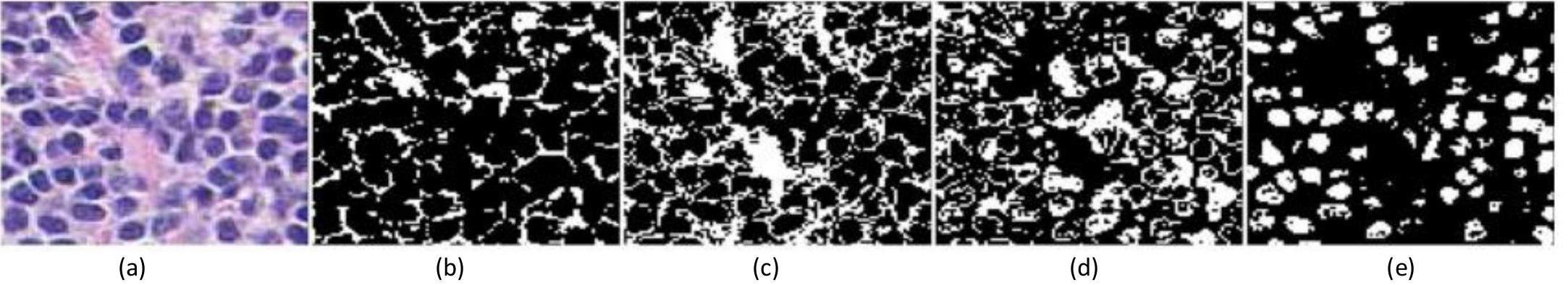}}
\caption{(a) Original HER2+ BC histopathology image with corresponding (b)-(e) class binarized scenes. The figure corresponds to Fig. 3 in~\cite{Fatakdawala-2010-Expectation}.}
\label{fig:EM}
\end{figure}

In~\cite{Fatakdawala-2010-Expectation}, a method is used to segment and detect lymphocytes in HER2+ breast cancer (BC) histopathology images. In this scheme, an EM-based (expectation Maximization) segmentation provides a specific initialization for enhancing the performance of active contours. Fig.~\ref{fig:EM} shows the results obtained by EM. 

\begin{figure}[!htbp]
\centering
\centerline{\includegraphics[width=0.7\textwidth]{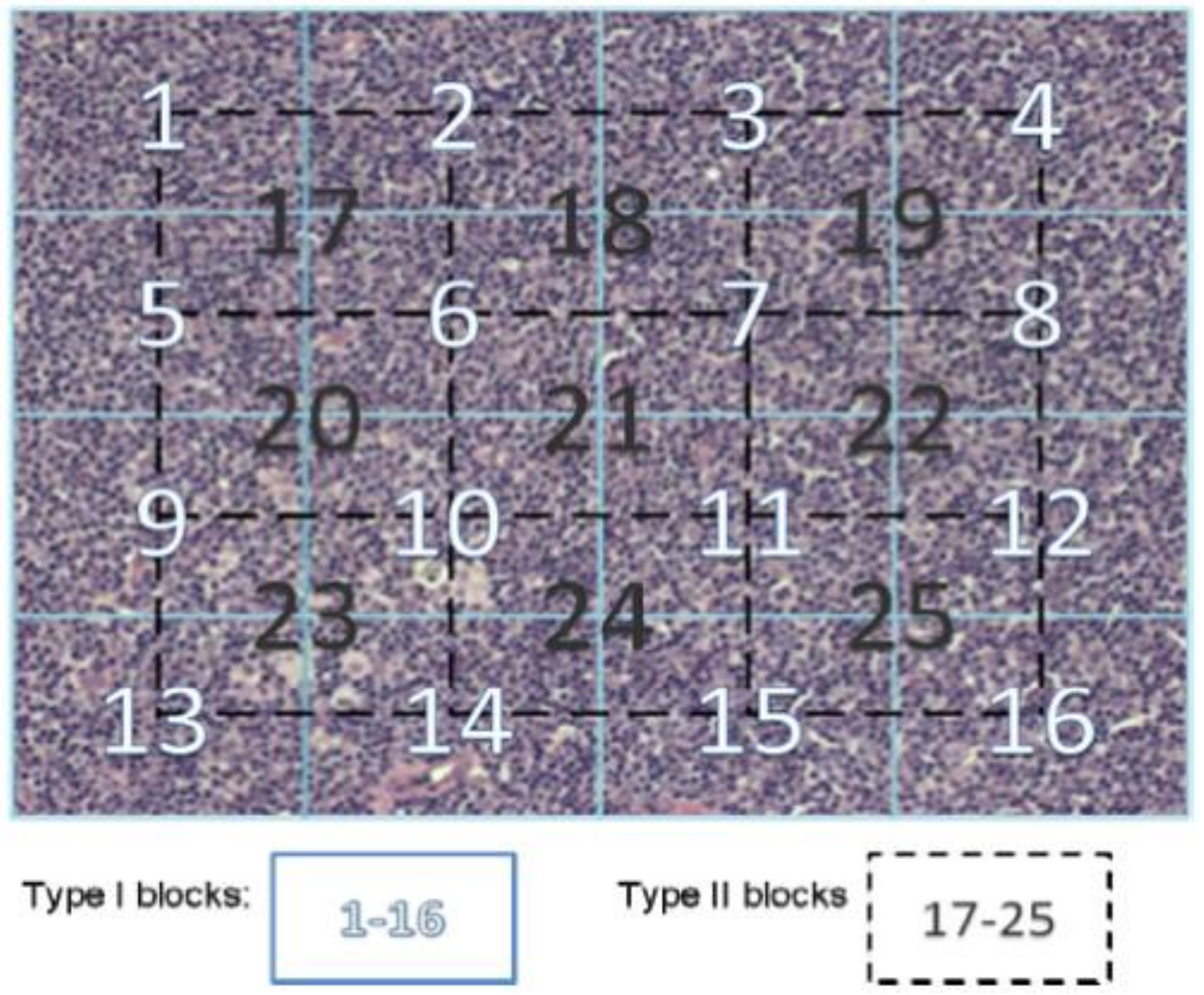}}
\caption{Division of the image to 25 blocks. This figure corresponds to Figure 2 in~\cite{Meng-2010-Histology}.}
\label{fig:PATCH}
\end{figure}

\begin{figure}[!htbp]
\centering
\centerline{\includegraphics[width=0.7\textwidth]{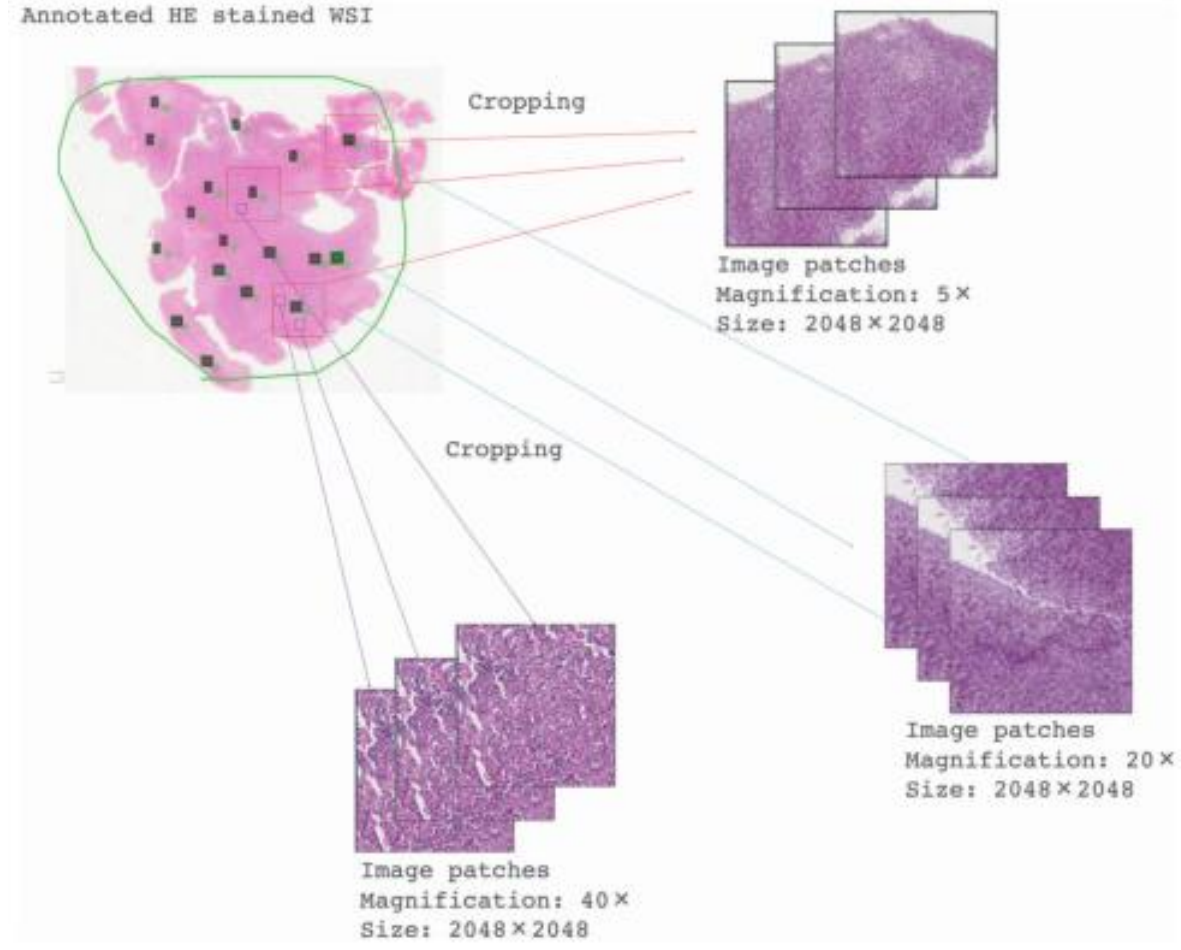}}
\caption{Procedure for creating image patches. The figure corresponds to Fig. 1 in~\cite{Miyoshi-2020-Deep}.}
\label{fig:PATCH2}
\end{figure}

\begin{figure}[!htbp]
\centering
\centerline{\includegraphics[width=0.7\textwidth]{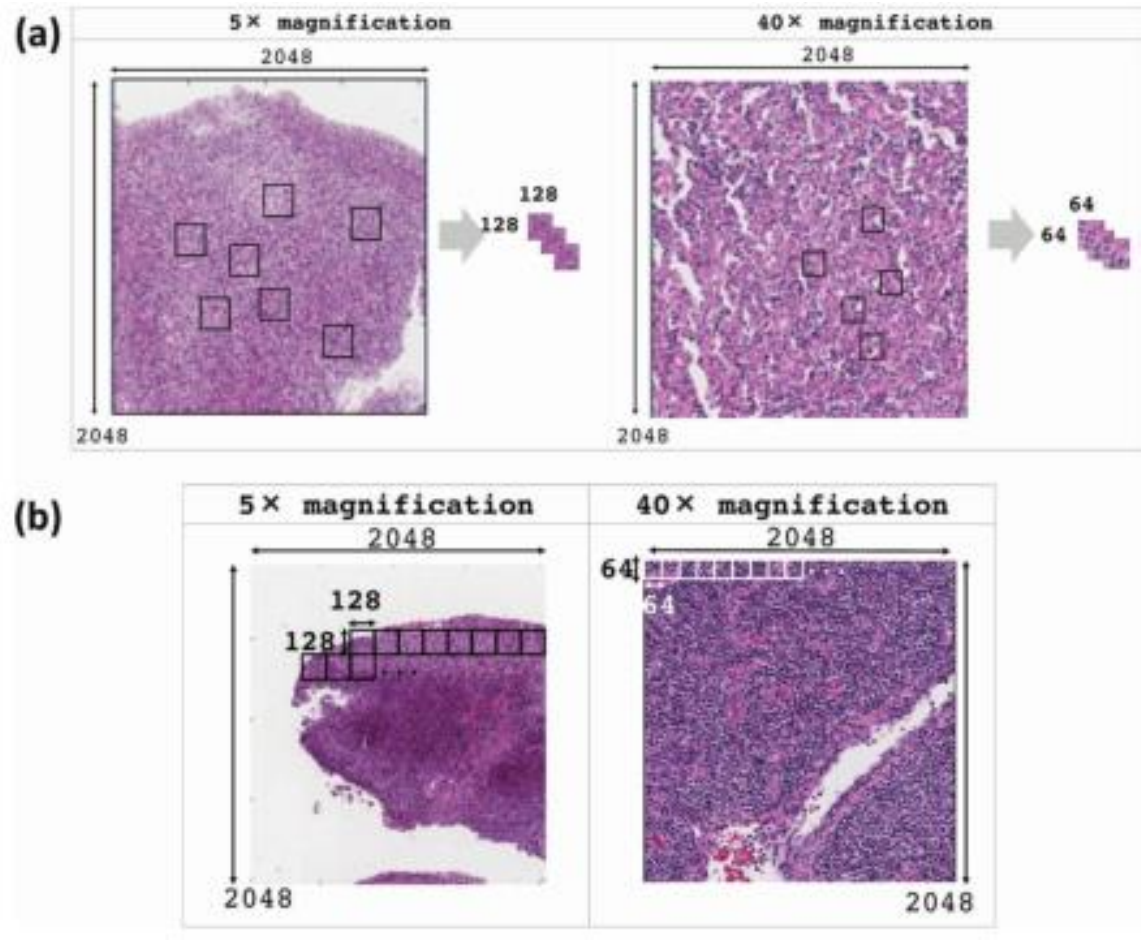}}
\caption{Preprocessing during the training and testing phase. The figure corresponds to Fig. 2 in~\cite{Miyoshi-2020-Deep}.}
\label{fig:PATCH3}
\end{figure}

In~\cite{Meng-2010-Histology} and~\cite{Meng-2013-Multimodal}, a framework is designed for histology images classification. In the feature preparation component of the framework, each image is divided into 25 blocks that are equal-sized (as shown in Fig.~\ref{fig:PATCH}). The purposes of the step include decreasing space complexity for feature extraction, the local features extracted by blocks can improve the results than the global features extracted without blocks and different features useful for classification in different regions of the image can be extracted. In~\cite{Han-2010-Multi, Ishikawa-2014-Gastric, Xiao-2017-Deep, Jamaluddin-2017-Tumor, Jiang-2018-Effective, Bentaieb-2018-Predicting, Alom-2019-Advanced, Senaras-2019-Segmentation, Mohlman-2020-Improving, Syrykh-2020-Accurate, Miyoshi-2020-Deep, Dif-2020-Efficient, Hashimoto-2020-Multi}, the original images are cropped into smaller patches. In~\cite{Jamaluddin-2017-Tumor}, 40,000 patches are extracted from the training set because the process of eliminating black borders, noise and white background area would increase the unnecessary time. In~\cite{Bentaieb-2018-Predicting}, all sides around tissue are cropped for reducing the computational amount. In~\cite{Miyoshi-2020-Deep}, 2048 pixel $\times$ 2048 pixel sized patches are cropped in the center of each annotation of WSI at a magnification of $\times$5, and 2048 pixel $\times$ 2848 pixel sized patches are cropped in the periphery of each annotation of WSI at a magnification of $\times$20 and$\times$40 (as shown in Fig.~\ref{fig:PATCH2}). Next, 128 pixel $\times$ 128 pixel sized patches are cropped from 2048 pixel $\times$ 2018 pixel sized patches at a magnification of $\times$5 during the training phase and test phase, and 64 pixel $\times$ 64 pixel sized patches are cropped from 2048 pixel $\times$ 2018 pixel sized patches at magnifications of $\times$20 and $\times$40 during the training phase and test phase (as shown in Fig.~\ref{fig:PATCH3}).

\begin{figure}[!htbp]
\centering
\centerline{\includegraphics[width=0.7\textwidth]{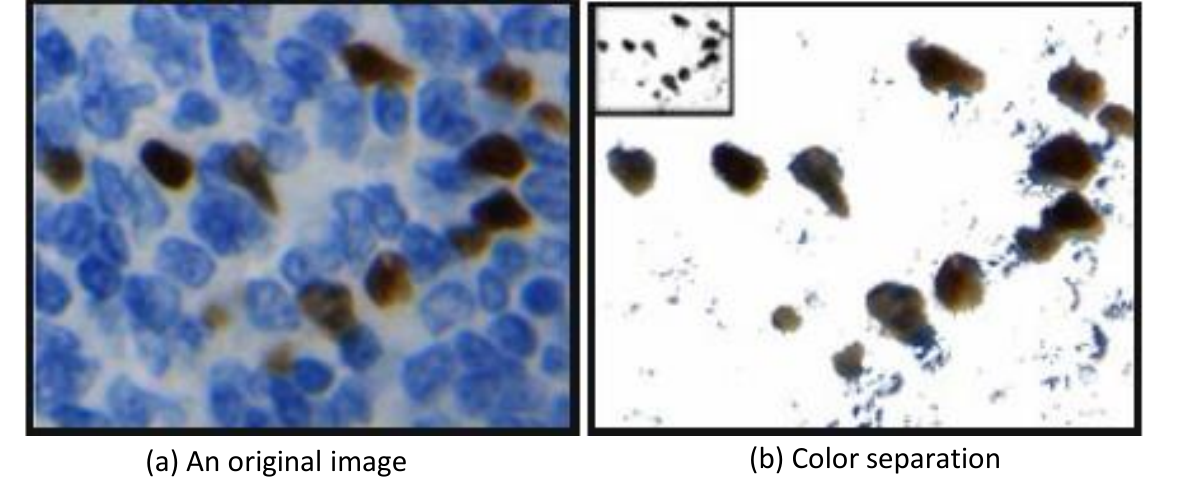}}
\caption{The preprocessing steps. The figure corresponds to Fig.3 in~\cite{Neuman-2010-Segmentation}.}
\label{fig:Rescale}
\end{figure}

In~\cite{Neuman-2010-Segmentation}, a method is designed for segmenting images of IHC stained FL tissue sections. Color separation and rescaling are used in the step of preprocessing. First, the brown in an image is separated by replacing all pixels outside the defined range of brown with white pixels (as shown in Fig.~\ref{fig:Rescale} (b)). Second, the Gaussian pyramid is used for rescaling the image (as shown in Fig.~\ref{fig:Rescale} (b), upper left, black box).

Since the contours of the images may be irregular, the Fourier shape descriptors are used to smooth and underline better the contours in~\cite{Belkacem-2010-Segmentation,Kong-2011-Partitioning,Kong-2011-Splitting,Samsi-2012-Efficient}.

\begin{figure}[htbp!]
\centering
\centerline{\includegraphics[width=0.7\textwidth]{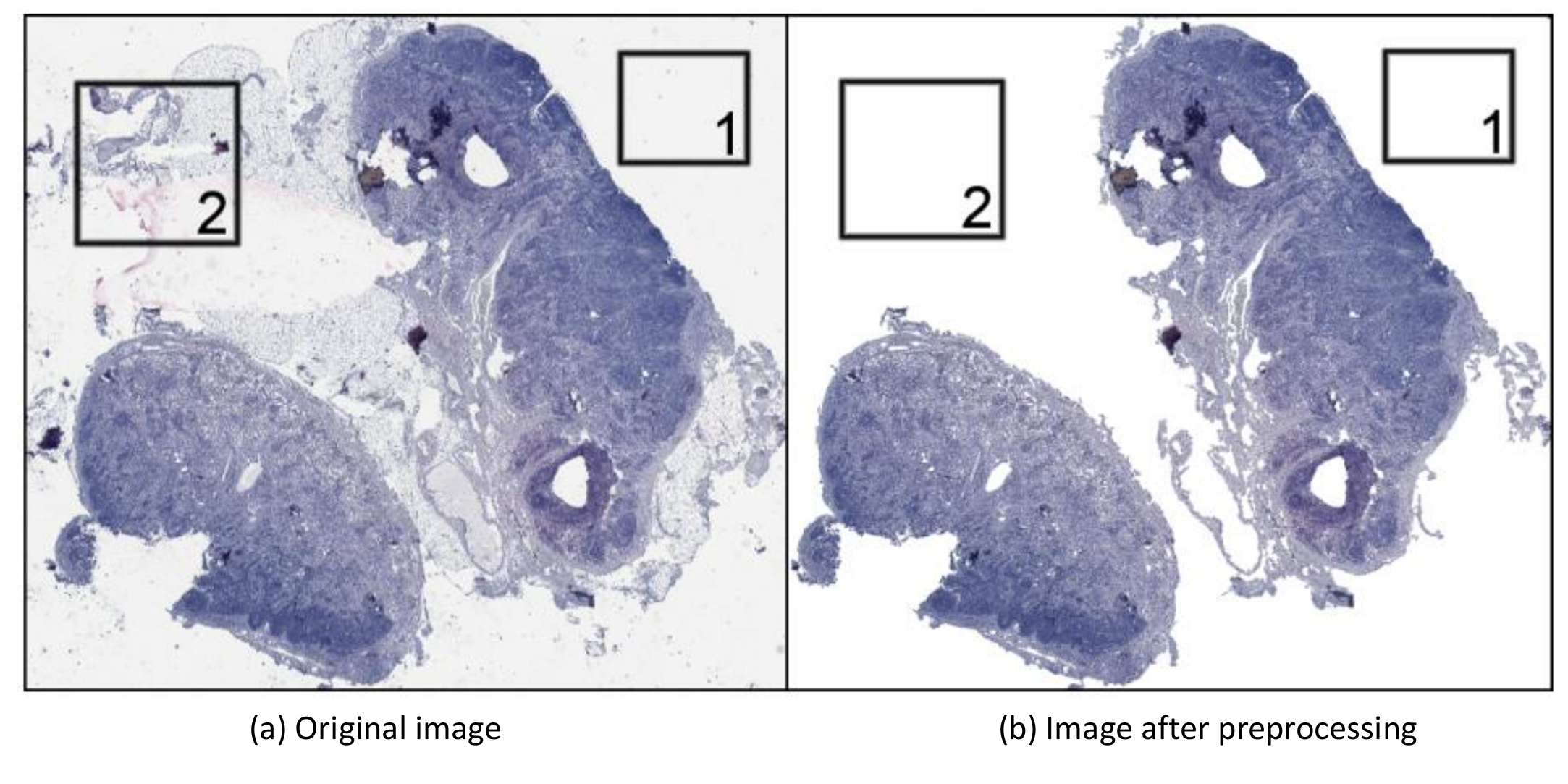}}
\caption{In Box 1, the background pixels are removed. In Box 2, tissue fragments are eliminated. The figure corresponds to Figure. 2 in~\cite{Belkacem-2010-Segmentation}.}
\label{fig:region label}
\end{figure}

In~\cite{Belkacem-2010-Segmentation}, the Gaussian filtering is firstly used to convolve the images. Next, the pixels marked as background in the original image are removed and the pixels marked as tissue are left by background thresholding. Finally, a region labeling method is used to make the remaining pixels are connected. Fig.~\ref{fig:region label} shows the effects of the preprocessing. In~\cite{Schafer-2013-Image} and~\cite{Es-2017-Decision}, the preprocessing methods are similar to~\cite{Belkacem-2010-Segmentation}. Firstly, a weighted averaging filter with weights according to a Gaussian distribution is used to smooth images. Secondly, an intensity threshold obtained from the histogram of the grayscale image is used to separate background points and object points, and the gray scale image is generated by averaging the RGB values of each point. Finally, a region-based segmentation is used to eliminate small isolated tissue areas and connect the object points.

In~\cite{Michail-2014-Morphological}, the preprocessing steps include: (1) The Gray-Level Run Length algorithm and the intensity histogram of nucleus are used to encapsulate the information that the CBs are of a higher non-uniformity than non-CBs. (2) The intensity threshold is used to eliminate the areas without nucleus. (3) Circular Hough Transform is used to images to detect nucleus. (4) Dilation is used to keep the proximate area of the nucleus.

\begin{figure}[!htbp]
\centering
\centerline{\includegraphics[width=0.7\textwidth]{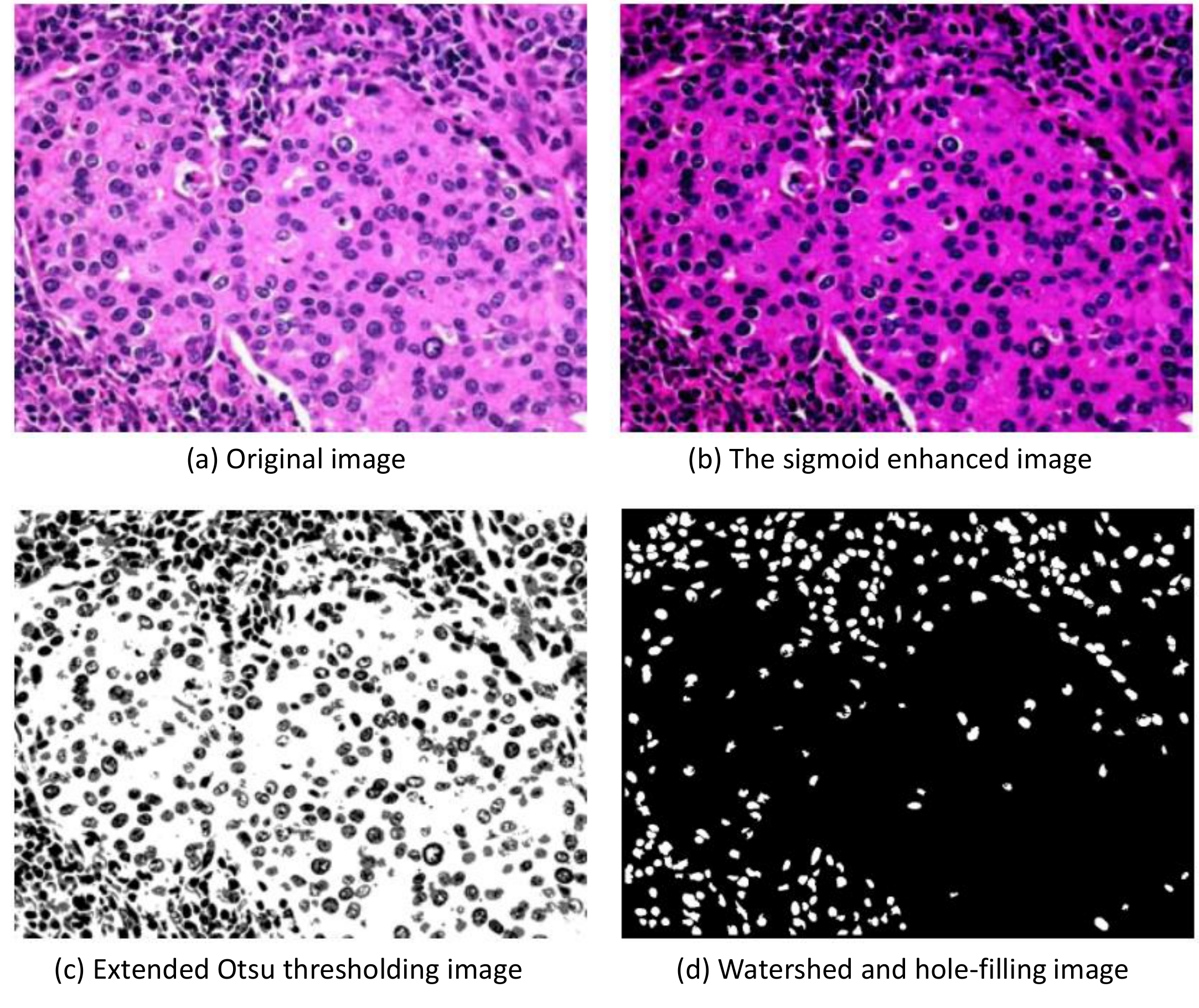}}
\caption{The effect of preprocessing in~\cite{Kuo-2014-Lymphatic}. The figure corresponds to Fig. 3 in~\cite{Kuo-2014-Lymphatic}.}
\label{fig:sigmoid}
\end{figure}

In~\cite{Kuo-2014-Lymphatic}, the preprocessing steps include: (1) Sigmoid contrast enhancement is used to strengthen the contrast of the image. (2) The enhanced image is converted to grayscale by using the V-component of HSV space transferred from RGB space. (3) The threshold for creating the binary image is chosen by extending the original Otsu's method. (4) The holes in images after watershed segmentation are filled by Hasan hole filling algorithm. Fig.~\ref{fig:sigmoid} shows the effect of preprocessing.

In~\cite{Linder-2019-Deep, Li-2020-Multi, Brancati-2019-Deep, Somaratne-2019-Improving, Senaras-2019-Segmentation, Kandel-2020-Novel, Thorat-2020-Classification,Dif-2020-Efficient, Hashimoto-2020-Multi}, data augmentation methods are used to avoid issue of biases and overfitting. The methods include left-right flip, rotation, perturbing color, adding jitter, scaling, shear, flipping (both horizontal and vertical), mirror operation. In~\cite{Dif-2020-Efficient}, the patches are rotated by 0 and 90 degrees after cropping. In~\cite{Hashimoto-2020-Multi}, the patches are rotated to augment data when the number of patches in a WSI is less than 3,000.

\subsection{Summary}
\begin{figure}[!htbp]
\centering
\centerline{\includegraphics[width=0.7\textwidth]{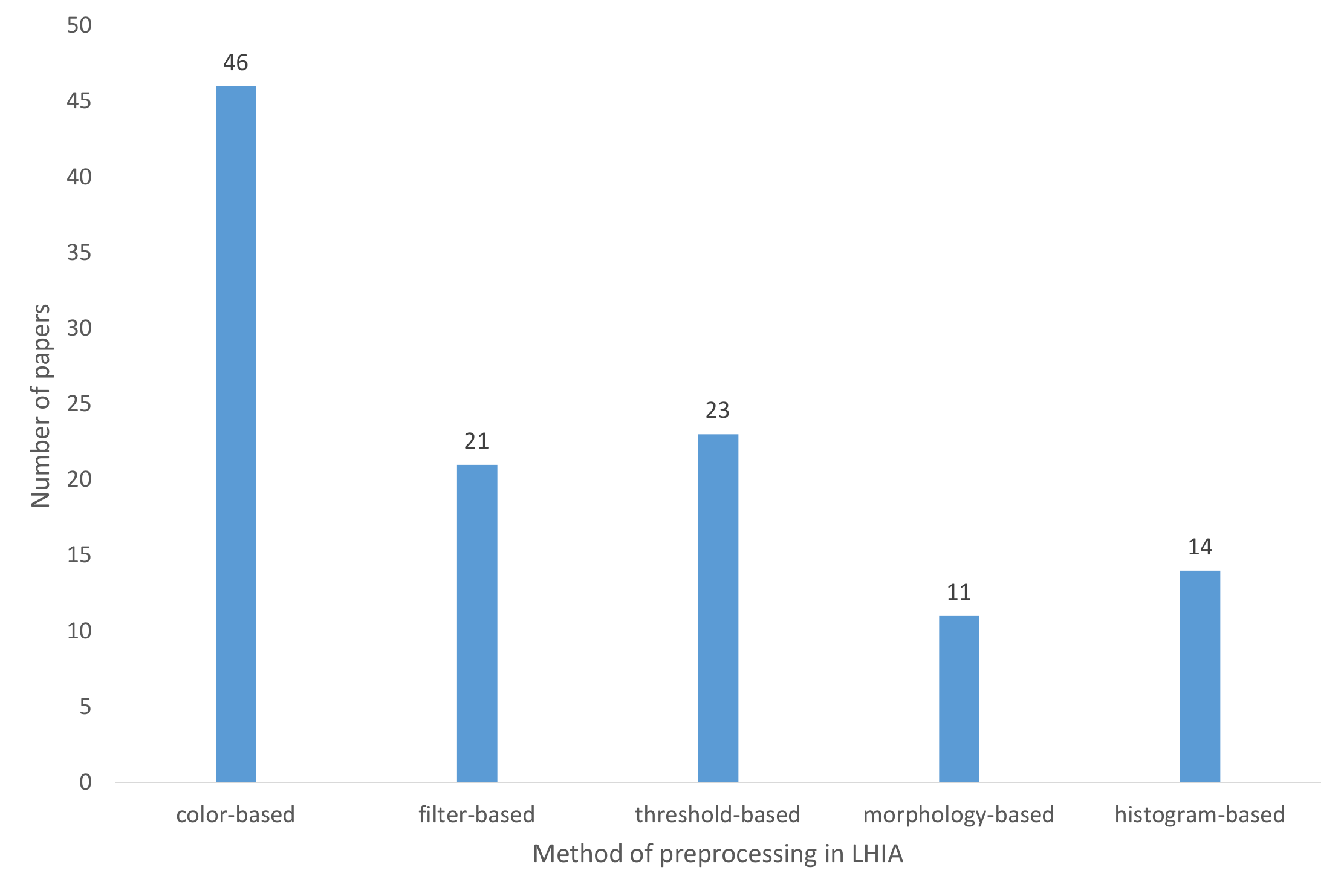}}
\caption{Statistical graphs of four preprocessing methods.}
\label{fig:preprocessing statis}
\end{figure}

In summary, we presented methods used in the preprocessing techniques of LHIA. AS shown in Fig.~\ref{fig:preprocessing statis}, a total of 46 papers used colour-based preprocessing methods, 23 papers used threshold-based methods, 11 papers used morphology-based method, and 14 papers used histogram-based methods. The colour-based methods are the most commonly used method in preprocessing of LHIA, and the HSV colour space is the most commonly used. In addition, the images are frequently cropped into smaller patches and data augmentation methods are also commonly used.

\section{Image Segmentation}
\label{Sec:4}

In this section, we have summarized the segmentation tasks in LHIA. Segmentation is a crucial step in image processing applications, which identifies ROIs, including nuclei, glands, and lymphocytes. The ROIs include follicular regions, lymphocytes, CBs, centrocytes, and neoplastic cells in lymphoma images. It is essential to correlate accurate identification of ROIs with pathologies~\cite{Ong-1996-Image}.

\subsection{Threshold-based Segmentation Methods}
\label{Sec:4.1}

Image thresholding is a widely used segmentation technique that generates binary images by selecting a reasonable threshold. The threshold is determined by using the difference in grayscale characteristics between the ROIs and background tissue in an image. The commonly used threshold-based segmentation methods in LHIA include intensity thresholding, Otsu thresholding and adaptive thresholding. Table.~\ref{Tab:Threshold-based Methods for Segmentation.} shows the references using threshold-based segmentation methods. 

In\cite{Zorman-2007-Symbol}, symbol-based ML method is used for segmenting FL images. The focus is on identifying lymphomas by finding follicles from microscopy images of FL. In the first stage, image preprocessing and feature extraction are implemented, then different rough set methods are used for pixel classification in the second stage. The mean brightness value of the R component of RGB image is used as a threshold for image binarization. Two decision trees are selected and a best overall accuracy of 86.65$\%$ is obtained from the first decision tree on 3627 training objects and 1813 test objects.

\begin{figure}[!htbp]
\centering
\centerline{\includegraphics[width=0.96\textwidth]{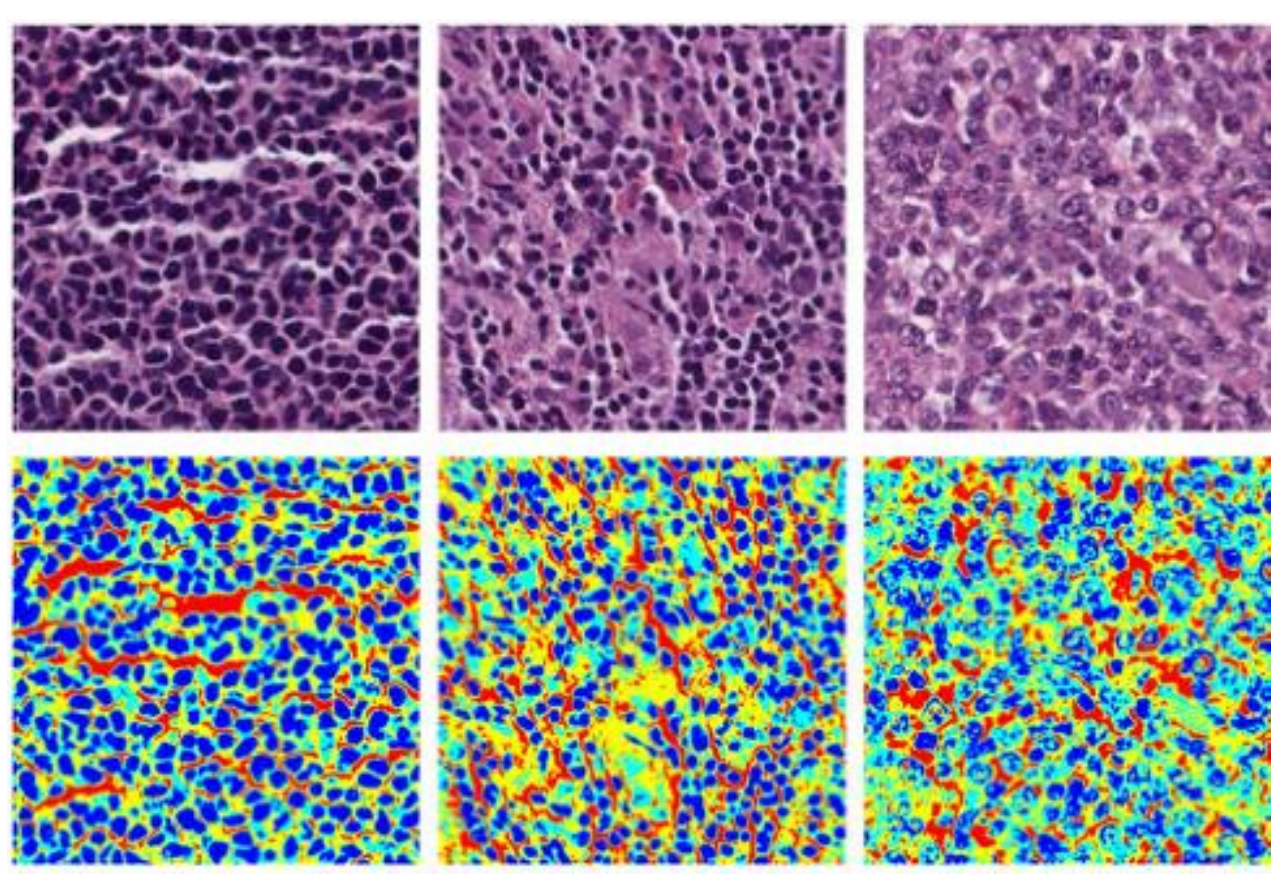}}
\caption{From left to right are grade I, grade II, grade III of FL respectively. Blue corresponds to nuclei, cyan corresponds to the cytoplasm, yellow corresponds to extracellular, red corresponds to the background and gray corresponds to red blood cells. The figure corresponds to Fig. 1 in~\cite{Sertel-2008-Texture}.}
\label{fig:k-means}
\end{figure}

In\cite{Sertel-2008-Texture, Sertel-2009-Histopathological}, since the red blood cells and the background area in the FL images show relatively consistent patterns, thus they are segmented by setting a threshold for the intensity value in the RGB color space. Fig.~\ref{fig:k-means} shows the segmentation results of~\cite{Sertel-2008-Texture}.

\begin{figure}[!htbp]
\centering
\centerline{\includegraphics[width=0.96\textwidth]{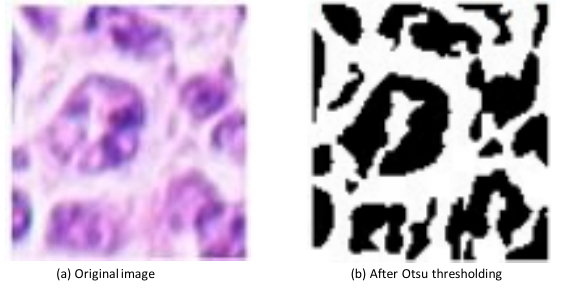}}
\caption{The result after Otsu thresholding. (a) and (b) respectively correspond (a) and (e) of Fig. 2 in~\cite{Michail-2014-Detection}.}
\label{fig:Otsu segment}
\end{figure}

In\cite{Belkacem-2009-Extraction, Michail-2014-Detection, Dimitropoulos-2014-Using, Codella-2016-Lymphoma}, Otsu thresholding is used in the segmentation step. In~\cite{Belkacem-2009-Extraction}, after the conversion of RGB color space to $L^*a^*b^*$ color space, the Otsu thresholding method is used to segment the background tissues for extracting ROIs in images. In~\cite{Michail-2014-Detection, Dimitropoulos-2014-Using}, a threshold of 0.37 is used to remove red blood cells, then Otsu thresholding is used to segment nuclei from extra-cellular and background tissues. Fig.~\ref{fig:Otsu segment} shows the result after Otsu thresholding segmentation. In~\cite{Codella-2016-Lymphoma}, Otsu thresholding is applied to the H$\&$E segmentation. Eosin segmentation and hematoxylin segmentation are realized by setting the highest threshold and the lowest threshold respectively.

In\cite{Neuman-2010-Segmentation}, a novel color image segmentation is used to help pathologists to investigate prognostic factors of FL. The segmentation method includes preprocessing, adaptive threshold, watershed and post-processing. The method is tested on 50 images and compared with the results of manual counting. The method only uses a mean time of 18 seconds which is the mean time of an image proceeding, and 100$\%$ objects are classified in the method.

In\cite{Mandyartha-2020-Global}, global thresholding and local thresholding are evaluated in 35 blood smear test images of acute lymphoblastic leukemia (ALL). Compared with ground truth, local thresholding obtains the mean Zijdenbos Similarity Index (ZSI), precision and recall is 92.5$\%$, 91.79$\%$ and 94.03$\%$ respectively, and global thresholding obtains the mean ZSI, precision and recall are 30.72$\%$, 23.38$\%$ and 99.39$\%$ respectively. The results show that global thresholding can be an alternative method when the number of white blood cell pixels and the number of other pixels reach a balance in an image.

\begin{table}[!htbp]
\renewcommand\arraystretch{2}
\centering
\caption{Threshold-based Methods for Segmentation.}
\label{Tab:Threshold-based Methods for Segmentation.}
\tiny
\begin{tabular}{cccc}
\hline
Reference & Year & Team & Segmentation Method\\
\hline

\cite{Zorman-2007-Symbol} & 2007 & M Zorman & Thresholding using mean brightness value\\

\cite{Sertel-2008-Texture} & 2008 & O Sertel & Thresholding the intensity values of RGB color space\\

\cite{Belkacem-2009-Extraction} & 2009 & K Belkacem-Boussaid & Otsu thresholding\\

\cite{Sertel-2009-Histopathological} & 2009 & O Sertel & Thresholding the intensity values of RGB color space\\

\cite{Neuman-2010-Segmentation} & 2010 & U Neuman & An adaptive threshold\\

\cite{Michail-2014-Detection} & 2014 & E Michail & \tabincell{l}{Red blood cells are removed by thresholding \\with value of 0.37; Otsu is used to segment nuclei}\\

\cite{Dimitropoulos-2014-Using} & 2014 & K Dimitropoulos & \tabincell{l}{Red blood cells are removed by thresholding \\with value of 0.37; Otsu is used to segment nuclei}\\

\cite{Codella-2016-Lymphoma} & 2016 & N Codella & Otsu thresholding \\

\cite{Senaras-2017-Foxp3} & 2017 & C Senaras & \tabincell{l}{Optimal adaptive thresholding method, \\a parameter free elliptical arc and line segment detector, \\Simple Linear Iterative Clustering}\\

\cite{Mandyartha-2020-Global} & 2020 & EP Mandyartha & Global and adaptive thresholding \\
\hline
\end{tabular}
\end{table}

\subsection{Clustering-based Segmentation Methods}
\label{Sec:4.2}

In the clustering algorithm, the goal is to group similar things together. The clustering algorithm usually does not use training data for learning, called unsupervised learning in ML. Contrary to supervised segmentation methods, unsupervised segmentation does not train a classifier. Unsupervised segmentation groups the pixels of an image into several clusters based on similar pixels with a property (color, texture, intensity value, etc.). The commonly used clustering-based segmentation method is $k$-means clustering. Table.~\ref{Tab:Clustering-based Methods for Segmentation.} shows the references using clustering-based segmentation methods.

$k$-means clustering is an iterative process of splitting $n$ objects into $k$ clusters. First, $k$ cluster center points are obtained by calculating the mean of pixels in each cluster. Second, the distance between each object and cluster center points is calculated. Third, an object is assigned to a cluster with minimum distance. Fourth, the cluster center points are recomputed through averaging all objects of a cluster. Finally, the best clustering result is obtained through repeated iterations.

\begin{figure}[!htbp]
\centering
\centerline{\includegraphics[width=0.96\textwidth]{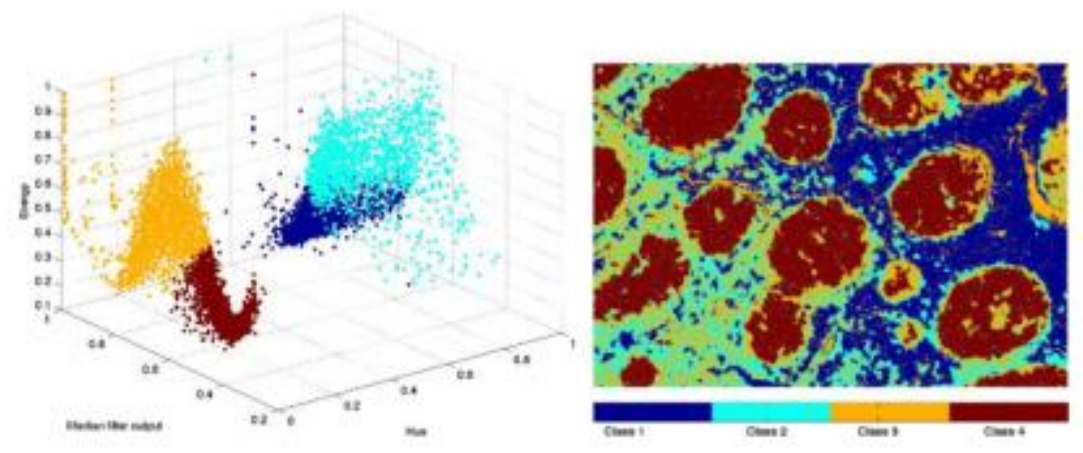}}
\caption{Example of scatter plot of test image after clustering. The figure corresponds to Fig. 3 in~\cite{Samsi-2010-Detection}.}
\label{fig:k-means2}
\end{figure}

In\cite{Sertel-2008-Texture, Sertel-2009-Histopathological, Oztan-2012-Follicular}, in an FL image, the components including nuclei, cytoplasm and extracellular are segmented by $k$-means clustering. Fig.~\ref{fig:k-means} (Sec.~\ref{Sec:4.1}) shows the segmentation results by $k$-means clustering. In\cite{Sertel-2008-Computerized}, $k$-means clustering is used to segment each cell into nuclei, cytoplasm and inter-nuclei materials. In~\cite{Samsi-2010-Detection, Samsi-2012-Efficient}, $k$-means is used to group four clusters including background, cells stained blue, follicle regions in the darker brown regions and the regions which show light brown. Fig.~\ref{fig:k-means2} shows the example of a scatter plot of test image after $k$-means clustering.

In\cite{Han-2010-Multi}, $k$-means clustering is applied to group three structures including mantle zone, germinal centers and inter-follicular regions. Besides, $k$-means clustering is again used for a finer segmentation.

In\cite{Shi-2016-Automated}, a hierarchical $k$-means clustering method is proposed for segmenting FL images. First, $k$-means clustering segments H$\&$E images into nuclei, cytoplasm and extracellular space. However, only core parts of nuclei are segmented, and most nuclei are segmented into cytoplasm because of low contrasts of outer edge in nuclei. Therefore, a second $k$-means clustering, namely hierarchical clustering, is used to segment the regions which are not nuclei into three parts (the three parts are described in the first clustering). 

In\cite{Arora-2013-Computer}, a method is proposed for FL grading. The method segments nuclei through using a locally defined clustering approach to identify the shape of nuclei. In the experimentation, 25 images are used to be training dataset 85 images are used to be test dataset. The method can be used to grade FL and accurately identify grade III.

\begin{table}[!htbp]
\renewcommand\arraystretch{2}
\centering
\caption{Clustering-based Methods for Segmentation.}
\label{Tab:Clustering-based Methods for Segmentation.}
\begin{tabular}{cccc}
\hline
Reference & Year & Team & Segmentation Method\\
\hline

\cite{Sertel-2008-Texture} & 2008 & O Sertel & $k$-means clustering\\

\cite{Sertel-2008-Computerized} & 2008 & O Sertel & $k$-means clustering, eccentricity\\

\cite{Sertel-2009-Histopathological} & 2009 & O Sertel & $k$-means clustering\\

\cite{Samsi-2010-Detection} & 2010 & S Samsi & $k$-means clustering\\

\cite{Han-2010-Multi} & 2010 & J Han & $k$-means clustering\\

\cite{Samsi-2012-Efficient} & 2012 & S Samsi & $k$-means clustering\\

\cite{Oztan-2012-Follicular} & 2012 & B Oztan & $k$-means clustering\\

\cite{Shi-2016-Automated} & 2016 & P Shi & Hierarchical $k$-means clustering\\

\hline
\end{tabular}
\end{table}

\subsection{Region-based Segmentation Methods}
\label{Sec:4.3}

A watershed algorithm is a segmentation method based on region, which segments ROIs by finding watershed lines. In the watershed algorithm, grey-level images are considered a topographic. The grey level of a pixel corresponds to its elevation, and the high grey level corresponds to a mountain, the low grey level corresponds to the valley. When the water level rises to a certain height, the water will overflow the current valley. A dam can be built on the watershed to avoid the collection of water in two valleys. Therefore, the image is divided into 2-pixel sets; one is the valley pixel set submerged by water, and the other is the watershed line pixel set. In the end, the lines formed by these dams partition the entire image to achieve the segmentation result of the image.

\begin{figure}[!htbp]
\centering
\centerline{\includegraphics[width=0.8\textwidth]{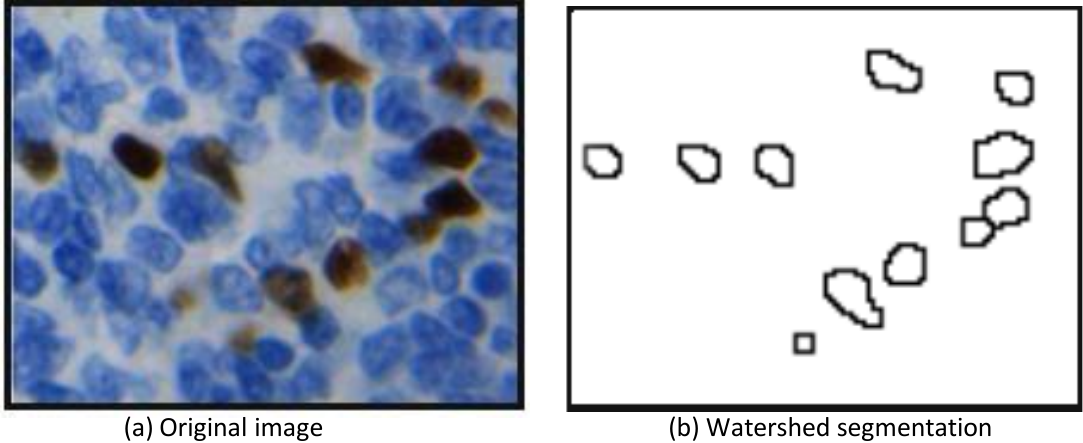}}
\caption{The segmentation result of watershed algorithm. The figure corresponds to Fig. 3 in~\cite{Neuman-2010-Segmentation}.}
\label{fig:watershed1}
\end{figure}

In\cite{Neuman-2010-Segmentation}, a watershed is used to segment nuclei as the second step of segmentation because the white components in the first step of segmentation (adaptive threshold map) are sometimes not related to only one nucleus. Fig.~\ref{fig:watershed1} shows the segmentation result of a watershed algorithm.

\begin{figure}[!htbp]
\centering
\centerline{\includegraphics[width=0.96\textwidth]{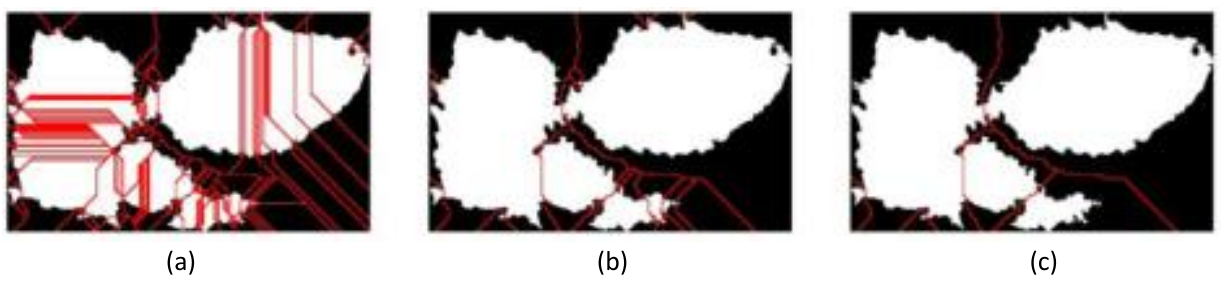}}
\caption{The process of iterative watershed algorithm. (a), (b) and (c) corresponds the first, second and third iteration. The figure corresponds to Fig. 4 in~\cite{Samsi-2010-Detection}.}
\label{fig:watershed2}
\end{figure}

In\cite{Samsi-2010-Detection}, multiple follicles may merge into one region in the process of $k$-means clustering. Therefore, the iterative watershed algorithm is used to split the overlapping regions until the segmentation ends. Fig.~\ref{fig:watershed2} shows the process of the iterative watershed algorithm.

In\cite{Zarella-2015-Lymph}, a method for predicting axillary lymph nodes is proposed. In the segmentation step, watershed transform is applied to segment candidate nuclei detected by SVM.

\subsection{Deep Learning based Segmentation Methods}
\label{Sec:4.5}

In addition to some traditional segmentation methods,\cite{Swiderska-2019-Learning, Bandi-2019-Resolution, Senaras-2019-Segmentation} use deep learning segmentation techniques in LHIA.

In\cite{Swiderska-2019-Learning}, four deep learning methods are used in automatically detecting lymphocytes of histopathology images. The four deep learning methods are: (1) patch classification using FCN; (2) U-net, which is used for segment cells; (3) detection using You Only Look Once (YOLO); (4) prediction using Locality Sensitive Method (LSM). The dataset consists of breast cancer, colon cancer and prostate cancer images, which includes 37 training images, 6 validation images and 40 testing images. Finally, U-net achieves the highest performance of an F1-score equalling 0.78. Besides, U-net achieves a higher agreement ($k$ = 0.72) with manual evaluation, compared with pathologists ($k$ = 0.64).

In\cite{Bandi-2019-Resolution}, a fully convolutional neural network (FCNN) is used for differentiating tissue and background regions for avoiding low-efficient scanning in empty background areas. There are seven convolutional layers, the first six convolutional layers have a ReLU activation function and the last convolutional layer has a softmax activation function. FCNN can achieve similar performance with U-net in segmentation. This dataset contains WSIs of organs in 10 categories, each with 10 WSIs, for a total of 100 WSIs. There are 10 CK8-18 stained and 10 H$\&$E stained lymph node WSIs. FCNN is compared with three traditional methods, achieves Dice scores from 0.97 to 0.98 in a resolution-agnostic network, a Dice score of 0.97 and sensitivity of 0.98 on eight testing images.

\begin{figure}[!htbp]
\centering
\centerline{\includegraphics[width=0.7\textwidth]{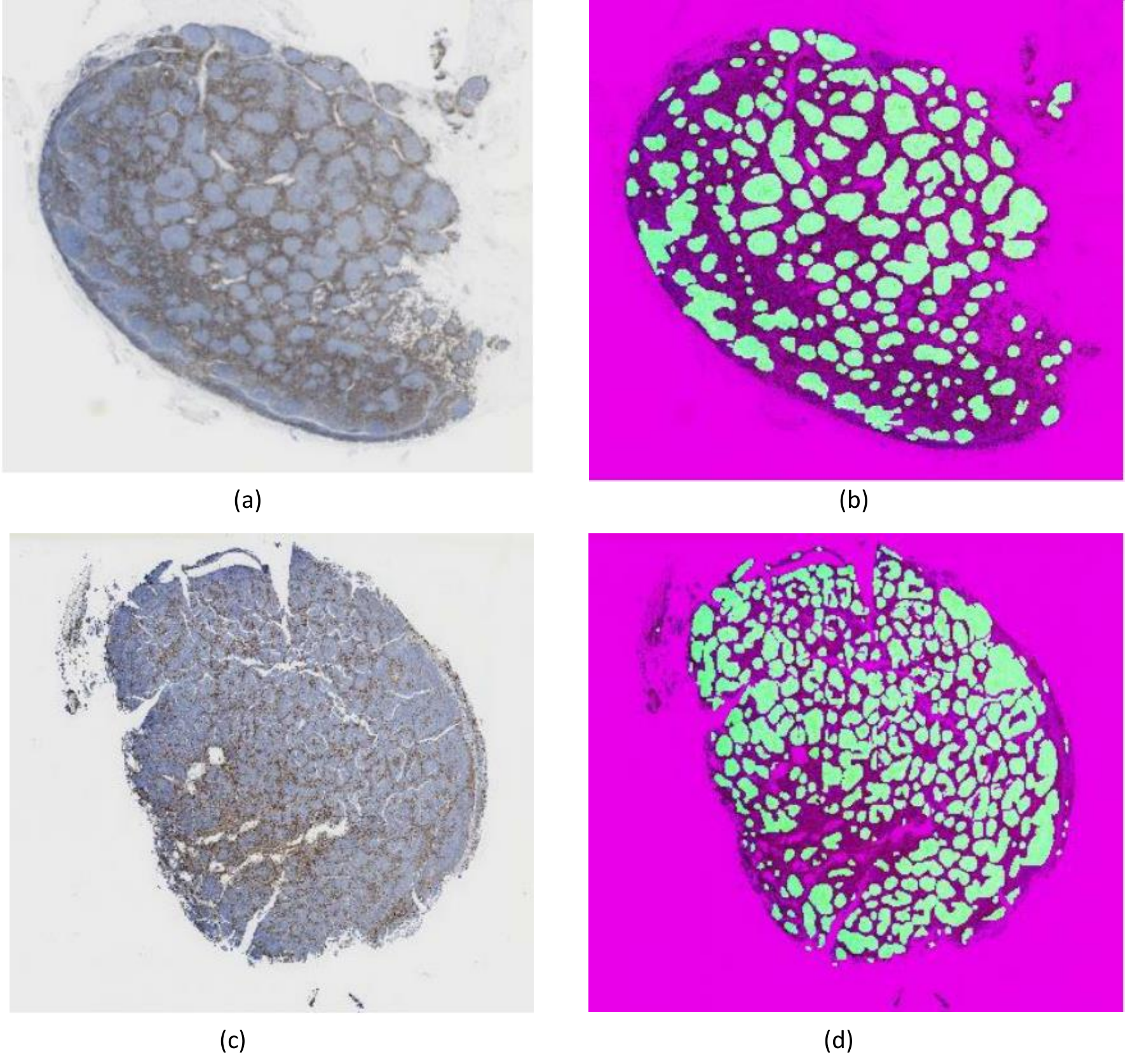}}
\caption{(a) and (c) are original images. (b) and (d) are the segmented results, the green regions are the detected follicles. The figure corresponds to Figure 4 in~\cite{Senaras-2019-Segmentation}.}
\label{fig:U-net1}
\end{figure}

In\cite{Senaras-2019-Segmentation}, a method is developed, which is used to automatically detect follicles in CD8 stained FL images. The method uses U-net to segment follicles in the whole WSI. The method is trained in more than 1000 follicles and is tested on eight whole digital slides. Finally, the method achieves an average Dice similarity coefficient equalling 85.6$\%$ compared with pathologists. Fig.~\ref{fig:U-net1} shows the segmentation results using U-net.

\subsection{Other Segmentation Methods}
\label{Sec:4.6}

In addition to the above segmentation methods, some other segmentation methods also achieve good results.

In\cite{Belkacem-2010-Segmentation}, a technique is developed to segment the follicular regions in the H$\&$E stained FL images. The technique is based on an active contour model and is initialized through users manually selecting seed points in follicular regions. The technique uses matched filter to remove noises and flatten the background of L channel after the color space conversion of RGB to $L^*a^*b^*$. The technique is tested on 40 FL images and the mean accuracy is 0.71, the standard deviation is 0.12.

\begin{figure}[!htbp]
\centering
\centerline{\includegraphics[width=0.96\textwidth]{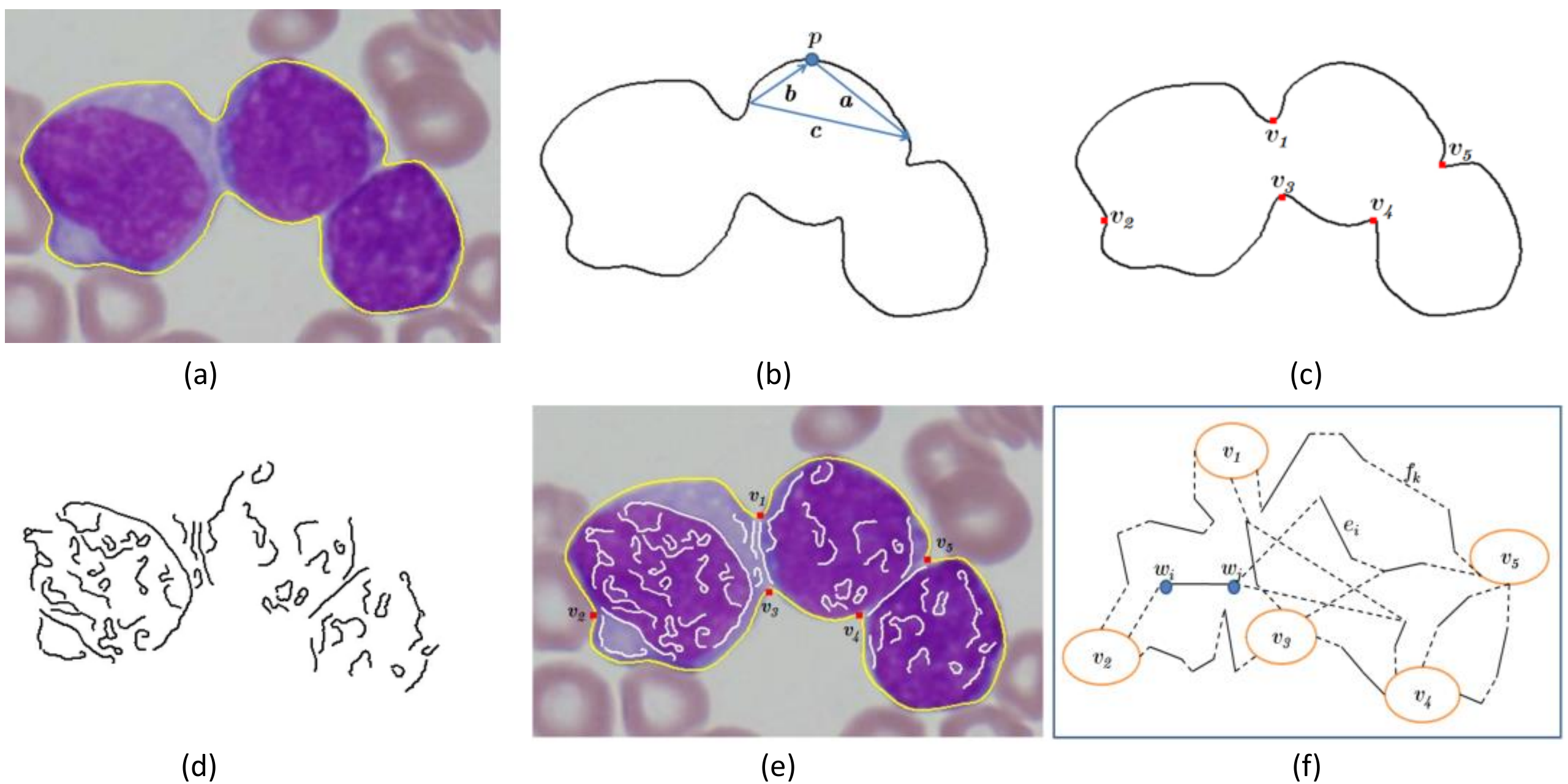}}
\caption{The process of constructing the concave vertex graph. (a) Original image with yellow boundary. (b) Detection of high curvature points. (c) Detection of concave points. (d) Detection of inner edges. (e) The original image is superimposed by outer boundary, concave vertices and inner edges. (f) The concave vertex graph. These figures (a)-(f) correspond to (a)-(f) of Fig. 3 in~\cite{Yang-2008-Automatic}.}
\label{fig:concave graph}
\end{figure}

In\cite{Yang-2008-Automatic}, an algorithm is proposed to separate touching cells. In the algorithm, the outer boundary is delineated by a L$_{2}$E robust estimation and a robust gradient vector flow snake, then concave points on the inner edges and boundary are detected automatically. These points construct a concave vertex graph. Finally, the touching cells are separated by recursively calculating the optimal path in the concave vertex graph after minimizing a cost function. The algorithm is tested on two datasets including 207 touching images and 3898 images and an average accuracy of 90.1$\%$ is obtained. Fig.~\ref{fig:concave graph} shows the process of constructing the concave vertex graph.

\begin{figure}[!htbp]
\centering
\centerline{\includegraphics[width=0.96\textwidth]{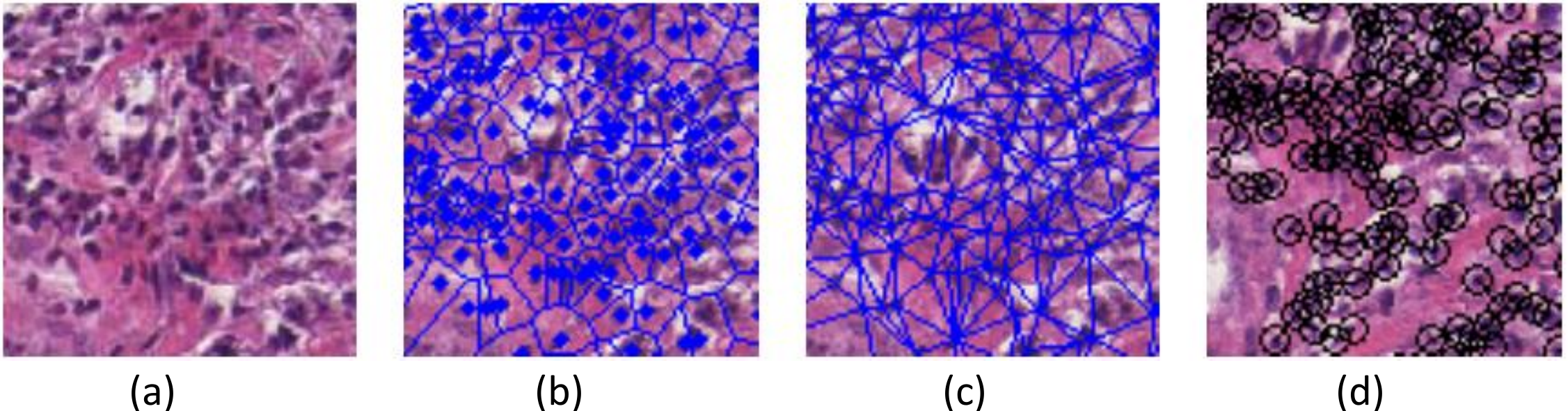}}
\caption{(a) Original image. (b) Voronoi diagram. (c) Delaunay triangulation. (d) Minimum spanning tree showed in segmented lymphocytes. The figure corresponds to Fig. 3 in~\cite{Basavanhally-2008-Manifold}.}
\label{fig:Bayesian and template}
\end{figure}

In\cite{Basavanhally-2008-Manifold}, a segmentation method is proposed to automatically detect lymphocytes, which consists of a Bayesian classifier and template matching. the segmentation algorithm takes advantage of the strong hematoxylin staining of lymph infiltration through the Bayesian classifier, template matching can be used to discard the false positive results because of the size and shape of the lymphocyte. Fig.~\ref{fig:Bayesian and template} shows the result of the proposed segmentation method.
 
\begin{figure}[!htbp]
\centering
\centerline{\includegraphics[width=0.96\textwidth]{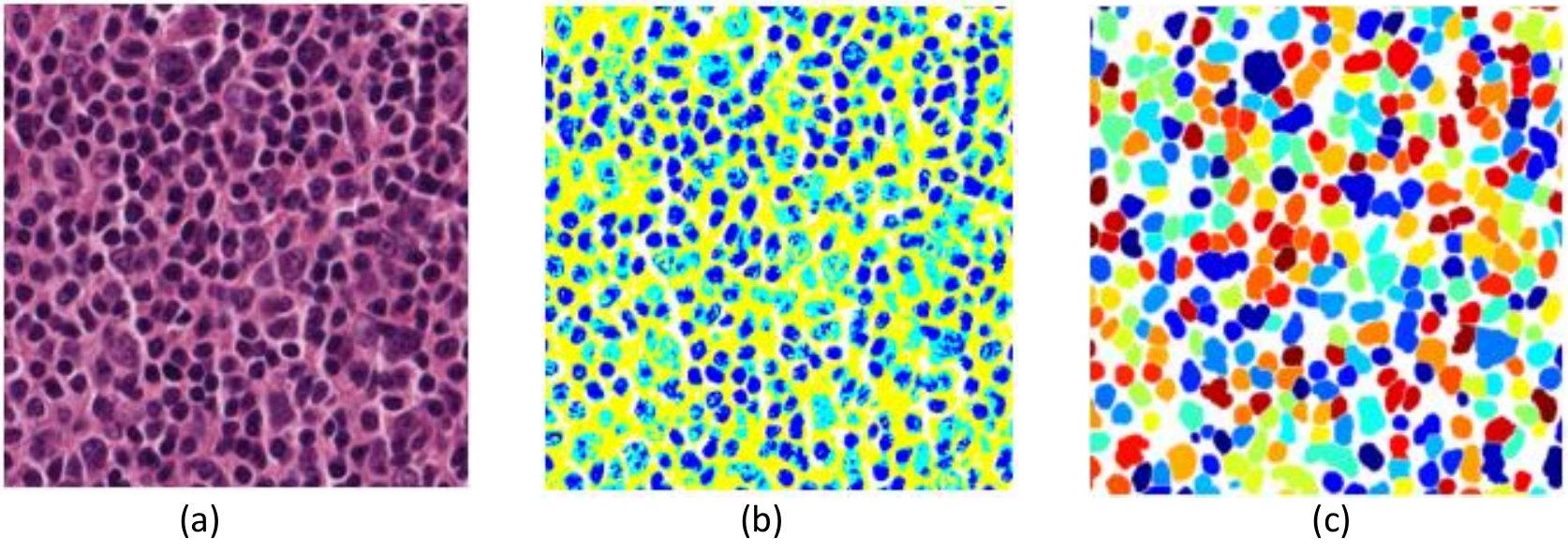}}
\caption{(a) Original image. (b) The segmentation result of mean-shift. (c) The final result. The figure corresponds to Figure 1 and Figure 2 in~\cite{Sertel-2010-Image}.}
\label{fig:mean-shift}
\end{figure}

In\cite{Sertel-2010-Image}, an image analysis system is proposed to quantitatively evaluate digitized FL tissue slides. In the segmentation step, the tissue is segmented into cytological components by the mean-shift algorithm for segmenting individual cells. The mean-shift algorithm does not need to know the number of clusters and does not control the shape of clusters. Then, morphological operations are used in post-processing. Fig.~\ref{fig:mean-shift} shows the process of segmentation.

In\cite{Cheng-2010-Identifying}, a multi-phase level set framework is used to initial segmentation of cells. The method avoids the vacuum and overlap problems. In the method, $n$ = 4 phase segmentation is selected to separate the ROIs for there are four regions with different intensity levels in original images.

In\cite{Sertel-2010-Computer}, a computer-aided detection system is proposed, which is used to automatically detect CB cells in FL images. In the segmentation step, the RGB images are converted into 1-D unitone images by Principal Components Analysis (PCA). Then, the unitone images are normalized to a range of [0, 1]. Next, the individual cells are segmented through two steps: (1) expectation, which is used to calculate the likelihood relative to the current estimation. (2) maximization, which is used to maximize the expected log likelihood.

In\cite{Kong-2011-Partitioning}, an integrated framework is proposed, which comprises an algorithm of supervised cell-image segmentation and a method of splitting touching-cell. In this method, the cell tissue and extracellular tissue are segmented. The color-texture feature extracted from the local neighborhood of each pixel is used as the input of the classification algorithm. Local Fourier transform (LFT) texture features are extracted from MDC color space (described in Sec.~\ref{Sec:3.1}). In the segmentation part, the boundaries are smoothed by the Fourier shape descriptor after differentiating between a touching-cell clump and a single non-touching cell. The approach is tested on 750 positive and 750 negative training patches and each image achieves a total error rate of 5.25$\%$.

In\cite{Kong-2011-Splitting}, a new algorithm is proposed, which is applied for splitting touching/overlapping cells. Similar to the segmentation part of~\cite{Kong-2011-Partitioning}, the boundaries are smoothed by Fourier shape descriptor after differentiating between a touching-cell clump and a single non-touching cell. Then, a novel iterative splitting algorithm is only used to split the touching-cell clump. Finally, the splitting method is tested on 21 FL images and obtain an average error rate which equals 5.2$\%$.

In\cite{Oger-2012-General}, a framework is proposed, which is used to segment follicular regions before histological grading. First, the mask of follicular boundaries is generated. Then, the boundaries are mapped onto the corresponding registered images. The framework is tested on 12 H$\&$E and IHC stained FL images. Finally, the maximum average of sensitivity, specificity, conformity and JAC is 0.935, 0.843, 0.198 and 0.57 respectively.

In\cite{Es-2017-Decision}, a system is introduced, which can automatically differentiate the categories for diffuse lymphoma cells. In the segmentation step, an image pixel is segmented into multiple pixel regions through the minimum variance quantization approach.

In\cite{Tosta-2017-Computational}, an unsupervised method used for segmenting the nuclear components is proposed. The method is based on fuzzy 3-partition entropy and genetic algorithm. The genetic algorithm initial population is needed to define, then the images after preprocessed are calculated and normalized, the fuzzy 3-partition entropy approach is used to define the values of each individual. The dataset used in the segmentation method includes 12 CLL, 62 FL and 99 MCL images. The mean value of accuracy of the method is 81.48$\%$.

In\cite{Tosta-2018-Fitness}, an unsupervised segmentation algorithm using a genetic algorithm is proposed for obtaining object diagnoses, which is applied to segment nuclei in CLL and FL images. The algorithm segments nuclei through evaluating different genetic algorithms fitness functions. SVM is used to classify using the features extracted from segmented regions. The algorithm is tested on 12 CLL and 62 FL images and obtains an accuracy of classification equalling 98.14$\%$.

\begin{figure}[!htbp]
\centering
\centerline{\includegraphics[width=0.96\textwidth]{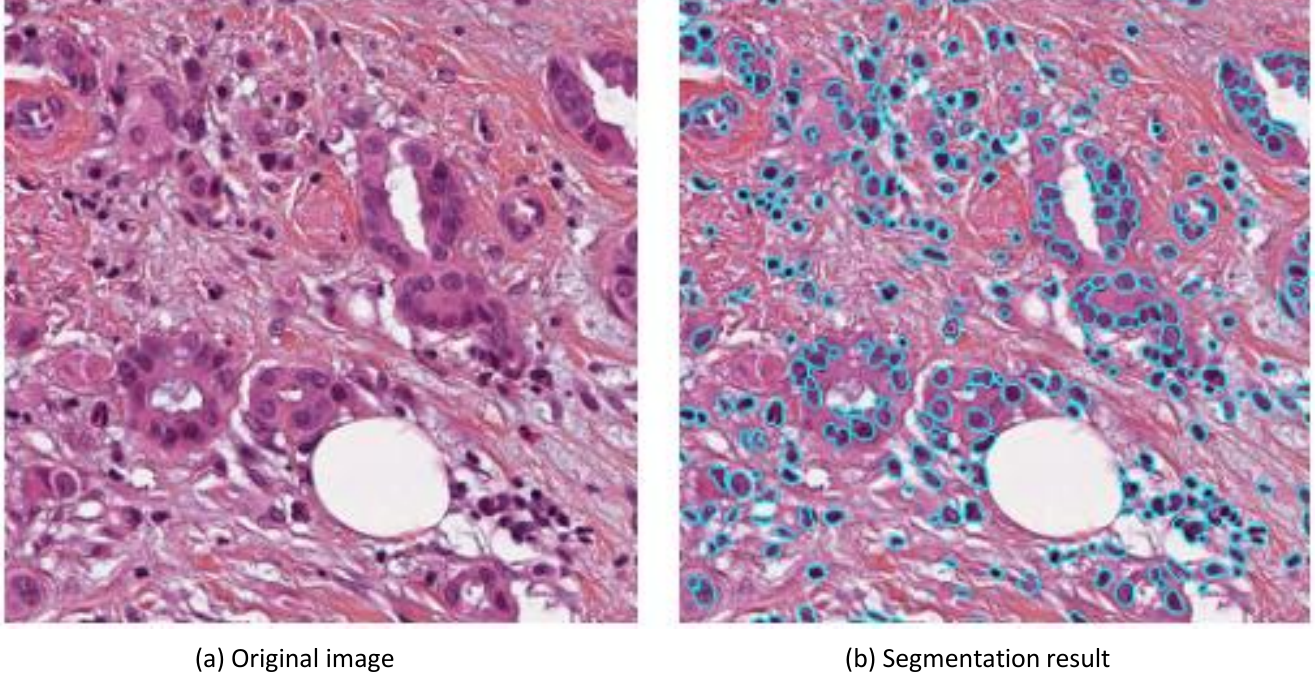}}
\caption{The result of segmentation. The figure corresponds to Fig. 1 in~\cite{Cheikh-2017-Spatial}.}
\label{fig:SLIC}
\end{figure}

In\cite{Cheikh-2017-Spatial}, a fast superpixel segmentation algorithm is applied. The number of expected superpixels is set to 3500 and the compact factor is set to 35. First, the color deconvolution algorithm (described in Sec.~\ref{Sec:3.1}) is applied to differentiate nuclear regions. Second, histogram stretching is used in the hematoxylin channel of the output of the color deconvolution algorithm. Third, the nuclei are extracted by image thresholding followed by filtering and small object removal. Finally, a superpixel segmentation algorithm is used because some segmented nuclei may overlap. The segmentation result is shown in Fig.~\ref{fig:SLIC}.

In\cite{Dimitropoulos-2017-Automated}, a new framework is presented, which is developed to detect and classify CBs in FL images stained with H$\&$E and PAX5. An energy minimization approach based on graph cuts is used in the segmentation of nuclei in PAX5 stained images. The approach is a kind of unsupervised approach. The process of segmentation is a labeling problem, the label stands for different cytological components. After the nuclei segmentation, there are many overlapping nuclei that need to be split. First, the spurious seeds are identified and discarded. Then, each nucleus is detected by using an ellipsoidal model. Finally, a validation criterion is applied to identify spurious seeds.

\subsection{Summary}

From the above summary of the segmentation methods in LHIA, the threshold-based segmentation methods and $k$-means clustering-based segmentation methods in the traditional methods are more commonly used. In terms of deep learning, U-net is an efficient segmentation method, and FCNN can also achieve similar performance to U-net with low computational cost. In addition to the traditional and deep learning segmentation methods, some other segmentation methods also achieve good performance.

\section{Feature Extraction}
\label{Sec:5}

In image processing tasks, it is expected to extract features from the output of the image segmentation. Feature extraction is a crucial step in histopathology image analysis, which is a process of extracting features of the object from segmented images. The extracted features are generally used in classification or detection tasks. The set which consists of all quantified features is defined as a feature set or a feature vector. The features are defined as original characteristics or attributes of an image, which include visual features and statistical features. Visual features generally include color, texture, shape, etc. Statistical features need to be measured manually, which include spectrum, histogram, etc. The extraction methods are crucial for the following classification and detection tasks. This section introduces the commonly used features and methods of extracting features in LHIA. Table.~\ref{Tab:Feature Extraction Methods1} and Table.~\ref{Tab:Feature Extraction Methods2} show the commonly used feature extraction methods, including traditional and deep learning-based methods.

\subsection{Visual Features Extraction}
\label{Sec:5.1}

The visual features of an image include color, shape, texture.

\subsubsection{Color Features Extraction}

Color features belong to the internal features of an image and describe the surface properties of images. Color features are widely used in CAD system.

In\cite{Zorman-2007-Symbol, Zorman-2011-Classification}, the contours of a follicle are different from the other follicles in the image, so the method of extracting features focuses on the distinction. First, the R component of RGB color space is extracted for it has the best contrast compared with the background areas. The mean brightness of the component is used as a threshold for image binarization. Because the density of white pixels in follicle contours is relatively high, so a method of measuring density is proposed. Then, four transformation matrices are obtained, which can provide a feature vector for each pixel of the original image. In the method, the transform matrices are obtained through scanning the binary image obtained from thresholding on column-by-column and row-by-row.

\begin{figure}[!htbp]
\centering
\centerline{\includegraphics[width=0.8\textwidth]{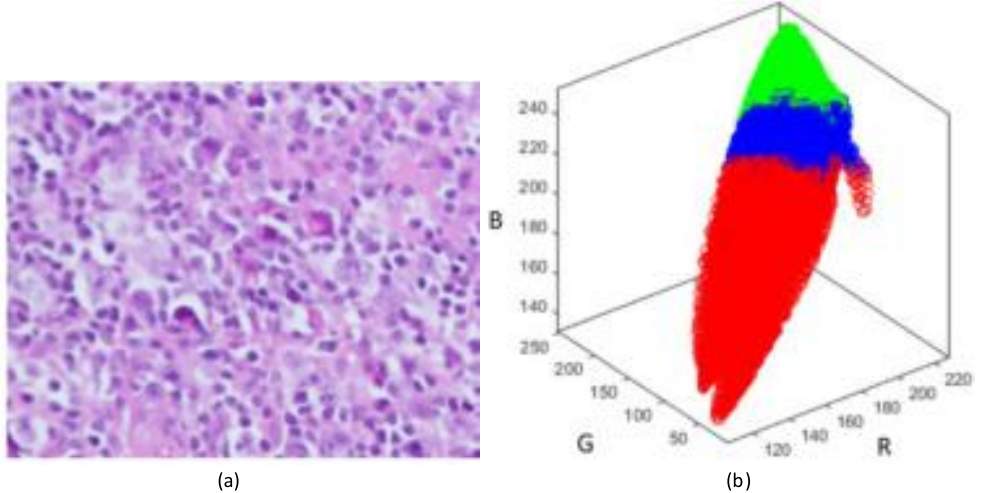}}
\caption{(a) Original image. (b) The red part represents extracellular spaces, the green part represents cytoplasm and the blue part represents nuclei. RGB intensities represent the coordinate axes. The figure corresponds to Figure 2 in~\cite{Shi-2016-Automated}.}
\label{fig:RGB intensity}
\end{figure}

RGB color space is a commonly used color space for color feature extraction. The references using RGB color space include~\cite{Akakin-2012-Content, Acar-2013-Tensor}. In~\cite{Shi-2016-Automated}, RGB color space is used to extract color intensities. Fig.~\ref{fig:RGB intensity} shows the distribution map of pixels in feature space after first clustering. In addition, the mean and standard deviation of a single pixel is also extracted in each color channel of RGB. In~\cite{Bianconi-2020-Experimental}, the color histograms are extracted in RGB color space.

In\cite{Sertel-2008-Computerized, Samsi-2010-Detection, Han-2010-Multi, Oger-2012-General, Akakin-2012-Content, Zarella-2015-Lymph}, HSV color space are used to extract color features. In~\cite{Samsi-2010-Detection, Samsi-2012-Efficient}, the pixel values in the H channel are extracted for clustering.

In\cite{Sertel-2010-Image}, the $L^*u^*v^*$ is used to construct feature space.

In\cite{Belkacem-2010-Effect}, the L channel of $L^*a^*b^*$ is used to extract morphological features. In~\cite{Akakin-2012-Content}, the color features extracted from $L^*a^*b^*$ represent images intensity and color information of images.

In\cite{Meng-2010-Histology}, each image is divided into 25 blocks, the color features are extracted from these blocks. The extracted color features include color dominant, color histogram and color moment. Same as~\cite{Meng-2010-Histology}, \cite{Meng-2013-Multimodal} first divides the original image into 25 blocks and a total of 505 visual features are extracted from each block. The extracted color features include color dominant, color moment, color histogram and edge histogram.

In\cite{Kong-2011-Partitioning}, LFT is used to extract color features from RGB color space.

In\cite{Fauzi-2015-Classification}, a color histogram with 64-dimension is extracted from the 64$\times$64 blocks of 3771 CBs and 4000 non-CBs. In~\cite{Bianconi-2020-Experimental}, color histogram of three dimension and marginal color histogram of one dimensions are extracted in RGB color space.

\subsubsection{Texture Features Extraction}

\begin{figure}[!htbp]
\centering
\centerline{\includegraphics[width=0.8\textwidth]{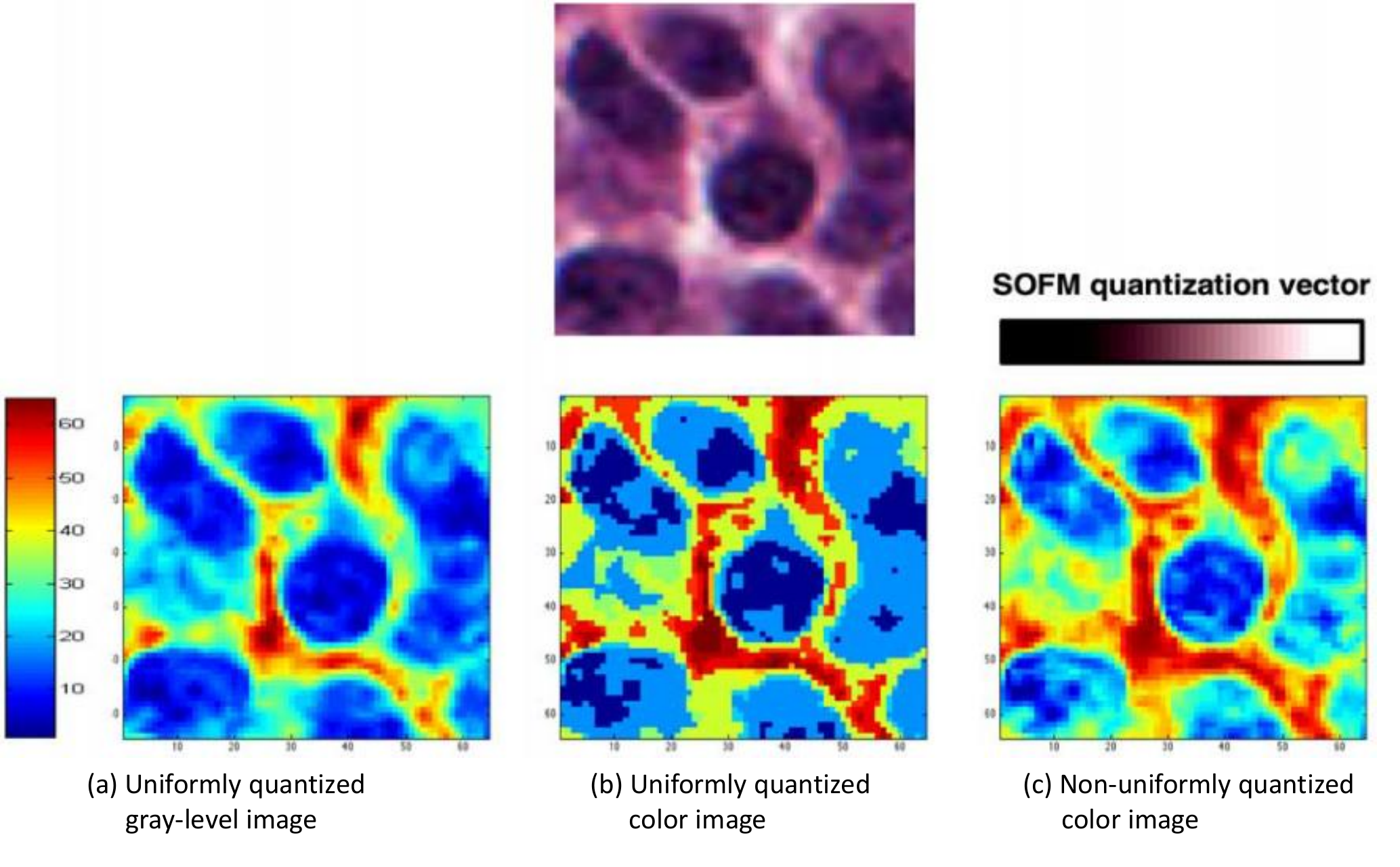}}
\caption{Quantized images produced by uniform gray-level, color quantization and SOFM. The figure corresponds to Fig. 3 in~\cite{Sertel-2008-Texture}.}
\label{fig:SOFM1}
\end{figure}

In\cite{Sertel-2008-Texture}, a new method applied to grade FL is proposed. This method improves the \emph{Gray Level Co-occurrence Matrix} (GLCM) by using self-organizing feature maps (SOFMs). After segmentation, SOFM extracts features of non-linear color quantization and color-texture. Color quantization is a process of constructing a color space through selecting several representative colors from the set of the whole RGB color space. The texture features are extracted by using GLCM from the quantized color space. This method is compared with gray-level and uniform and color quantization method, Fig.~\ref{fig:SOFM1} shows the compare results. From the figure, the proposed method provides a better enhanced image that can produce more detailed texture features. The method is tested on 510 images through $k$-fold cross validation. Finally, a Bayesian classifier is applied, and an overall correct classification rate of 88.9$\%$ and 100$\%$ sensitivity are obtained.

In\cite{Sertel-2009-Histopathological}, first, $k$-means clustering is used to segmentation (in Sec.~\ref{Sec:4.2}). Second, a method for constructing color texture features is proposed, which is based on the combination of the model-based intermediate representation (MBIR) with self-organizing maps (SOM). Third, the dimension of feature space is reduced by the combination of PCA and linear discriminant analysis (LDA). Finally, Bayesian is used as the classifier on the dataset including 180, 240 and 90 grade I, grade II and grade III images respectively. This method has the best identification on the grade III with a sensitivity equalling 98.9$\%$ and a specificity equalling 98.7$\%$, besides, the overall classification accuracy is 85.5$\%$.

The co-occurrence matrix provides a statistical approach, which is used to characterize the spatial dependencies of gray levels of an image\cite{Haralick-1973-Textural}. In\cite{Sertel-2008-Computerized}, co-occurrence matrix method is used to extract homogeneity feature. In~\cite{Samsi-2010-Detection}, energy metric is extracted for determining the locations of follicle. In~\cite{Oger-2012-General}, a co-occurrence matrix is used to quantify the texture features. In~\cite{Samsi-2012-Efficient}, there are two texture features are extracted. One is the output of a grayscale image after a median filter, another is the energy obtained using a co-occurrence matrix. In~\cite{Oztan-2012-Follicular}, co-occurrence extracts five color texture features including homogeneity, energy, contrast, correlation and entropy.

In\cite{Belkacem-2009-Extraction}, a method for automatically identifying CB cells and non-CB cells is proposed. The step of cell identification includes color space conversion of RGB to $L^*a^*b^*$, Otsu thresholding, morphology operations, extraction of area and perimeter of the cell. Then, the principal component analysis (PCA) is applied to extract color texture features in three color spaces including RGB, HSI and $L^*a^*b^*$. Next, a quadratic discriminant analysis (QDA) is used for classification through 10-fold cross validation. The H$\&$E stained dataset includes 218 CB and 218 non-CB images, the dataset is divided with an 80$\%$-20$\%$ ratio. Finally, a classification accuracy equalling 82.56$\%$ is obtained. In~\cite{Belkacem-2010-Computer}, the same method and dataset are used for automatically detecting CB and non-CB images of FL. In the experiment result, the classification accuracy, sensitivity and specificity are 82.56$\%$, 86.67$\%$ and 86.96$\%$ respectively.

In\cite{Belkacem-2010-Effect}, a method similar to\cite{Belkacem-2009-Extraction} is applied to recognize CB cells from non-CB cells in FL images. In this method, color texture features extracted by using PCA are used in the QDA classifier and variance is used to quantify color texture features. 

In\cite{Cheng-2010-Identifying}, a total of 173 extracted texture features include Zernike features, Daubechies, wavelet features, Gabor features, skeleton features and Haralick features. SVM Recursive Feature Elimination (SVM-RFE) is applied to eliminate the irrelevant features. In\cite{Meng-2013-Multimodal}, the texture features extracted from each block include texture co-occur, texture Tamura, texture wavelet, texture Gabor and local binary patterns (LBPs). In~\cite{Acar-2013-Tensor}, the extracted texture features include entropy, homogeneity, mean and contrast.

In\cite{Sertel-2010-Computer}, the \emph{Gray-Level Run Length Matrix} (GLRLM) is applied to derive higher order statistics to represent the texture of images. The features include the features of a wide spectrum and texture. In~\cite{Michail-2014-Morphological}, the Gray-Level Run Length (GLRL) algorithm is applied for calculating the Gray-Level Non-uniformity (GLN) of nuclei, GLN is used for quantifying the texture abnormalities. The mean and skewness of histogram are extracted by calculating the grayscale histogram of nuclei. In~\cite{Shi-2016-Automated}, skewness and kurtosis are extracted. Skewness is used to measure the symmetry of local signals, kurtosis is used to measure the signal intensity under a normal distribution.

In\cite{Di-2015-Different}, an analysis is conducted, which is applied for cells and tissues. In this work, different texture features are extracted from different color spaces and are used in classification, for the purpose of determining the feature set which can be used in different classification tasks. In order to determine the feature set, the existing method of grayscale of color images is generalized. The generalized method is used in calculating matrices of gray-level co-occurrence, gray-level difference and gray-level run-length. SVM is applied for classification using the IICBU-2008 dataset to compare the feature sets. The experiment is conducted in five color spaces, and the result in HSV is the best.

In\cite{Bianconi-2020-Experimental}, texture features including contrast, correlation, energy and entropy are extracted by Gray-Level Co-occurrence Matrices (GLCM).

In\cite{Fatakdawala-2010-Expectation}, the first-order statistical features of texture including standard deviation and average intensity are extracted is used for $k$-means clustering.

In\cite{Meng-2010-Histology}, texture features extracted from blocks include texture co-occur, texture wavelet, texture Tamura, texture Gabor and LBPs.

In\cite{Kong-2011-Partitioning}, an LFT algorithm is developed to speed up the process of extracting texture features, which is based on image shifting and image integral and is applied on MDC color space.

In\cite{Akakin-2012-Content}, the texture features extracted using co-occurrence histogram includes mean, standard deviation, correlation, energy, contrast, entropy and homogeneity. The texture feature extracted using normalized co-occurrence matrix include mean, entropy and homogeneity.

In~\cite{Sandhya-2013-Automated}, Log-Gabor filter is used to extract texture features which include entropy, contrast, homogeneity and spatial frequency. In~\cite{Bianconi-2020-Experimental}, the mean and standard deviation are extracted through Gabor filters.

\begin{figure}[!htbp]
\centering
\centerline{\includegraphics[width=0.8\textwidth]{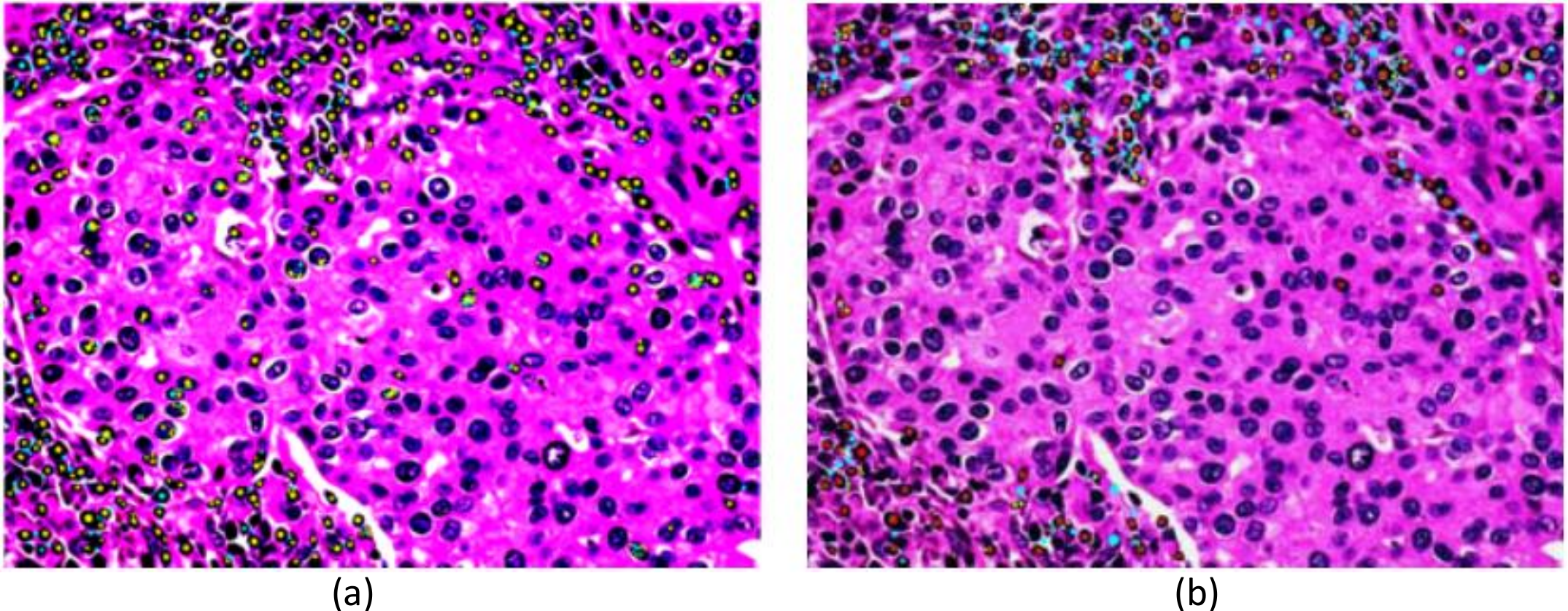}}
\caption{The key points retrieved by SIFT (color online). (a) Key points without morphological refining. (b) Key points after morphological refining. The figure corresponds to Fig. 4 in~\cite{Kuo-2014-Lymphatic}.}
\label{fig:SIFT}
\end{figure}

SIFT is an algorithm for describing local characteristics of an image in CV, it extracts features through finding angles, location and the grayscale at the extreme points. Features extracted through SIFT have characteristics that are scale and rotation invariant. In~\cite{Kuo-2014-Lymphatic}, a method for automatically detecting lymphocytes is proposed. Scale-Invariant Feature Transform (SIFT) is used in the step of feature extraction. Fig.~\ref{fig:SIFT} shows the key points retrieved by SIFT.

\subsubsection{Shape Feature Extraction}

In\cite{Yang-2008-Automatic}, the contour of boundary of the touching cells is extracted by the followed steps: First, L$_2$E robust estimation is applied to produce a rough estimation for the boundaries. Second, a robust gradient vector flow extracts the contour from the background.

In\cite{Sertel-2008-Computerized, Cooper-2009-Feature}, the size of a cell, eccentricity are extracted for feature construction. In~\cite{Belkacem-2009-Extraction}, the area and perimeter of cells are extracted. In~\cite{Sertel-2010-Image}, area and nuclear to cytoplasm ratio are extracted in the feature vector. In~\cite{Belkacem-2010-Effect}, the largest area is extracted through Otsu thresholding, opening, closing, labeling and area classification. In~\cite{Oztan-2012-Follicular}, the morphological features extracted include the length of major axes, length of minor axes and area of ellipses. In~\cite{Arora-2013-Computer}, two features are extracted from segmented images, which include curvature and center of the nuclei. In~\cite{Sandhya-2013-Automated}, the extracted shape features include area, bounding box, perimeter, convex area, solidity, major axes and minor axes. In~\cite{Zarella-2015-Lymph}, six features are extracted including perimeter, area, aspect-ratio, circularity and two measurements quantifying the shape of nuclei. In~\cite{Wang-2016-Deep}, a total of 28 geometrical and morphological features are extracted, which include the percentage of region of the tumor over the region of whole tissue, the area ratio, mean and the longest axis. In~\cite{Chen-2016-Identifying}, the extracted geometrical and morphological features include max, mean, skewness, variance and kurtosis, etc.

In\cite{Michail-2014-Morphological}, individual nuclei and their size construct the first feature through connected component labeling. Then, the perimeter of nuclei is extracted. Next, aspect ratio and ellipse residual are extracted after estimating the best fitting ellipse through the Orthogonal Distance Regression (ODR).
 
\subsection{Statistical Feature Extraction}
\label{Sec:5.2}

In\cite{Sertel-2008-Computerized}, skewness, energy, mean intensity, kurtosis and entropy are extracted for obtaining the intensity characteristics.

\subsection{Graph-based Feature Extraction}
\label{Sec:5.3}

In\cite{Kuo-2014-Lymphatic}, Voronoi diagram, Delaunay triangulation and minimum spanning tree are constructed for extracting the arrangement information of lymphocyte nuclei because lymphocyte detection is not enough to describe the abnormalities lymphatic infiltration. Voronoi diagram is constructed using the segmented objects to produce a tessellation for an image. Delaunay triangulation is constructed through connecting the centroid of the adjacent diagram of the Voronoi diagram. Suppose there is a connected graph $G$, $G$ may have multiple spanning trees, and these spanning trees have all the vertices of $G$. When traversing all the vertices of these spanning trees, the spanning tree with the least cost is the minimum spanning tree. 

In\cite{Basavanhally-2008-Manifold}, a method combining manifold learning with graph-based features, which is used to identify lymphocytic infiltration in histopathology images of breast cancer. Graph-based features extracted in the method include Voronoi diagram, Delaunay triangulation, minimum spanning tree, nuclear features and Varma-zisserman texton-based features. Manifold learning is used for feature dimensionality reduction. The dataset includes 22, 10 and 9 cases of high, medium and low lymphocytic infiltration. SVM is used to classification through 3-fold cross-validation and an accuracy equalling 89.50$\%$ $\pm$ 6.22 $\%$ is obtained.

\begin{figure}[!htbp]
\centering
\centerline{\includegraphics[width=0.8\textwidth]{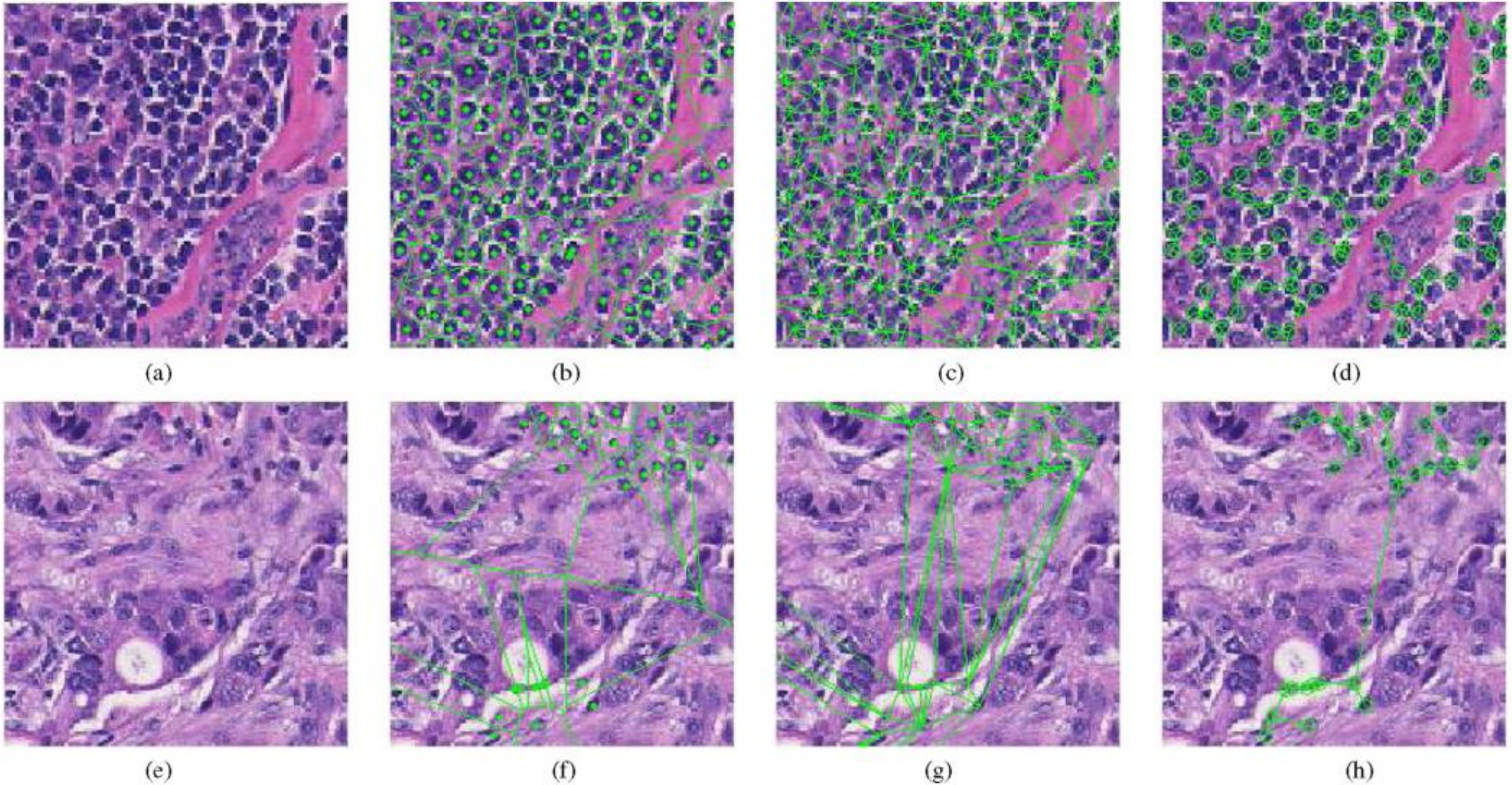}}
\caption{(a) and (e) stand for high and low levels of lymphocytic infiltration. (b) and (f) stand for corresponding Voronoi diagrams. (c) and (g) stand for corresponding Delaunay triangulations. (d) and (h) stand for corresponding minimum spanning trees. The figure corresponds to Fig. 7 in~\cite{Basavanhally-2009-Computerized}.}
\label{fig:graph-based features}
\end{figure}

In\cite{Basavanhally-2009-Computerized}, a method for detecting and grading lymphocytic infiltration of breast cancer is proposed. a method for detecting and grading lymphocytic infiltration of breast cancer is proposed. First, the lymphocytes are detected by region growing algorithm and Markov random field algorithm. Second, 50 features are extracted from three graphs including the Voronoi diagram, Delaunay triangulation and minimum spanning tree (as shown in Fig.~\ref{fig:graph-based features}). The three graphs are constructed with the center of single detected lymphocytes as vertices. Next, the high-dimensional feature vector is reduced through graph embedding. Finally, the method is tested on a  dataset including 41 H$\%$E stained images. SVM is applied to classification through more than 100 3-fold cross validation and a classification accuracy that is higher than 90$\%$ is obtained.

In\cite{Oztan-2012-Follicular}, a method is proposed for grading FL images, which is based on a multi-scale feature analysis. The cell graph is used to describe the structural tissue of cell and represents the histological image of undirected and unweighted graphs. The graph nodes are composed of cytological components and the graph edges consist of approximate adjacencies. The features extracted from cell graphs are used for classification. Three classifiers including SVM, Bayesian and $k$-nearest-neighbors ($k$NN) are selected for classification and the dataset includes 510 FL images. The result shows that when the cell graph features are combined with MBIR and low-level features, the accuracy is higher.

In\cite{Ishikawa-2014-Gastric}, a method for automatically detecting cancer in H$\&$E stained images of the gastric lymph nodes is proposed. In this method, three features are combined, which include Higher-order Local  Auto-Correlation (HLAC) feature, wavelet feature, Delaunay feature. These features are connected into one vector after being calculating in images, and then SVM is used as a classifier to detect cancer. The dataset includes 19 training images and 381 testing images. In addition, each image is divided into 120 patches. Finally, the best results of sensitivity and specificity are 95.7$\%$ and 82.1$\%$.

\subsection{Deep Learning Feature Extraction}
\label{Sec:5.4}

In addition to the traditional methods of low-level feature extraction described above, some advanced deep learning methods have been widely used in LHIA in recent years. There is no need to manually extract features in deep neural networks, and better features can be automatically extracted. A neural network extracts visual features which can be overlooked by human visual inspection. The convolutions in the neural network can filter a large number of features contained in an image. For example, in the feature extraction of edge detection, convolutions can help determine the number of nuclei with abnormal shapes and the density of nuclei neighborhoods.

\begin{figure}[!htbp]
\centering
\centerline{\includegraphics[width=1\textwidth]{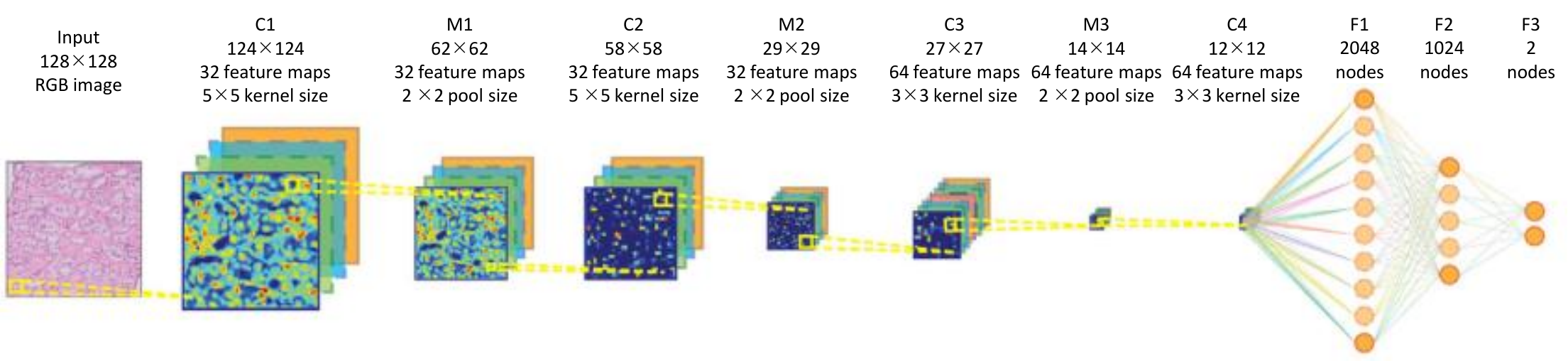}}
\caption{The figure corresponds to Figure 1 in~\cite{Litjens-2016-Deep}.}
\label{fig:CNN feature extraction}
\end{figure}

In\cite{Litjens-2016-Deep}, CNN replaces traditional methods for feature extraction. Traditional methods rely on hand-made quantitative feature extractors for feature extraction, while CNN can automatically extract image features. Fig.~\ref{fig:CNN feature extraction} shows a CNN model. C1, C2, C3, and C4 are convolutional layers for feature extraction through continuously extracting higher-level features from the image patch. In~\cite{Mohlman-2020-Improving}, the CNNs through 12 initial feature filters obtain a network containing 258,327 parameters. The parameters.

In\cite{Sheng-2020-Blood}, Faster R-CNN is used in classifying lymphoma. The input image is transformed into a tensor, then the tensor is sent into a CNN which is pre-trained or initialized to generate a feature map. Then, the region proposal network is used to calculate the feature to obtain a region proposal containing the target. A new tensor is obtained by using the region of interest pooling, and the tensor is put into the fully connected layers for classification.

In\cite{Hashimoto-2020-Multi}, a new method based on CNN is proposed, which is combined with multiple instances, domain adversarial and multi-scale. The method automatically detects tumor-specific features. In the CNN network, the feature extractor is obtained by training a single-scale DA-MIL network, then the multi-scale DA-MIL network is trained through plugging the feature extractor into the CNN network.

In\cite{Roberto-2021-Fractal}, an ensemble model is proposed, which is based on handcrafted fractal features and deep learning. 300 fractal features are obtained through features extraction. These features compose an image after being reshaped into a 10*10*3 matrix, and the image is the input to the CNN.

In\cite{Dif-2021-Transfer}, the convolutional layer of CNN extracts features, the pooling layer reduces dimension. A hierarchical feature extraction module is generated by stacking the convolutional and pooling layer. The feature map is reshaped into a 3D matrix and is flattened to form a vector for feature extraction.

\subsection{Other Features Extraction}
\label{Sec:5.5}

\begin{figure}[!htbp]
\centering
\centerline{\includegraphics[width=0.7\textwidth]{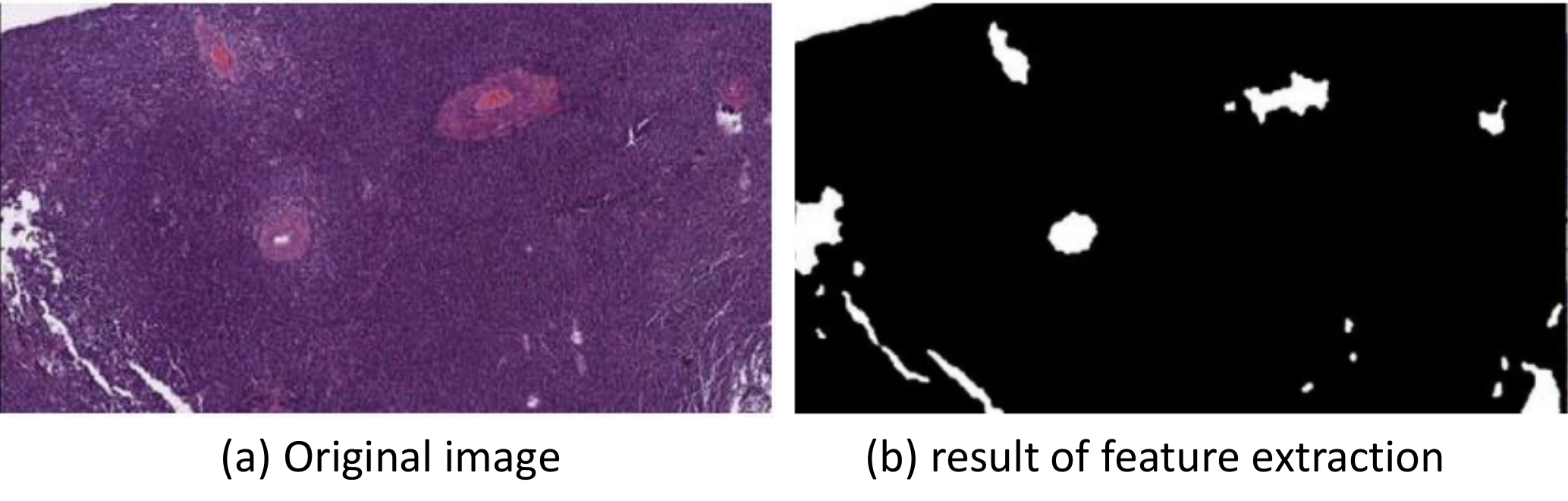}}
\caption{The result of high-level feature extraction. The figure corresponds to Fig. 4 in~\cite{Cooper-2009-Feature}.}
\label{fig:matching high level}
\end{figure}

In\cite{Cooper-2009-Feature}, a method of automatic non-rigid registration is proposed, which is used in different staining histopathology images. This method extracts matching high level features on the down-sampled images of original images. Fig.~\ref{fig:matching high level} shows the result of feature extraction. Then, the match candidates are formed by matching features between the base images and float images. The dataset in this experiment includes 5 pairs of FL slides, one is stained with CD3 and another is stained with H$\&$E in each pair. The statistical result demonstrates that the accuracy of this method is equivalent to that of manual registration.

In\cite{Orlov-2010-Automatic}, a two-stage method is used to classify three types of lymphoma: CLL, FL and MCL. In the outer level, Fourier, Chebyshev and wavelets, Chebyshev of Fourier and wavelets of Fourier are calculated. The set of global features is calculated in the inner level, then the calculated features are fused into a feature vector.

In\cite{Michail-2014-Detection}, a method is proposed, which is developed for automatically detecting malignant cells in microscopic images of FL. The cell nuclei are first detected after preprocessing and segmentation, next the expectation maximization algorithm is used to split touching cells. Then, the candidate CBs are selected by their size, shape and intensity histogram. Then, in the step of feature extraction, the image consisting of cell and its surroundings is used as a feature vector for classifying CBs and non-CBs. The data set including 70 CBs images and 110 non-CBs images is divided into training set and validation set at 80$\%$ to 20$\%$. A LDA classifier is applied for classification, and average 82.58$\%$ of malignant cells are detected.

\subsection{Summary}
\label{Sec:5.6}

This section has described feature extraction methods, including traditional and deep learning-based methods. In traditional methods, the extracted features are generally color features, texture features and shape features, and the commonly used traditional methods include the conversion of color space, GLCM, co-occurrence matrix, LBP, Gabor filter, SIFT and so on. However, advanced algorithms are researched and applied in LHIA with the development of MV. Deep learning is now being studied by more and more people, especially in the field of MV, and it has also shown good performance. With the advancement of deep learning technology, feature extraction is no longer limited to manual extraction because deep neural networks can automatically extract better features. 
 
\begin{table}[!htbp]
\renewcommand\arraystretch{2}
\centering
\caption{Feature Extraction Methods.}
\label{Tab:Feature Extraction Methods1}
\tiny
\begin{tabular}{cccc}
\hline
Reference & Year & Team & Details\\
\hline

\cite{Zorman-2007-Symbol} & 2007 & M Zorman & R component of the RGB image\\

\cite{Sertel-2008-Texture} & 2008 & O Sertel & \tabincell{l}{Color Texture Analysis Using SOFM, \\the gray level co-occurrence features}\\

\cite{Basavanhally-2008-Manifold} & 2008 & A Basavanhally & Graph-Based Architectural Features\\

\cite{Yang-2008-Automatic} & 2008 & L Yang & contour\\

\cite{Sertel-2008-Computerized} & 2008 & O Sertel & color, texture, morphological features\\

\cite{Cooper-2009-Feature} & 2009 & L Cooper & matching high level features\\

\cite{Basavanhally-2009-Computerized} & 2009 & AN Basavanhally & Graph-based architectural\\

\cite{Belkacem-2009-Extraction} & 2009 & K Belkacem-Boussaid & geometrical features 
vector and texture color features\\

\cite{Sertel-2009-Histopathological} & 2009 & O Sertel & morphological and topological features, color, texture\\

\cite{Belkacem-2010-Computer} & 2010 & K Belkacem-Boussaid & Geometric and color texture features\\

\cite{Sertel-2010-Image} & 2010 & O Sertel & $L^*u^*v^*$\\

\cite{Cheng-2010-Identifying} & 2010 & J Cheng & \tabincell{l}{49 Zernike features, 30 Daubechies 4 wavelet features, \\60 Gabor features, 5 skeleton features, \\13 Haralick features and 16 morphological features}\\

\cite{Orlov-2010-Automatic} & 2010 & NV Orlov & Transform-based global features\\

\cite{Sertel-2010-Computer} & 2010 & O Sertel & gray-level run length matrix (GLRLM)\\

\cite{Samsi-2010-Detection} & 2010 & S Samsi & texture and color\\

\cite{Belkacem-2010-Effect} & 2010 & K Belkacem-Boussaid & Morphometric and color texture features\\

\cite{Fatakdawala-2010-Expectation} & 2010 & H Fatakdawala & texture \\

\cite{Meng-2010-Histology} & 2010 & T Meng & color and texture feature, C-RSPM model\\

\cite{Han-2010-Multi} & 2010 & J Han & color and texture feature\\

\cite{Kong-2011-Partitioning} & 2011 & H Kong & color texture feature\\

\cite{Zorman-2011-Classification} & 2011 & M Zorman & R component of the RGB image\\

\cite{Oger-2012-General} & 2012 & M Oger & \tabincell{l}{Each pixel is represented by a feature \\vector of color and texture information.}\\

\cite{Akakin-2012-Content} & 2012 & HC Akakin & color, texture\\

\cite{Samsi-2012-Efficient} & 2012 & S Samsi & color, gray level feature, texture\\

\cite{Oztan-2012-Follicular} & 2012 & B Oztan & \tabincell{l}{the combination of cell-graph based features \\with the MBIR and color-textural features}\\

\cite{Arora-2013-Computer} & 2013 & B Arora & no of centers, length of curvature\\

\cite{Sandhya-2013-Automated} & 2013 & BS Sandhya & morphological and texture features\\

\cite{Meng-2013-Multimodal} & 2013 & T Meng & A set of 505 visual features\\

\cite{Acar-2013-Tensor} & 2013 & E Acar & 72 color and texture features\\

\cite{Ishikawa-2014-Gastric} & 2014 & T Ishikawa & HLAC feature, Wavelet feature, Delaunay feature\\

\cite{Michail-2014-Detection} & 2014 & E Mchail & \tabincell{l}{The whole image of the cell with its \\surroundings are used as a feature vector}\\

\hline
\end{tabular}
\end{table}

\begin{table}[!htbp]
\renewcommand\arraystretch{2}
\centering
\caption{Continue: Feature Extraction Methods.}
\label{Tab:Feature Extraction Methods2}
\tiny
\begin{tabular}{cccc}
\hline
Reference & Year & Team & Details\\
\hline

\cite{Michail-2014-Morphological} & 2014 & E Mchail & morphological and textural features\\

\cite{Kuo-2014-Lymphatic} & 2014 & YL Kuo & Scale-invariant feature transform\\ 

\cite{Zarella-2015-Lymph} & 2015 & MD Zarella & color-based feature, shape-based feature\\

\cite{Fauzi-2015-Classification} & 2015 & MFA Fauzi & 64-dimensional color histogram features\\

\cite{Di-2015-Different} & 2015 & C Di Ruberto & GLCM features, GLRLM features, GLDM features\\

\cite{Chen-2016-Automatic} & 2016 & J Chen & Extension of FCN\\

\cite{Song-2016-Histopathology} & 2016 & Y Song & \tabincell{l}{the local DSIFT features and patch-level features that are \\learned using a deep belief network (DBN) model}\\

\cite{Wang-2016-Deep} & 2016 & D Wang & geometrical and morphological features\\

\cite{Chen-2016-Identifying} & 2016 & R Chen & geometrical and morphological features\\

\cite{Shi-2016-Automated} & 2016 & P Shi & \tabincell{l}{Color intensities, mean and standard deviation \\value of pixels, local texture features}\\

\cite{Codella-2016-Lymphoma} & 2016 & N Codella & CNN\\

\cite{Litjens-2016-Deep} & 2016 & G Litjens & CNN\\

\cite{Xiao-2017-Deep} & 2017 & K Xiao & CNN\\

\cite{Roberto-2017-Features} & 2017 & GF Roberto & percolation theory\\

\cite{Brancati-2019-Deep} & 2019 & N Brancati & Supervised Encoder FusionNet\\

\cite{Somaratne-2019-Improving} & 2019 & UV Somaratne & A model using transfer learning with fine-tuning\\

\cite{Li-2020-Deep} & 2020 & D Li & GOTDP-MP-CNNs\\

\cite{Sheng-2020-Blood} & 2020 & B Sheng & faster R-CNN\\

\cite{Mohlman-2020-Improving} & 2020 & JS Mohlman & CNNs\\

\cite{Hashimoto-2020-Multi} & 2020 & N Hashimoto & Multi-scale domain-adversarial multiple-instance CNN\\

\cite{Bianconi-2020-Experimental} & 2020 & F Bianconi & 13 image descriptors\\

\cite{Kandel-2020-Novel} & 2020 & I Kandel & CNN\\

\cite{Syrykh-2020-Accurate} & 2020 & C Syrykh & BNN\\

\cite{Nugaliyadde-2020-Rcnn} & 2020 & A Nugaliyadde & RCNN\\

\cite{Candelero-2020-Selection} & 2020 & D Candelero & Evolutionary Algorithms\\

\cite{Martins-2021-Hermite} & 2021 & AS Martins & A Hermite polynomial algorithm\\

\cite{Azevedo-2021-Evaluation} & 2021 & TA Azevedo Tosta & \tabincell{l}{Statistical and Haralick’s features extracted \\from wavelet and ranklet transforms}\\

\cite{Roberto-2021-Fractal} & 2021 &GF Roberto & An ensemble model based on handcrafted fractal features and deep learning\\

\cite{Dif-2021-Transfer} & 2021 & N Dif & Transfer learning from synthetic labels\\

\hline
\end{tabular}
\end{table}

\section{Classification Methods}
\label{Sec:6}

In this section, we introduce the classification tasks in LHIA. After feature extraction, different classifiers are selected to classify the histopathological images. In LHIA, common classification tasks include staging and grading malignant lymphomas (such as FL), classification of several malignant lymphomas (such as CLL, MCL and FL) and classification of CB and non-CB. The methods of classification include traditional ML methods and deep learning-based methods. 
 
\subsection{Traditional Machine Learning based Classification Methods}

In\cite{Basavanhally-2009-Computerized}, SVM is applied to discriminate high and low lymphocytic infiltration samples. In~\cite{Akakin-2012-Content}, SVM is used to classify FL and neuroblastoma, and the radial basis function (RBF) is selected in SVM. In~\cite{Oztan-2012-Follicular, Michail-2014-Morphological}, SVM is used to grade FL. In~\cite{Sandhya-2013-Automated}, SVM is used to classify CBs and non-CBs. The steps of preprocessing and feature extraction are described in Sec.~\ref{Sec:3} and Sec.~\ref{Sec:5}. Finally, an accuracy of 84.53$\%$ is obtained in 70 training datasets and 30 test datasets. In~\cite{Di-2015-Different}, SVM is used for classification and validates the texture features extracted from the HSV color space are superior to the features of other color spaces. In~\cite{Shi-2016-Automated, Song-2017-Low}, SVM is used to classify three lymphomas (CLL, FL and MCL). In~\cite{Tosta-2017-Application}, SVM is used to classify FL and MCL.

In\cite{Acar-2013-Tensor}, a framework for grading FL is developed. In the framework, the image is represented by a third-order tensor including images, features and scales. The basic patterns of each mode can be extracted from these tensor models. A total of 500 FL images are analyzed through SVM classifier and the accuracy can reach up to 90$\%$.

In\cite{Sertel-2008-Texture, Sertel-2009-Histopathological, Oztan-2012-Follicular}, Bayesian classifier is selected to grade FL. In~\cite{Dimitropoulos-2017-Automated}, a Bayesian classifier is applied to classify CBs in FL tissue samples. In\cite{Sertel-2009-Histopathological, Oztan-2012-Follicular, Fauzi-2015-Classification}, $k$NN classifier is selected to classify FL into three histological grades. In\cite{Belkacem-2009-Extraction, Belkacem-2010-Computer, Belkacem-2010-Effect}, QDA is selected to classify CBs and non-CBs.

Weighted-Neighbor Distance (WND), which relies on the distance of sample to class. The training process of the Naive Bayes Network (BBN) classifier is to estimate the class prior probability based on the training set, and estimate the conditional probability for each attribute. RBF is a real-valued function whose value depends only on the distance from the origin. In\cite{Orlov-2010-Automatic}, a method is proposed to classify CLL, FL and MCL. WND, BBN and RBF are used as classifiers after feature extraction. In the training set, there are 57 images for each type of lymphoma. In the test set, CLL, FL and MCL include 56, 82 and 65 images respectively. Finally, an average classification rate equalling 99$\%$ is obtained by WND.

In\cite{Meng-2010-Histology}, a framework is proposed for improving the classification accuracy, which is based on the Collateral Representative Subspace Projection Modeling (C-RSPM) classifier. Each classifier of C-RSPM uses the known-class data consisting of different types in training dataset~\cite{Quirino-2006-Collateral}. In this framework, the original image is divided into 25 blocks, and one C-RSPM is constructed in each block. A multimodal fusion algorithm is used to determine the final class label after classification. IICBU-2008 dataset is divided into training set and test set at a ratio of 2 to 1. IICBU-2008 dataset is divided into the training set and test set at a ratio of 2 to 1. In the experiment, an average classification accuracy of 92.7$\%$ is obtained, while the highest average classification accuracy previously reported using the same data set is 85$\%$. In~\cite{Meng-2013-Multimodal}, the steps of preprocessing and feature extraction are similar to~\cite{Meng-2010-Histology}, and described in Sec.~\ref{Sec:3.6} and Sec.~\ref{Sec:5}. The C-RSPM classifier is used for classification through three-fold cross-validation. The IICBU-2008 dataset is used in the experiment and the highest classification accuracy of 96.81$\%$ is obtained compared with other classifiers including SVM, BayesNet and BBN, etc.

In\cite{Zorman-2007-Symbol, Zorman-2011-Classification}, symbol-based ML methods are used to classify FL. In\cite{Schmitz-2012-Automated}, a minimum distance classifier is used to classify pixels, which is a supervised approach. In\cite{Ribeiro-2018-Analysis}, a random forest classifier is used to classify three NHL into CLL, FL and MCL.

\begin{table}[!htbp]
\renewcommand\arraystretch{2}
\centering
\caption{Traditional Classification Methods.}
\label{Tab:Traditional Classification Methods}
\begin{tabular}{cccc}
\hline
Reference & Year & Team & Details\\
\hline
\cite{Sertel-2008-Texture} & 2008 & O Sertel & Bayesian classifier\\

\cite{Basavanhally-2009-Computerized} & 2009 & AN Basavanhally & SVM\\

\cite{Belkacem-2009-Extraction} & 2009 & K Belkacem-Boussaid & QDA\\

\cite{Sertel-2009-Histopathological} & 2009 & O Sertel & Bayesian and $k$NN\\

\cite{Belkacem-2010-Computer} & 2010 & K Belkacem-Boussaid & QDA\\

\cite{Orlov-2010-Automatic} & 2010 & NV Orlov & WND, RBF and BBN\\

\cite{Belkacem-2010-Effect} & 2010 & K Belkacem-Boussaid & QDA\\

\cite{Meng-2010-Histology} & 2010 & T Meng & C-RSPM\\

\cite{Schmitz-2012-Automated} & 2012 & A Schmitz & minimum distance classifier\\

\cite{Akakin-2012-Content} & 2012 & HC Akakin & SVM\\ 

\cite{Oztan-2012-Follicular} & 2012 & B Oztan & SVM, Bayesian, $k$NN\\

\cite{Sandhya-2013-Automated} & 2013 & BS Sandhya & SVM\\

\cite{Acar-2013-Tensor} & 2013 & E Acar & SVM\\

\cite{Michail-2014-Morphological} & 2014 & E Mchail & SVM\\

\cite{Fauzi-2015-Classification} & 2015 & MFA Fauzi & $k$NN\\

\cite{Di-2015-Different} & 2015 & C Di Ruberto & SVM\\

\cite{Song-2016-Histopathology} & 2016 & Y Song & SVM\\

\cite{Shi-2016-Automated} & 2016 & P Shi & SVM\\

\cite{Tosta-2017-Application} & 2017 & TAA Tosta & SVM\\

\cite{Song-2017-Low} & 2017 & Y Song & SVM\\

\cite{Dimitropoulos-2017-Automated} & 2017 & K Dimitropoulos & Bayesian\\

\cite{Ribeiro-2018-Analysis} & 2018 & MG Ribeiro & Random forest\\

\cite{Tosta-2018-Fitness} & 2018 & TAA Tosta & SVM\\
\hline
\end{tabular}
\end{table}

\subsection{Deep Learning based Classification Methods}

\begin{figure}[!htbp]
\centering
\centerline{\includegraphics[width=0.8\textwidth]{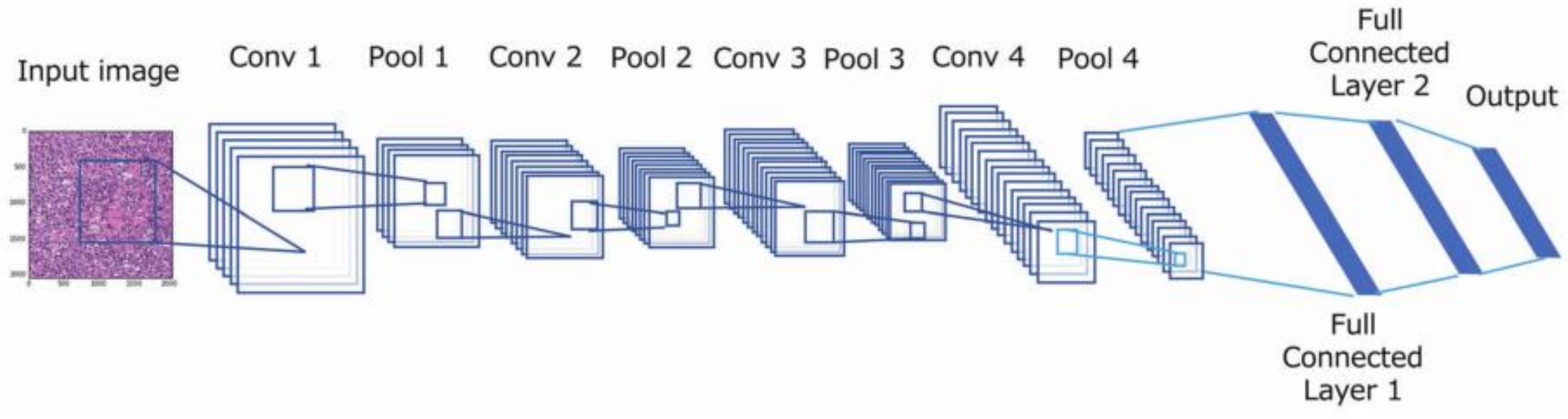}}
\caption{The architecture of the deep neural network. The figure corresponds to Fig. 4 in~\cite{Miyoshi-2020-Deep}.}
\label{fig:deep neural network}
\end{figure}

In\cite{Arora-2013-Computer}, a method is proposed to grade FL. The step of segmentation is described in Sec.~\ref{Sec:4.2}, two features are fed a classifier which is composed of a hidden layer with 10 neurons. In\cite{Miyoshi-2020-Deep}, a deep neural network is used to classify lymphomas including DCBCL, FL and reactive lymphoid hyperplasia. The proposed network includes 11 layers containing four convolutional and two fully connected layers (as shown in Fig.~\ref{fig:deep neural network}). 388 sections are prepared into WSIs which are divided into patches with different magnifications ($\times$5, $\times$20 and $\times$40). Finally, the deep neural network wins pathologists with an accuracy of 97$\%$ which demonstrates the method is potential in diagnosing malignant lymphoma.

\begin{figure}[!htbp]
\centering
\centerline{\includegraphics[width=0.8\textwidth]{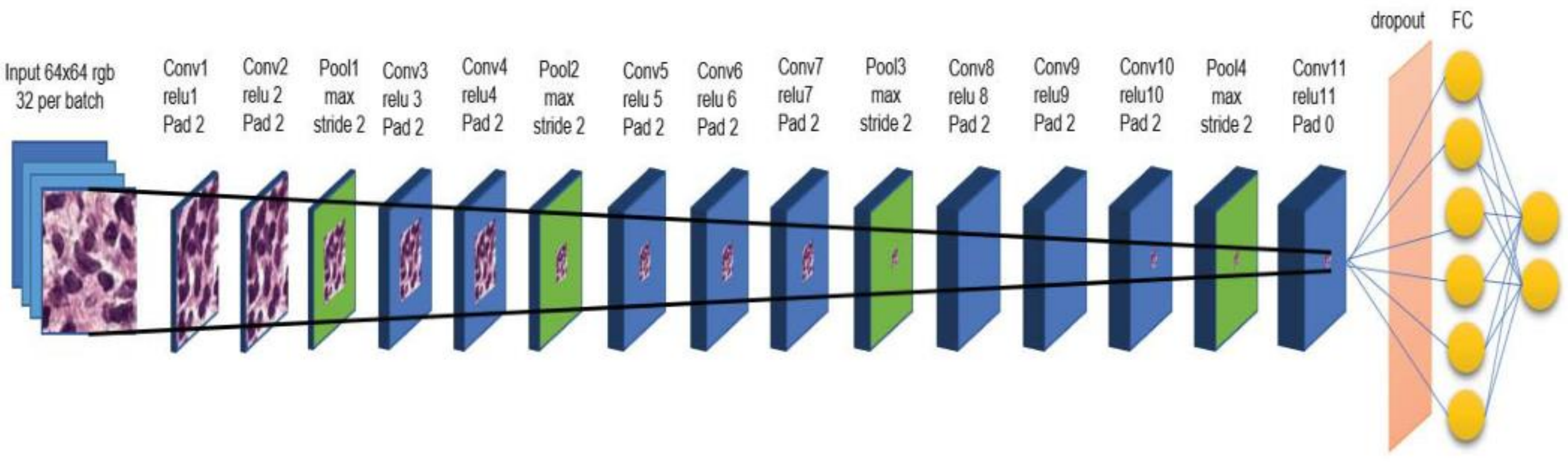}}
\caption{The architecture of the proposed deep learning model. The figure corresponds to Fig. 3 in~\cite{Jamaluddin-2017-Tumor}.}
\label{fig:model based on LeNet}
\end{figure}

In\cite{Jamaluddin-2017-Tumor}, a deep learning model based on CNN is developed to classify normal and tumor slides in images of Camelyon16. The proposed deep learning model has 12 convolutional layers and max pooling, and the size of inputs image is 64$\times$64$\times$3. The architecture of the model is shown in Fig.~\ref{fig:model based on LeNet}. The masks of tumor region are generated through the model, then a random forest classifier is utilized for classification. Finally, an AUC of 0.94 is obtained, which surpasses the result of the winner of the Camelyon16 Challenge.

In\cite{El-2019-Automated}, a CNN model is applied to classify four lymphomas including benign lymph node, DLBCL, Burkitt lymphoma and small lymphocytic lymphoma. In\cite{Wollmann-2018-Adversarial}, a new method is proposed for four-classification in dataset Camelyon17. The method combines two networks including Cycle-Consistent Generative Adversarial Network (CycleGAN) and a densely connected deep neural network.

In\cite{Zhu-2019-Novel}, a classification framework is developed, which is based on the method of multi-space image reconstruction and technology of transfer learning. The classification results can be more accurate through the auxiliary information obtained by the method of multi-space reconstruction. The VGG16 is applied to reconstruct images and then the long short-term memory (LSTM) is used for the selection and refinement of features. The softmax classifier is utilized in the classification task. IICBU-2008 lymphoma dataset is applied in the framework, and each image is divided into 336 patches with 64$\times$64 pixels. Finally, an ACC of 98.94$\%$ is obtained, which is higher than other results of\cite{Shamir-2008-Wndchrm, Meng-2010-Histology, Codella-2016-Lymphoma}.

In\cite{Bai-2019-Nhl}, a classification model is developed. The first step is the conversion of color space (as described in Sec.~\ref{Sec:3.1}), the original images are converted into blue ratio and $L^*a^*b$ spaces. Then, texture and statistical features are extracted in the patches of blue ratio images and color features are extracted from $L^*a^*b$ color space. These features are processed to obtain hand-crafted representations and then a random forest classifier is applied for classification. Next, the images are cropped and used as input of a google inception net, and the softmax classifier is used in classification. The IICBU-2008 dataset is applied in the model and an overall ACC equalling 0.991 and an AUC equalling 0.9978 are obtained. The results show the model is an efficient classification method of NHL images.

In\cite{Brancati-2019-Deep}, a residual CNN is used in the classification of lymphoma subtypes. Residual CNN is compared with other deep neural networks including U-net and ResNet. The IICBU-2008 dataset is used in the method and an ACC of 97.67$\%$ is obtained. The result is higher than the results of U-net and ResNet, this shows that the method is efficient.

In\cite{Li-2020-Deep}, an transfer learning platform with 17 CNNs is designed to improve the diagnosis of DLBCL. The dataset is collected from three hospitals: 500 DLBCL images and 505 non-DLBCL images are from hospital A, 1467 square images of DLBCL and 1656 square images of non-DLBCL are from hospital B, 204 DLBCL images and 198 non-DLBCL images are from hospital C. 80$\%$ of the dataset are used as the training set, 10$\%$ are used as the validation set and the last 10$\%$ are used as the testing set. Finally, a diagnostic rate close to 100$\%$ is obtained and the result shows that the platform has the efficiency of clinical diagnosis.

In\cite{Sheng-2020-Blood}, faster Region-based CNNs (R-CNNs) are used to classify different images of lymphomas. In the process of faster R-CNN, the input image is first represented as a tensor, then the tensor is used as the input of a pre-trained CNN for the purpose of generating a feature map. Next, the region proposal network is used to compute the feature for obtaining the region proposal network. The region network may contain the target. The dataset includes images of lymphocytes, lymphoma cells and blasts and is divided into 60$\%$ training set, 20$\%$ validation set and 20$\%$ test set. Finally, the result of detection rate higher than 96$\%$ is obtained which shows the method has a potential in the diagnosis of lymphoma.

\begin{figure}[!htbp]
\centering
\centerline{\includegraphics[width=0.8\textwidth]{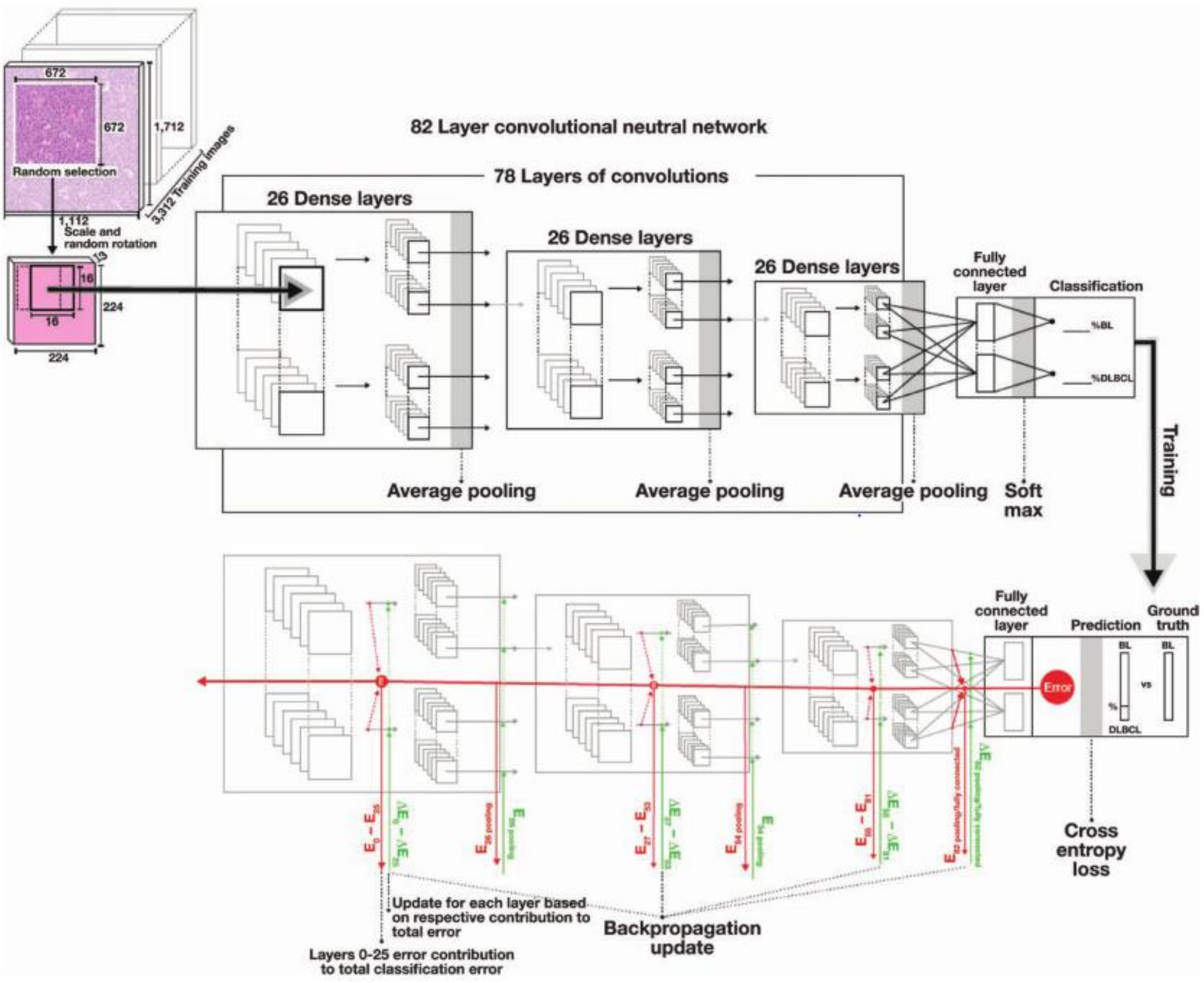}}
\caption{The architecture of the proposed CNN. The figure corresponds to Figure 2 in~\cite{Mohlman-2020-Improving}.}
\label{fig:BL and DLBCL}
\end{figure}

In\cite{Mohlman-2020-Improving}, a CNN is applied to classify Burkitt lymphoma and DLBCL. The architecture of the CNN is shown in Fig.~\ref{fig:BL and DLBCL}, there are 78 convolutional layers that are divided into three equal modules. A total of 10818 images of BL and DLBCL are collected in the experiment, an AUC of 0.92 and 94$\%$ cases are correctly classified. The results show that the proposed CNN is a promising tool for classifying BL and DLBCL.

In\cite{Kandel-2020-Novel}, a new CNN is proposed to classify histopathology images. The CNN model is comprised of 15 convolutional and two fully connected layers. The PCam dataset is used for testing the CNN and an AUC of 95.46$\%$ is obtained.

In\cite{Hashimoto-2020-Multi}, a novel CNN is proposed to classify the malignant lymphomas subtype. The CNN combines some frameworks including multiple-instance, multi-scale and domain adversarial. 196 clinical samples of malignant lymphoma are used in the method, and the result of ACC higher than 0.871 wins other methods. The result shows that the method is efficient in classifying malignant lymphomas.

\begin{table}[!htbp]
\renewcommand\arraystretch{2}
\centering
\caption{Deep Learning Methods for Classification.}
\label{Tab:Deep Learning Methods for Classification}
\begin{tabular}{cccc}
\hline
Reference & Year & Team & Details\\
\hline

\cite{Arora-2013-Computer} & 2013 & B Arora & A hidden layer with 10 neurons\\

\cite{Jamaluddin-2017-Tumor} & 2017 & MF Jamaluddin & Deep learning model based on LeNet\\

\cite{Wollmann-2018-Adversarial} & 2018 & T Wollmann & CycleGAN and desely connected deep neural network\\

\cite{El-2019-Automated} & 2019 & H El Achi & CNN\\

\cite{Zhu-2019-Novel} & 2019 & H Zhu & VGG16\\

\cite{Bai-2019-Nhl} & 2019 & J Bai & google inception net\\

\cite{Brancati-2019-Deep} & 2019 & N Brancati & Residual CNN\\

\cite{Li-2020-Deep} & 2019 & D Li & Transfer learning platform with 17 CNNs\\

\cite{Sheng-2020-Blood} & 2020 & B Sheng & Faster R-CNN\\

\cite{Kandel-2020-Novel} & 2020 & I Kandel & CNN\\

\cite{Hashimoto-2020-Multi} & 2020 & N Hashimoto & CNN\\
\hline
\end{tabular}
\end{table}

\subsection{Summary}

From the reviewed works, SVM and KNN are the commonly used traditional classification methods in LHIA. Deep learning classification methods are the mainstream methods and achieve good results, and the CNN network is the most commonly used deep learning method.

\section{Detection Methods}
\label{Sec:7}

In this section, we introduce the detection tasks in LHIA. The detection tasks of LHIA include the detection of malignant cells, CB, follicles and lymphocytic infiltration. In many cases, the detection tasks are carried out through classification methods. 

\subsection{Traditional Machine Learning based Detection Methods}

In\cite{Basavanhally-2008-Manifold}, SVM is used to identify lymphocytic infiltration in breast cancer. In~\cite{Ishikawa-2014-Gastric}, a method for automatically detecting cancer in the gastric lymph node is proposed (as described in Sec.~\ref{Sec:5.3}). SVM is used as a classifier for classification. In~\cite{Zarella-2015-Lymph}, SVM successfully predicts the status of lymph node metastasis in breast carcinoma, 101 patients participate in the experiment and a correct prediction rate equalling 88.4$\%$ is obtained.

In\cite{Kuo-2014-Lymphatic}, a method is developed, which is applied for automatically detecting lymphocytic infiltration in H$\&$E stained images of breast cancer. The steps of preprocessing and feature extraction are described in Sec.~\ref{Sec:3.6} and Sec.~\ref{Sec:5.1}. The dataset includes 47 slides (training set includes 10 slides, validation set includes 37 slides), 40 slides are randomly selected for testing. SVM is applied for classification, and the highest accuracy of 93.23$\%$ is obtained.

\begin{figure}[!htbp]
\centering
\centerline{\includegraphics[width=0.7\textwidth]{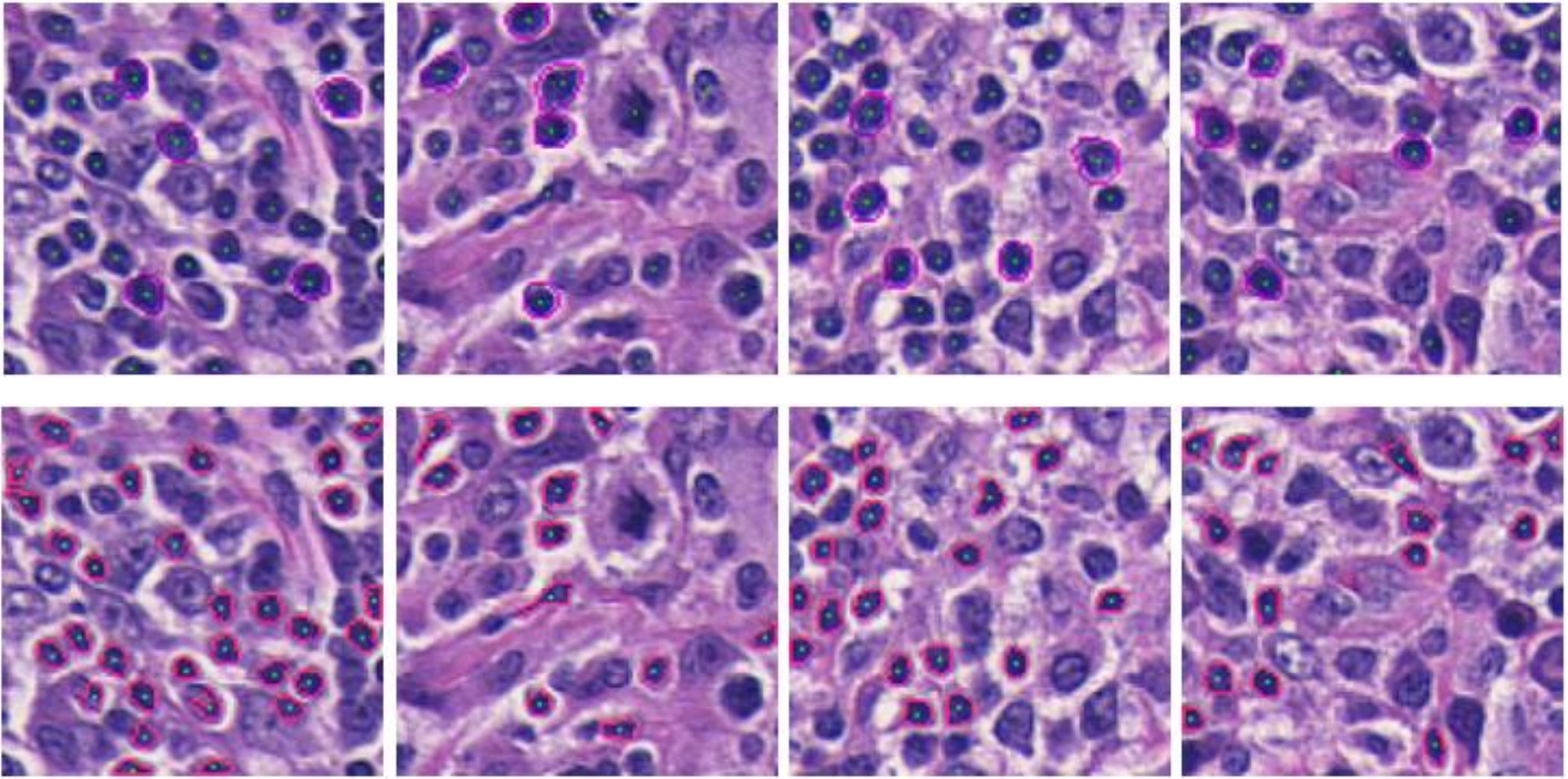}}
\caption{First row: manually annotations. Second row: lymphocytes identified by the framework. The figure corresponds to Fig. 2 in~\cite{Cheng-2010-Identifying}.}
\label{fig:lymphocytes in breast cancer}
\end{figure}

\begin{figure}[!htbp]
\centering
\centerline{\includegraphics[width=0.7\textwidth]{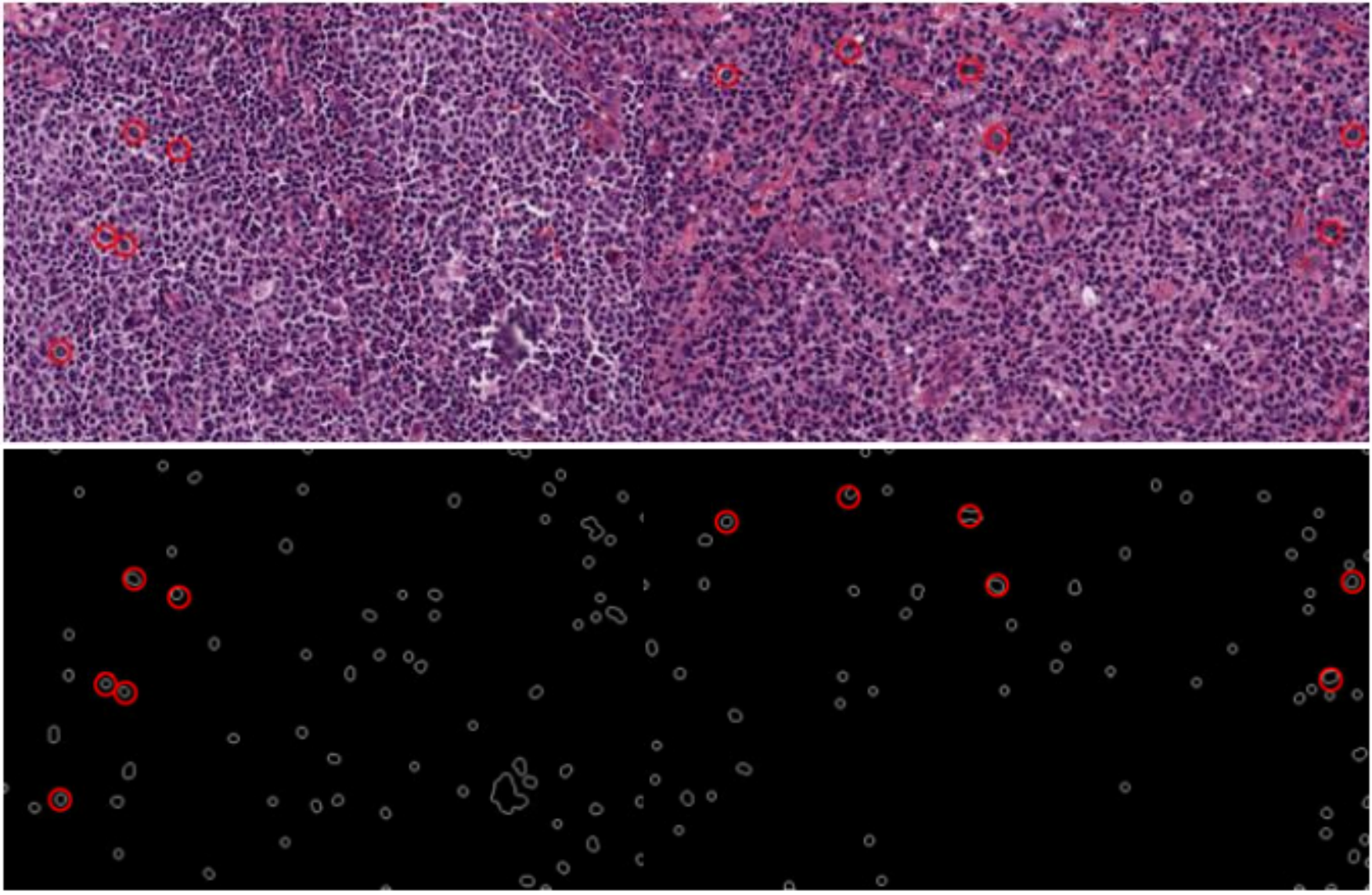}}
\caption{First row: manually annotations. Second row: The CBs identified by the framework. The figure corresponds to Fig. 3 in~\cite{Cheng-2010-Identifying}.}
\label{fig:CBs in FL}
\end{figure}

In\cite{Cheng-2010-Identifying}, a framework is proposed to identify lymphocytes and CBs. SVM-RFE is selected for classification after the steps of segmentation, feature extraction and feature selection. SVM-RFE uses SVM as a classifier and recursively eliminates features that are irrelevant. Two datasets are used, the first dataset includes 10 H$\&$E stained breast cancer images, the second dataset includes 5 H$\&$E stained FL images. Ten trials are performed through five-fold cross-validation in the experiment, lymphocytes are identified in the first dataset and CBs are identified in the second dataset. Finally, a series of results (Dice = 0.73, Overlap (equals to JAC) = 0.57, SEN = 0.57, SPE = 1, PPV = 1, HD = 4.58, MAD = 0.77) are obtained in the first dataset, and the results of SEN = 0.38, FPR = 82.85 are obtained in the second dataset. The results show that the framework has a good performance in identifying lymphocytes and CBs. Fig.~\ref{fig:lymphocytes in breast cancer} shows the lymphocytes identified in breast cancer images. Fig.~\ref{fig:CBs in FL} shows CBs identified in FL images.

In\cite{Michail-2014-Detection}, a method for automatically detecting malignant cells in FL images is proposed (as described in Sec.~\ref{Sec:5.5}). The LDA classifier is used for classification and an average of 82.58$\%$ of malignant cells are detected. In\cite{Basavanhally-2009-Computerized}, lymphocytes are detected by an algorithm that combines region growing and Markov random field (MRF).

\begin{table}[!htbp]
\renewcommand\arraystretch{2}
\centering
\caption{Traditional Detection Methods.}
\label{Tab:Traditional Detection Methods}
\begin{tabular}{cccc}
\hline
Reference & Year & Team & Details\\
\hline
\cite{Basavanhally-2008-Manifold} & 2008 & A Basavanhally & SVM\\

\cite{Basavanhally-2009-Computerized} & 2009 & AN Basavanhally & \tabincell{l}{An algorithm which combines \\region growing and MRF.}\\

\cite{Cheng-2010-Identifying} & 2010 & J Cheng & SVM-RFE\\

\cite{Ishikawa-2014-Gastric} & 2014 & T Ishikawa & SVM\\

\cite{Michail-2014-Detection} & 2014 & E Mchail & LDA\\

\cite{Kuo-2014-Lymphatic} & 2014 & YL Kuo & SVM\\

\cite{Zarella-2015-Lymph} & 2015 & MD Zarella & SVM\\

\hline
\end{tabular}
\end{table}

\subsection{Deep Learning based Detection Methods}

\begin{figure}[!htbp]
\centering
\centerline{\includegraphics[width=0.7\textwidth]{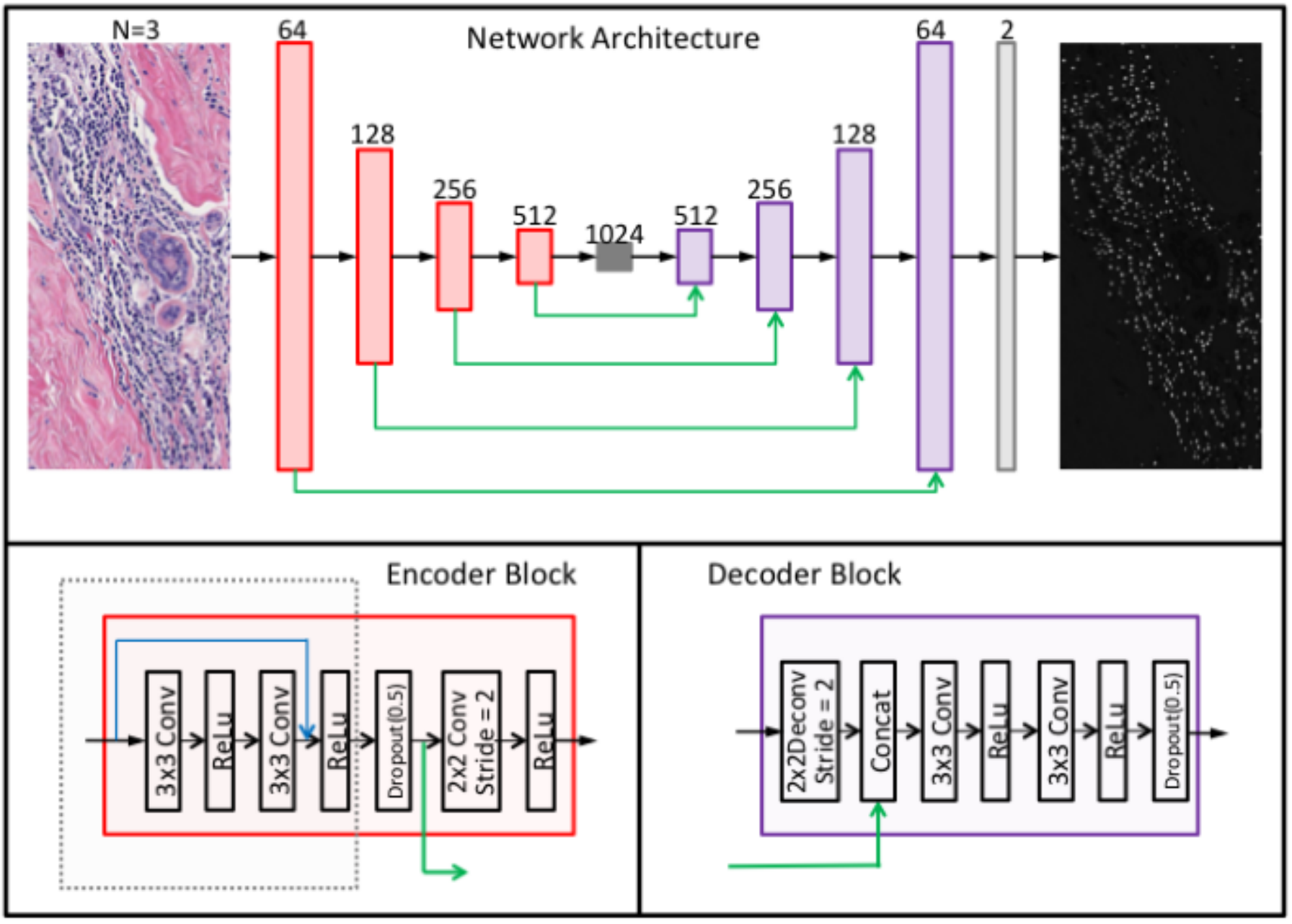}}
\caption{The whole architecture of the network. The figure corresponds to Fig. 1 in~\cite{Chen-2016-Automatic}.}
\label{fig:FCN}
\end{figure}

\begin{figure}[!htbp]
\centering
\centerline{\includegraphics[width=0.7\textwidth]{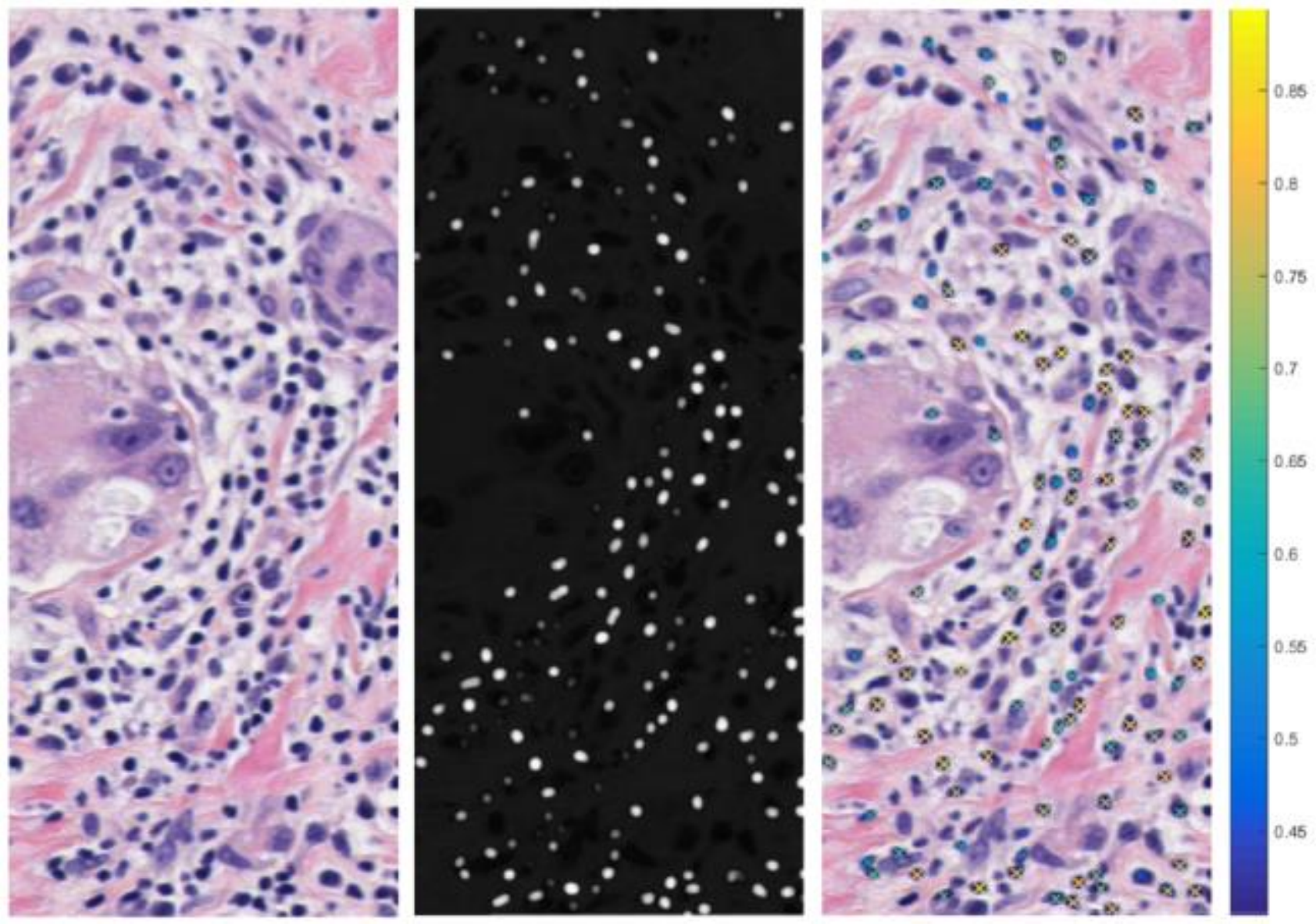}}
\caption{Left: Original image. Middle: A probability map. Right: The visualization result. The color column shows the confidence score of detection. The figure corresponds to Fig. 3 in~\cite{Chen-2016-Automatic}.}
\label{fig:extension of FCN}
\end{figure}

In\cite{Chen-2016-Automatic}, a deep neural network model is designed for detecting lymphocytes in H$\&$E stained images. The model is an extension of FCN (as shown in Fig.~\ref{fig:FCN}) and includes two parts: encoder component and decoder component. In the FCN, each encoder block (red boxes) uses a residual learning function to process images and transforms images into upper scale, then the highest level abstraction is extracted by the bridge block, the decoder block restores the extracted abstraction. In the reference, an extension of FCN is developed: (1) the information is decoded in the bridge block. (2) Rich features are retained in decoder blocks. (3) Pooling layers are replaced by convolution layers. (4) Shortcut link is set in each encoder block. (5) Dropout layers are set in blocks of encoder and decoder. Two datasets are used in the experiment, the first dataset includes 100 fields of view (FOV), the second dataset includes 7335 image patches. Fig.~\ref{fig:extension of FCN} shows an example of detection result. 

\begin{figure}[!htbp]
\centering
\centerline{\includegraphics[width=0.8\textwidth]{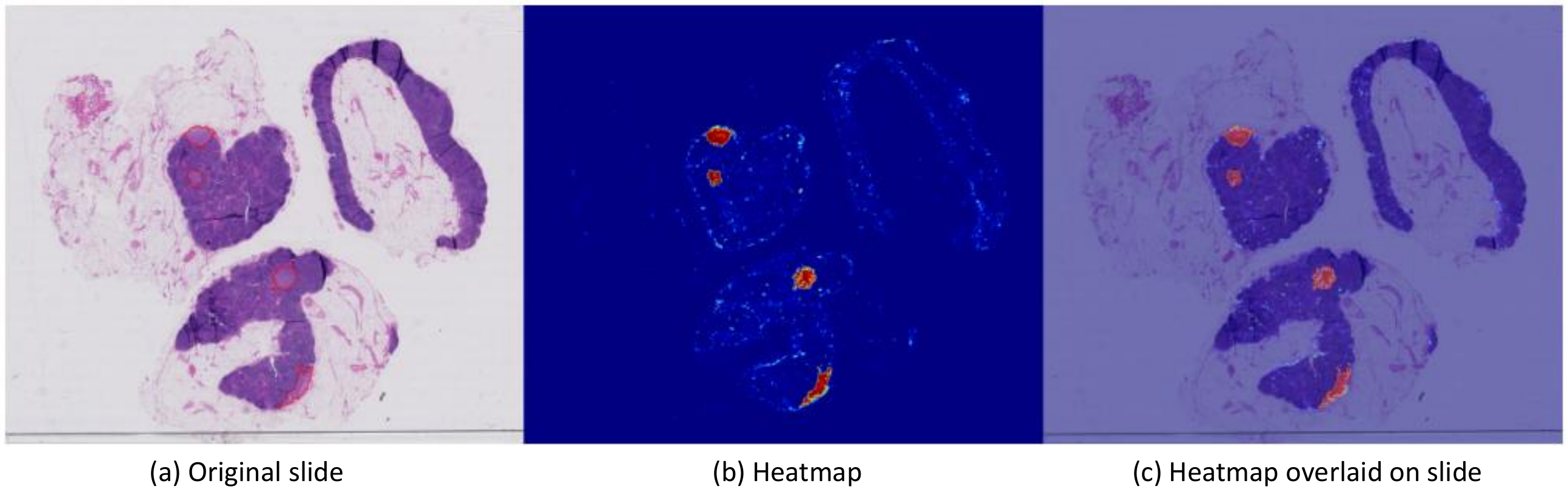}}
\caption{Visualization result of tumor detection. The figure corresponds to Figure. 3 in~\cite{Wang-2016-Deep}.}
\label{fig:GoogLeNet detection}
\end{figure}

In\cite{Wang-2016-Deep}, a framework for detecting cancer metastasis in images of Camelyon16 is proposed. In the method, a deep convolutional neural network is trained with millions of training patches to classify tumor and normal patches. Then, tumor probability heatmaps are constructed for the localization task and the classification task based on the slide. Fig.~\ref{fig:GoogLeNet detection} shows the Visualization result of tumor detection. The GoogLeNet is selected as the deep neural network in the classification module. GoogLeNet has 27 layers and more than 6 million parameters, with characteristics of a fast speed and stability. Finally, an AUC equalling 0.925 and a tumor localization score equalling 0.733 are obtained. The framework wins the tasks in the challenge of Camelyon16.

\begin{figure}[!htbp]
\centering
\centerline{\includegraphics[width=0.8\textwidth]{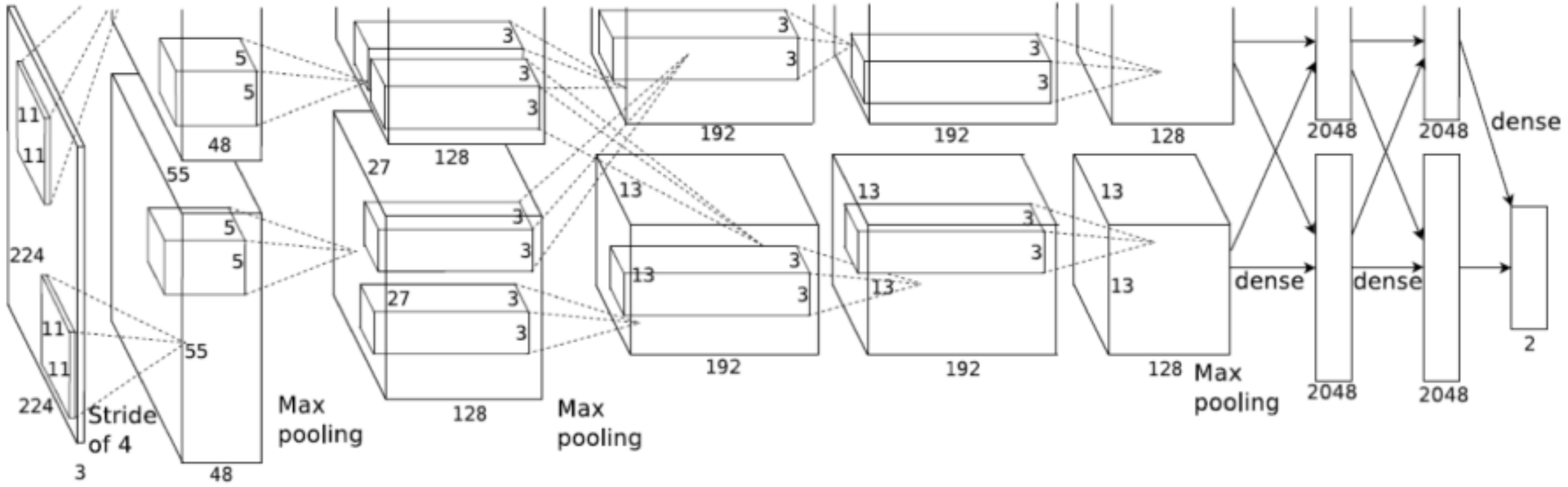}}
\caption{The architecture of AlexNet layer. The figure corresponds to Fig. 2 in~\cite{Chen-2016-Identifying}.}
\label{fig:AlexNet}
\end{figure}

In\cite{Chen-2016-Identifying}, a deep CNN is trained with millions of patches from the Camelyon16 dataset. The steps of preprocessing and feature extraction are described in Sec.~\ref{Sec:3} and Sec.~\ref{Sec:5}. AlexNet is selected in the classification module to discriminate tumor and normal tissue tiles. AlexNet has five convolutional layers, three fully-connected layers, 60 million parameters. Fig.~\ref{fig:AlexNet} shows the architecture of the AlexNet layer. Finally, a sensitivity equalling 0.96, a specificity equalling 0.89 and an AUC equalling 0.90 are obtained. The evaluation result shows the framework is efficient.

In\cite{Xiao-2017-Deep}, a method for identifying and classifying cancer in the dataset of Camelyon17 is developed. ResNeXt architecture with an inputs size of 224$\times$224$\times$3 is selected to initially detect breast cancer metastases of lymph nodes through a tumor probability map. ResNeXt is a deep CNN with 101 layers and has a characteristic of stability and high efficiency. Next, an ensemble learning approach is applied for two classification tasks including 5 types of pN-stage of lymph node metastases. Finally, an average kappa score of 99.5$\%$ is obtained in the three-class classification task through five fold cross validation, an average kappa score of 89.57$\%$ is obtained in the two-class classification task through five fold cross validation.

In\cite{Liu-2017-Detecting}, a framework is proposed to automatically detect and localize tumors of small pixels with 100$\times$100 in microscopy images with a size of 100,000$\times$100,000. A CNN architecture called InceptionV3 is utilized to carry out the detection task in the Camelyon16 dataset. InceptionV3 can process more and richer space features and increase feature diversity. In the experiment, 92.4$\%$ of tumors are detected. The result surpasses the result of 82$\%$ detected by the previous method and the result of 73$\%$ detected manually. Besides, an AUC of more than 97$\%$ is obtained in the Camelyon16 dataset. These results show that the proposed method can reduce the FNR of metastasis detection.

\begin{figure}[!htbp]
\centering
\centerline{\includegraphics[width=0.8\textwidth]{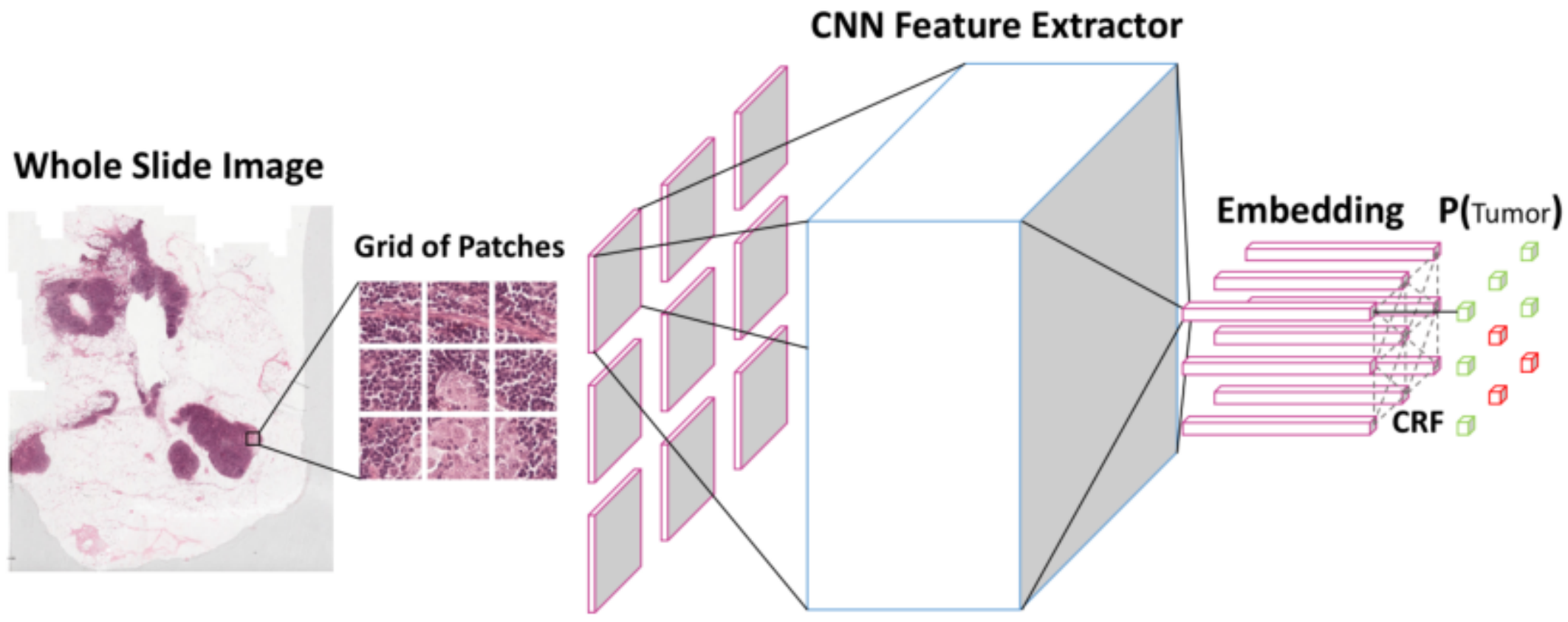}}
\caption{The architecture of NCRF. The figure corresponds to Figure 1 in~\cite{Li-2018-Cancer}.}
\label{fig:NCRF}
\end{figure}

In\cite{Li-2018-Cancer}, a deep learning framework called neural conditional random field (NCRF) is developed to detect metastasis in lymph nodes. Each WSI is divided into small patches, and the spatial correlations of neighboring patches are considered by NCRF through a fully connected CRF. The CRF is incorporated on the top of the feature extraction module of CNN. Fig.~\ref{fig:NCRF} shows the architecture of NCRF. Finally, the framework is applied on the Camelyon16 dataset and an average FROC score equalling 0.8069 is obtained. The result is higher than the previous best result of 0.8074 reported in~\cite{Wang-2016-Deep}.

In\cite{Somaratne-2019-Improving}, a deep learning model is proposed to identify FL. The model improves the identification through transfer learning with fine-tuning. The model increases the accuracy from 58$\%$ to 100$\%$ in a private dataset.

\begin{table}[!htbp]
\renewcommand\arraystretch{2}
\centering
\caption{Deep Learning Methods for Detection.}
\label{Tab:Deep Learning Methods for Detection}
\tiny
\begin{tabular}{cccc}
\hline
Reference & Year & Team & Details\\
\hline

\cite{Chen-2016-Automatic} & 2016 & J Chen & Extension of FCN\\

\cite{Wang-2016-Deep} & 2016 & D Wang & GoogLeNet\\

\cite{Chen-2016-Identifying} & 2016 & R Chen & AlexNet\\

\cite{Xiao-2017-Deep} & 2017 & K Xiao & ResNeXt\\

\cite{Liu-2017-Detecting} & 2017 & Y Liu & InceptionV3\\

\cite{Somaratne-2019-Improving} & 2019 & UV Somaratne & Transfer learning\\

\hline
\end{tabular}
\end{table}

\subsection{Summary}

The reviewed works show that SVM and $k$nn are the commonly used traditional detection methods of LHIA. Deep learning-based detection methods are the main-stream methods and achieve good results, and CNN is the most commonly used deep learning method.

\section{Methodology Analysis}
\label{Sec:8}

This section selects some more suitable methods from each task for LHIA analysing.

\subsection{Analysis of Image Preprocessing in LHIA}

In image preprocessing of LHIA, colour-based techniques are widely used. In addition to these commonly used methods, other methods can also preprocess the image well for subsequent experiments. In~\cite{Fatakdawala-2010-Expectation}, the EM-based method is useful for analyzing other biomedical images. In~\cite{Meng-2010-Histology,Meng-2013-Multimodal}, the method of dividing an image into blocks is suitable for relatively large images in order to reduce the space complexity of feature extraction and extract different local features. In~\cite{Belkacem-2010-Segmentation, Schafer-2013-Image, Es-2017-Decision}, the region-based method can remove isolated tissue areas in the image, so the image with many noises in the background can be preprocessed with this method. In~\cite{Linder-2019-Deep, Li-2020-Multi, Brancati-2019-Deep, Somaratne-2019-Improving, Senaras-2019-Segmentation, Kandel-2020-Novel, Thorat-2020-Classification,Dif-2020-Efficient, Hashimoto-2020-Multi}, data augmentation methods are used to avoid the issue of biases and overfitting, Data enhancement methods are used to avoid bias and overfitting problems, but in addition to these, this method can also solve the problem of too few images in the dataset. 

\subsection{Analysis of Image Segmentation in LHIA}

In the process of image segmentation in LHIA, the threshold-based and $k$-means clustering methods are widely used. In~\cite{Senaras-2017-Foxp3}, optimal adaptive thresholding is used to minimize under- and over-segmented nuclei. The sensitivity of 100$\%$ and maximum precisions of 100$\%$ show the method is useful in segmenting histopathology images. However, the method of global and local threshold only obtains a precision of 91.79$\%$ in~\cite{Mandyartha-2020-Global}. Adaptive thresholding can also obtain a good result in segmenting liver cysts~\cite{Zhu-2017-Segmentation}. The threshold segmentation method has the advantages of simple calculation, high calculation efficiency, fast speed, and easy implementation. It has been widely used in applications where calculation efficiency is important. It has a better segmentation effect on images with large contrast between the target and the background, and it can always use closed and connected boundaries to define non-overlapping areas. But it is not suitable for multi-channel images and images with little correlation between eigenvalues, and it is difficult to obtain accurate results for image segmentation problems where there is no obvious grayscale difference in the image or the grayscale value range of each object has a large overlap.

The main advantages of $k$-Means include relatively simple principle, easy implementation, fast convergence speed, better clustering effect, better interpretability of the algorithm, and the main parameter that needs to be adjusted is only the number of clusters $k$. However, there are some problems when using $k$-means. For example, the value of $k$ is not easy to select, and it is difficult to converge for data sets that are not convex. And if the data of each hidden category is not balanced, the clustering effect is not good. With the iterative method, the result may only be a local optimum, and the method is more sensitive to noise and abnormal points.

In addition to the above-mentioned commonly used segmentation methods, there are some more potential segmentation methods, such as U-net~\cite{Swiderska-2019-Learning}. U-net network is shaped like U and is suitable for medical image segmentation~\cite{Ronneberger-2015-U}. The U-net structure is symmetrical, with the encoder on the left to extract features and the decoder on the right to output the encoded features as pictures. Now, U-net is not only used for the segmentation of lymphoma histopathology images, but also for corneal neuropathology images~\cite{Colonna-2018-Segmentation}. In~\cite{Ojeda-2020-Convolutional}, the parasites in the blood are segmented by U-net to complete the experiment more efficiently.



\subsection{Analysis of Classification and Detection Methods in LHIA}

In the process of classification and detection methods in LHIA, SVM is the most commonly used method, but the result of SVM is not very excellent. For example, SVM is used to classify CBs and non-CBs in~\cite{Sandhya-2013-Automated}, an accuracy of 84.53$\%$ is obtained. The result does not look good in the current situation where classification methods are emerging one after another. SVM can achieve much better results than other algorithms on the small sample training set, with a low generalization error rate and fast classification speed. However, SVM is difficult to implement for large-scale training set samples, has difficulties in multi-classification problems, and is sensitive to the selection of parameters and kernel functions. 

The $k$nn algorithm is relatively simple, easy to understand and implement, and does not require parameter estimation or training. $k$nn is suitable for classifying rare events. For multi-classification problems, it performs better than SVM. However, when the sample is unbalanced, it may cause classification errors.

Most of the current image classification and detection tasks use deep learning methods, and CNN is the most widely used algorithm. Compared with general neural networks, the most prominent feature of CNN is the addition of convolutional layers and pooling layers. It does not need to extract specific artificial features for specific image data sets or classification methods but simulates the visual processing mechanism of the brain to image hierarchy abstraction, automatic screening of features, so as to achieve classification. In recent years, CNN has also achieved good results in the classification and detection of histopathological images of lymphoma. For example, CNN is used to classify normal and tumor slides in images of Camelyon16, an AUC of 0.94 which surpasses the result of the winner of Camelyon16 Challenge is obtained~\cite{Jamaluddin-2017-Tumor}. the residual CNN is used to classify lymphomas subtypes, an ACC of 97.67$\%$ which is higher than U-net and ResNet is obtained~\cite{Brancati-2019-Deep}. 92.4$\%$ of tumors are detected through a CNN architecture called Inception V3 in~\cite{Liu-2017-Detecting}. CNN obtains good results on histopathological images of lymphoma, but also on other pathological images, such as breast cancer~\cite{Titoriya-2019-Breast,Kumar-2017-Comparative}, prostate cancer~\cite{Duran-2020-Prometeo,Lucas-2019-Deep}, colon cancer~\cite{Yildirim-2021-Classification,Tasnim-2021-Deep}. 

\subsection{Potential Methods of LHIA}

In this section, we introduce some potential methods of LHIA.

\subsubsection{Potential Methods of Segmentation for LHIA}

In the summarized papers, in addition to the commonly used segmentation methods, there are some other potential methods.

\begin{figure}[!htbp]
\centering
\centerline{\includegraphics[width=0.96\textwidth]{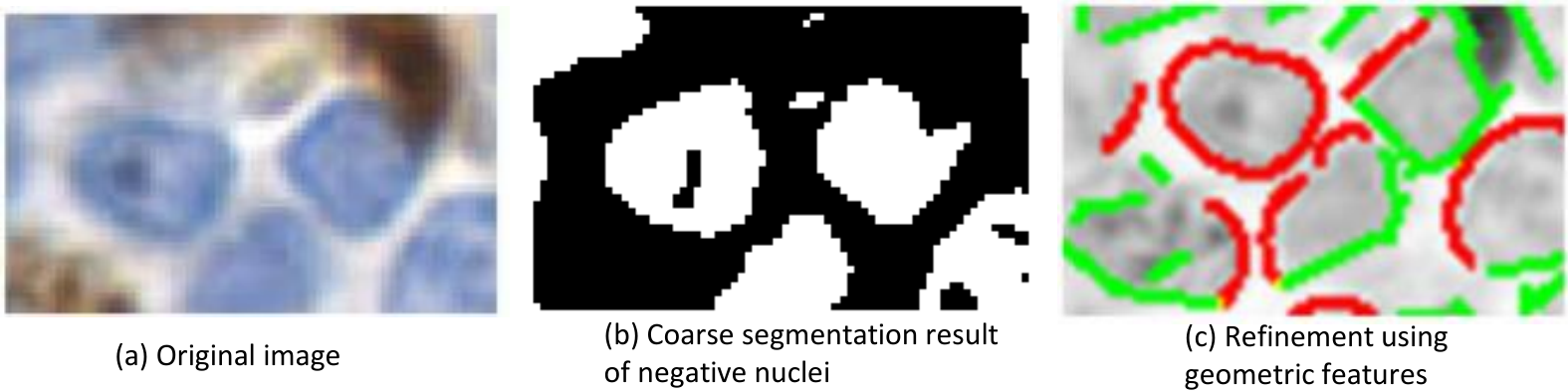}}
\caption{The process of coarse segmentation. The figure corresponds to Figure 1 in~\cite{Senaras-2017-Foxp3}.}
\label{fig:OAT1}
\end{figure}

\begin{figure}[!htbp]
\centering
\centerline{\includegraphics[width=0.7\textwidth]{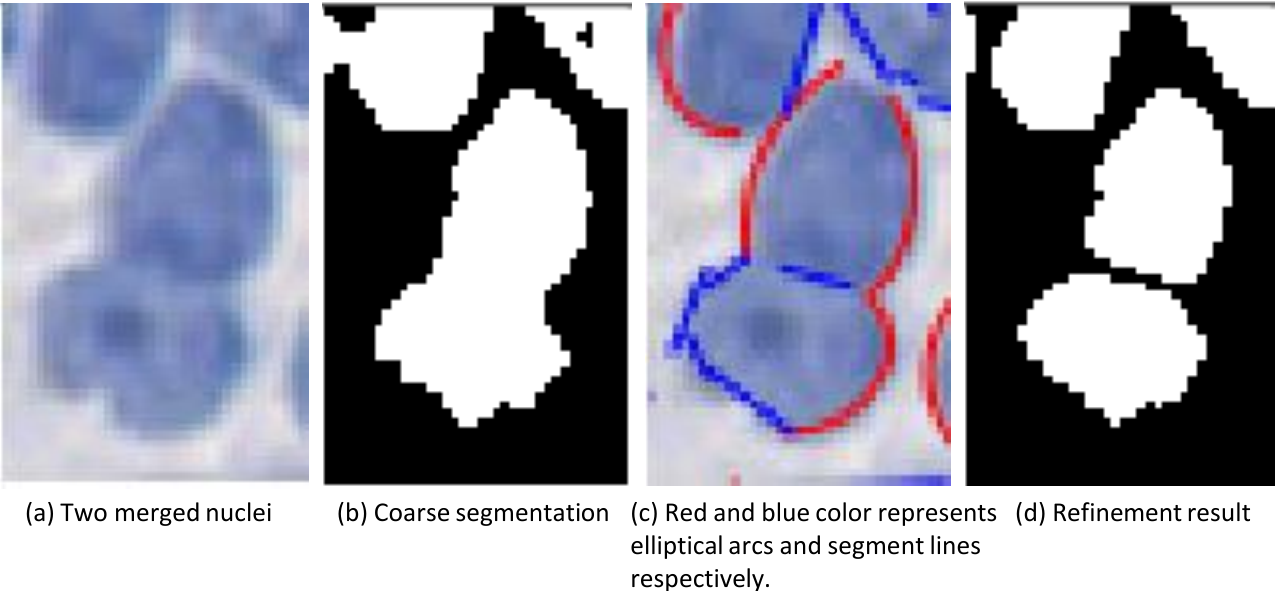}}
\caption{The process of splitting overlapping cells. The figure corresponds to Figure 3 in~\cite{Senaras-2017-Foxp3}.}
\label{fig:OAT2}
\end{figure}

In\cite{Senaras-2017-Foxp3}, a new method for detecting and counting nuclei from FOXP3 stained FL images is proposed. The method defines a new process of adaptive thresholding, which is called optimal adaptive thresholding (OAT). OAT is used to minimize under- and over-segmented nuclei for coarse segmentation. Then, a parameter free elliptical arc and line segment detector are integrated to refine segmentation results and split most of overlapping nuclei. Finally, the remaining overlapping nuclei are split using \emph{Simple Linear Iterative Clustering} (SLIC). The dataset used in the method is composed of 13 ROIs including 769 negative nuclei and 88 positive nuclei. The maximum sensitivity for detecting negative nuclei and positive nuclei is 100$\%$ and 95$\%$ respectively, the maximum precisions are both 100$\%$. The results show that the proposed method can be used to study the influence of FOXP3 positive cell nuclei on the prognosis of FL. Fig.~\ref{fig:OAT1} shows the results of coarse segmentation and refinement. Fig.~\ref{fig:OAT2} shows the process of splitting overlapping cells. The method is potential in splitting overlapping nuclei.

In\cite{Fatakdawala-2010-Expectation}, expectation-maximization driven geodesic active contour with overlap resolution (EMaGACOR) algorithm is used to automatically detect and segment lymphocytes in histopathology images of breast cancer. First, the expectation maximization algorithm is used to cluster the tissue components into relative groups. Next, the centroids of the clusters are used as seed points for automatically initializing the magnetostatic active contour (MAC), which is applied as the Geodesic Active Contour (GAC) model. The method is tested on 100 HER+ breast cancer histology images and its detection sensitivity is more than 86$\%$, the positive predictive value is more than 64$\%$. Besides, the method can split more than 90$\%$ overlapping lymphocytes, HD and MAD are 2.1 and 0.9 pixels respectively. The results show that the segmentation method is efficient and helpful in analyzing other biomedical images.

In~\cite{Tosta-2017-Application}, an evaluation of evolutionary algorithm is presented, which is used to segment neoplastic cellular in lymphoma images. First, the R channel of RGB color space is histogram equalized and filtered by a Gaussian filter. Next, fuzzy 3-partition is used to define two thresholds to identify neoplastic nuclei, cytoplasm and background tissue. The method is tested on 62 FL and 99 MCL images and obtains a mean value accuracy equalling 99.38$\%$. The result shows that the method is potential in segmenting cells in histopathology images.

\subsubsection{Potential Methods of Feature Extraction for LHIA}

In the summarized papers, in addition to the commonly used feature extraction methods, there are some other potential methods.

In\cite{Cooper-2009-Feature}, a method of automatic non-rigid registration extracts matching high level features on the down-sampled images of original. Then, the match candidates are formed by matching features between the base images and float images. The accuracy of the method is equivalent to manual registration. The method can be applied to extract traditional features.

In\cite{Song-2016-Histopathology}, a supervised intra-embedding algorithm is designed, which has a multilayer neural network model. The algorithm can address some problems of Fisher vectors including high dimensionality and bursty visual elements. In the algorithm, the ConvNet-based Fisher vectors are transformed into a more discriminative feature representation. The classification ACC of the algorithm in the IICBU-2008 dataset is up to 99.2$\%$. The result shows the method is efficient in enhancing the ConvNet-based Fisher vectors.

In\cite{Roberto-2021-Fractal}, features are obtained from histology images through techniques including multidimensional fractal models and defining convolutional descriptors with different CNN models. The method shows that the feature association extracted from different methods is a potential field for further research.

\subsubsection{Potential Methods of Classification for LHIA}

In the summarized papers, in addition to the commonly used classification methods, there are some other potential methods.

In\cite{Song-2016-Histopathology}, a method for classifying histopathology images is proposed, which is based on the Fisher vector. A dimension reduction method for resolving high dimension of features is designed, two local descriptions including texture-based features and unsupervised features are combined to obtain Fisher vectors in image-level. SVM is applied for classification through four-fold cross validation using the IICBU-2008 dataset and a classification accuracy equalling 93.3$\%$ is obtained, which is higher than other methods. The result shows that the method is potential in classifying other datasets.

In\cite{Jiang-2018-Effective}, a hierarchical classification model is developed, which is based on the labels' statistics. First, the images are converted into a grayscale channel. Then, each image is divided into 130 non-overlapped patches. Next, the sparse autoencoder is used to extract texture features of these patches. Finally, an extreme learning machine and SVM classifier are applied to classify NHL into FL, CLL and MCL. The dataset includes 31,200 training patches, 17,420 testing patches. The method obtains an ACC of 97.96$\%$ higher than the results of other methods. The result shows that the method is potential in classifying other datasets.

In\cite{Martins-2019-Colour}, the fractal features extracted from RGB and $L^*a^*b$ color spaces are used in classification of histopathology images of lymphoma. The fractal features are concatenated to be feature vectors and the Hermite polynomial classifier is selected to evaluate the performance of the method. The method is potential in classifying other cancers.

\subsubsection{Potential Methods of Detection for LHIA}

In\cite{Sertel-2010-Computer}, a two-step method is used to detect CBs in FL images. In the first step, non-CBs are detected through the size and shape of cells. In the second step, CBs are screened through utilizing the texture of non-CBs. A detection accuracy of 80.7$\%$ is obtained through 100 ROIs of 10 FL samples. In~\cite{Belkacem-2010-Effect}, CBs are detected through extracting color, morphology features.

\begin{figure}[!htbp]
\centering
\centerline{\includegraphics[width=0.6\textwidth]{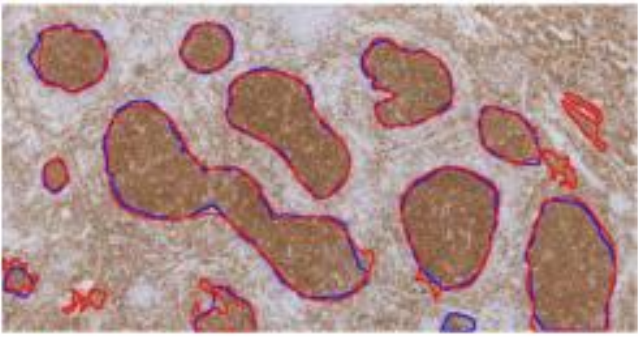}}
\caption{The blue boundaries: manually annotations, red boundaries: generation of the algorithm. This figure corresponds to Fig. 6 in~\cite{Samsi-2010-Detection}.}
\label{fig:CD10 follicles}
\end{figure}

In\cite{Samsi-2010-Detection}, the work is to identify follicles in FL images, for the purpose of improving FL grading. Different measurements of color and texture are used in the identification of follicles, an iterative watershed algorithm is used to split merged regions. Finally, smooth follicle boundaries are obtained through Fourier descriptors. eight CD10 stained FL images are used in the experiment, a result of an average similarity score equaling 87.11$\%$ is obtained when comparing the computer with manual segmentation. Fig.~\ref{fig:CD10 follicles} shows the follicles identified by the algorithm.

\begin{figure}[!htbp]
\centering
\centerline{\includegraphics[width=0.6\textwidth]{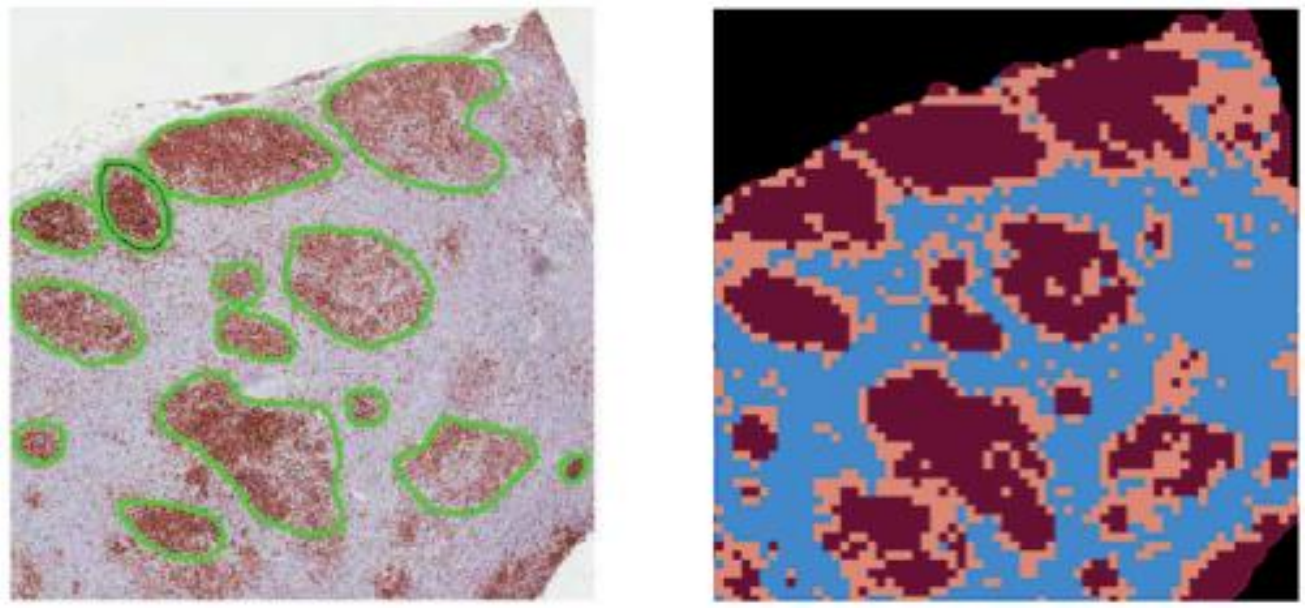}}
\caption{Left: manually identification (green lines). Right: identification result of the algorithm (dark brown and light brown parts). This figure corresponds to Figure 2 in~\cite{Han-2010-Multi}.}
\label{fig:Gaussian detection of follicle}
\end{figure}

In\cite{Han-2010-Multi}, a method for detecting follicle in IHC stained FL images is proposed. The novelty of this method is processing WSIs and Giga-byte scaled images using multi-resolution. The images are set to Gaussian pyramids with multi-resolution to represent the visual patterns at different scales. The images are set to Gaussian pyramids with multi-resolution to represent the visual patterns at different scales. Two IHC stained images are used in the experiment, and an average segmentation accuracy of 83.09$\pm$6.25 is obtained. Fig.~\ref{fig:Gaussian detection of follicle} shows the identification result.

In\cite{Oger-2012-General}, a method for automatically detecting follicle regions in FL images is proposed, which is a feature-based clustering method. Color and texture features are extracted through conversion of color space, co-occurrence matrix. Each pixel in images is represented by a feature vector constructed by the extracted features. Then, the feature vectors are classified into follicles, intra-follicular regions, mixture and background regions through $k$-means classifier. Finally, a method through accurate follicle boundaries is used to improve the quality of segmenting follicle regions.

In\cite{Samsi-2012-Efficient}, the module of detecting follicle areas in FL images includes the following steps: First, the extraction of color and texture through conversion of color space, median filter and co-occurrence matrix. Second, feature vectors are constructed by the extracted features. Finally, the feature vectors are used in $k$-means clustering to obtain follicle areas.

\begin{figure}[!htbp]
\centering
\centerline{\includegraphics[width=0.6\textwidth]{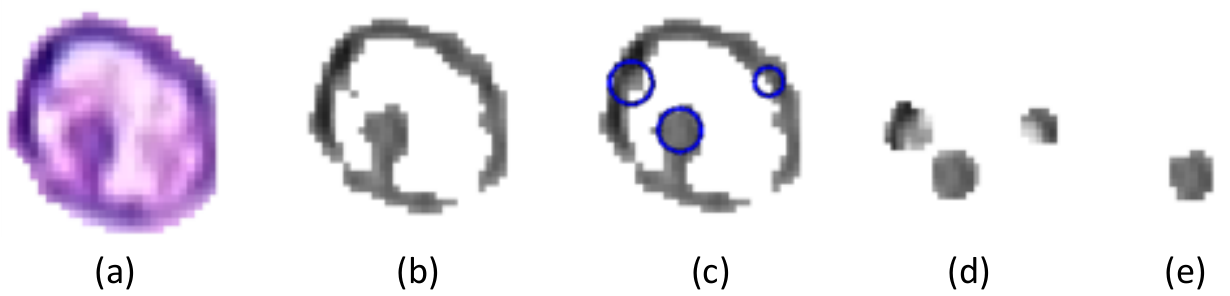}}
\caption{(a) Original image. (b) The area containing nuclei. (c) Hough transform. (d) The inner area of the circle detected by the Hough transform. (e) Nuclei. The figure corresponds to Fig. 2 in~\cite{Dimitropoulos-2014-Using}.}
\label{fig:adaptive detect CBs}
\end{figure}

In\cite{Dimitropoulos-2014-Using}, a method for detecting CBs in FL microscopic images is proposed. The step of segmentation is described in Sec~\ref{Sec:4}. In the step of detection, an algorithm for detecting nuclei is developed. Fig.~\ref{fig:adaptive detect CBs} shows the steps of the algorithm. Intensity threshold is used to filter out areas that do not contain nuclei, the numbers of different intensity areas are estimated by expectation maximization with minimum description length. Next, Hough transform is used in the areas containing nuclei to detect nuclei. Finally, in the step of classification, a neuro-fuzzy classifier with multiple levels is developed to classify CBs and non-CBs. The method is based on knowledge or rules, so the classifier can be constructed through the form of fuzzy rules in the knowledge of pathologists. The dataset includes 120 CBs and 180 non-CBs and an average number equalling 90.35$\%$ of CBs are detected by the classifier. The performance of this classifier is better than other classifiers including Adaboost, SVM, neural network and classification tree.

\section{Conclusions and Future Work}
\label{Sec:9}

This paper summarizes the LHIA methods based on machine vision. The data sets, evaluation methods and indicators, image preprocessing, segmentation, feature extraction, classification, and detection are comprehensively analyzed and summarized.

Reviewing all the work, we can find that since 2010, deep learning has promoted technological changes. Deep learning replaces artificial features and improves the accuracy of segmentation, classification, and detection. The Camelyon dataset and IICBU-2008 dataset are commonly used in LHIA. In image segmentation, $k$-means and clustering are more commonly used methods, and U-net has become the mainstream image segmentation method. In image preprocessing, the methods based on color and threshold are used frequently. There are also many methods for extracting local features by dividing an image into blocks. Among the feature extraction methods, traditional features such as color, texture, and shape are the most commonly used features, but as deep learning becomes more and more powerful, deep learning features gradually replace manual feature extraction methods. Among classification and detection methods, deep learning methods are also the current mainstream methods, especially CNN-based methods. 

In the future, diagnosing and treating lymphoma methods based on deep learning has extensive application prospects. Especially since the new crown epidemic outbreak in 2019, medical image analysis technology has received more and more attention. At present, medical image analysis is still an emerging area. Therefore, developing a system that requires a small amount of calculation, small memory, and interpretability is essential.

\section*{Acknowledgements}
This work is supported by the ``National Natural Science Foundation of China'' (No. 61806047), 
``Sichuan Science and Technology Program'' (No. 2021YFH0069, 2021YFQ0057, 1614 2022YFS0565). 
We also thank Miss Zixian Li and Mr. Guoxian Li for their important discussion.

\section*{Conflicts of Interest}
There is no conflicts of interest in this paper.

\bibliography{Xiaoqi}
\end{document}